\documentclass[a4paper,10pt,notitlepage]{report}
\pdfoutput=1
\usepackage[bookmarks=false]{hyperref}
\usepackage[numbers]{natbib}
\usepackage[margin=1in]{geometry}

\usepackage[utf8]{inputenc}

\usepackage{myTutorialPaper,notation}

\begin{document}
\title{\textbf{Using Inertial Sensors for Position and Orientation Estimation}}

\author{Manon~Kok$^\star$, Jeroen~D.~Hol$^\dagger$ and Thomas~B.~Sch\"on$^\ddagger$ \\
\small{$^\star$Department of Engineering, University of Cambridge, Cambridge, United Kingdom} \\
\small{E-mail: mk930@cam.ac.uk} \\
\small{$^\dagger$Xsens Technologies B.V., Enschede, the Netherlands} \\
\small{E-mail: jeroen.hol@xsens.com} \\
\small{$^\ddagger$Department of Information Technology, Uppsala University, Sweden} \\
\small{E-mail: thomas.schon@it.uu.se} \\
}


\newcommand{\coverTitle}{Using Inertial Sensors for Position and Orientation Estimation}
\newcommand{\coverYear}{2017}

\newcommand{\coverAuthors}{Manon~Kok$^\star$, Jeroen~D.~Hol$^\dagger$ and Thomas~B.~Sch\"on$^\ddagger$ \\ \vspace{3mm}
\small{$^\star$Delft Center for Systems and Control, Delft University of Technology, the Netherlands\footnote{At the moment of publication Manon Kok worked as a Research Associate at the University of Cambridge, UK. A major part of the work has been done while she was a PhD student at Link\"oping University, Sweden.}} \\
\small{E-mail: m.kok-1@tudelft.nl} \\
\small{$^\dagger$Xsens Technologies B.V., Enschede, the Netherlands} \\
\small{E-mail: jeroen.hol@xsens.com} \\
\small{$^\ddagger$Department of Information Technology, Uppsala University, Sweden} \\
\small{E-mail: thomas.schon@it.uu.se} 
}

\begin{titlepage}
\begin{center}
%

\vspace*{2.5cm}
%
{\Huge \bfseries \coverTitle  \\[0.4cm]}

%
{\Large \coverAuthors \\[1.5cm]}

\renewcommand\labelitemi{\color{red}\large$\bullet$}
\begin{itemize}
\item {\Large \textbf{Please cite this version:}} \\[0.4cm]
\normalsize
Manon Kok, Jeroen D. Hol and Thomas B. Sch\"on (2017), "Using Inertial Sensors for Position and Orientation Estimation", Foundations and Trends in Signal Processing: Vol. 11: No. 1-2, pp 1-153. http://dx.doi.org/10.1561/2000000094 
\end{itemize}

\end{center}

\vspace{1cm}

\begin{abstract}
\noindent In recent years, \gls{mems} inertial sensors (3D accelerometers and 3D gyroscopes) have become widely available due to their small size and low cost. Inertial sensor measurements are obtained at high sampling rates and can be integrated to obtain position and orientation information. These estimates are accurate on a short time scale, but suffer from integration drift over longer time scales. To overcome this issue, inertial sensors are typically combined with additional sensors and models. In this tutorial we focus on the signal processing aspects of position and orientation estimation using inertial sensors. We discuss different modeling choices and a selected number of important algorithms. The algorithms include optimization-based smoothing and filtering as well as computationally cheaper extended Kalman filter and complementary filter implementations. The quality of their estimates is illustrated using both experimental and simulated data.
\end{abstract}

\vfill

\end{titlepage}

\glsreset{mems}

\tableofcontents

\chapter{Introduction}
\label{cha:introduction} 
In this tutorial, we discuss the topic of position and orientation estimation using inertial sensors. We consider two separate problem formulations. The first is estimation of orientation only, while the other is the combined estimation of both position and orientation. The latter is sometimes called \textit{pose estimation}. We start by providing a brief background and motivation in \Sectionref{sec:intro-background} and explain what inertial sensors are and give a few concrete examples of relevant application areas of pose estimation using inertial sensors. In \Sectionref{sec:intro-imusForPose}, we subsequently discuss how inertial sensors can be used to provide position and orientation information. Finally, in \Sectionref{sec:intro-outline} we provide an overview of the contents of this tutorial as well as an outline of subsequent chapters. 

\section{Background and motivation}
\label{sec:intro-background}
\begin{figure}
  \centering
    \subfigure[Left bottom: an Xsens MTx \gls{imu}~\citep{xsens-tutorial}. Left top: a Trivisio Colibri Wireless \gls{imu}~\citep{trivisio-tutorial}. Right: a Samsung Galaxy S4 mini smartphone.]{
  \includegraphics[height = 0.35\textwidth]{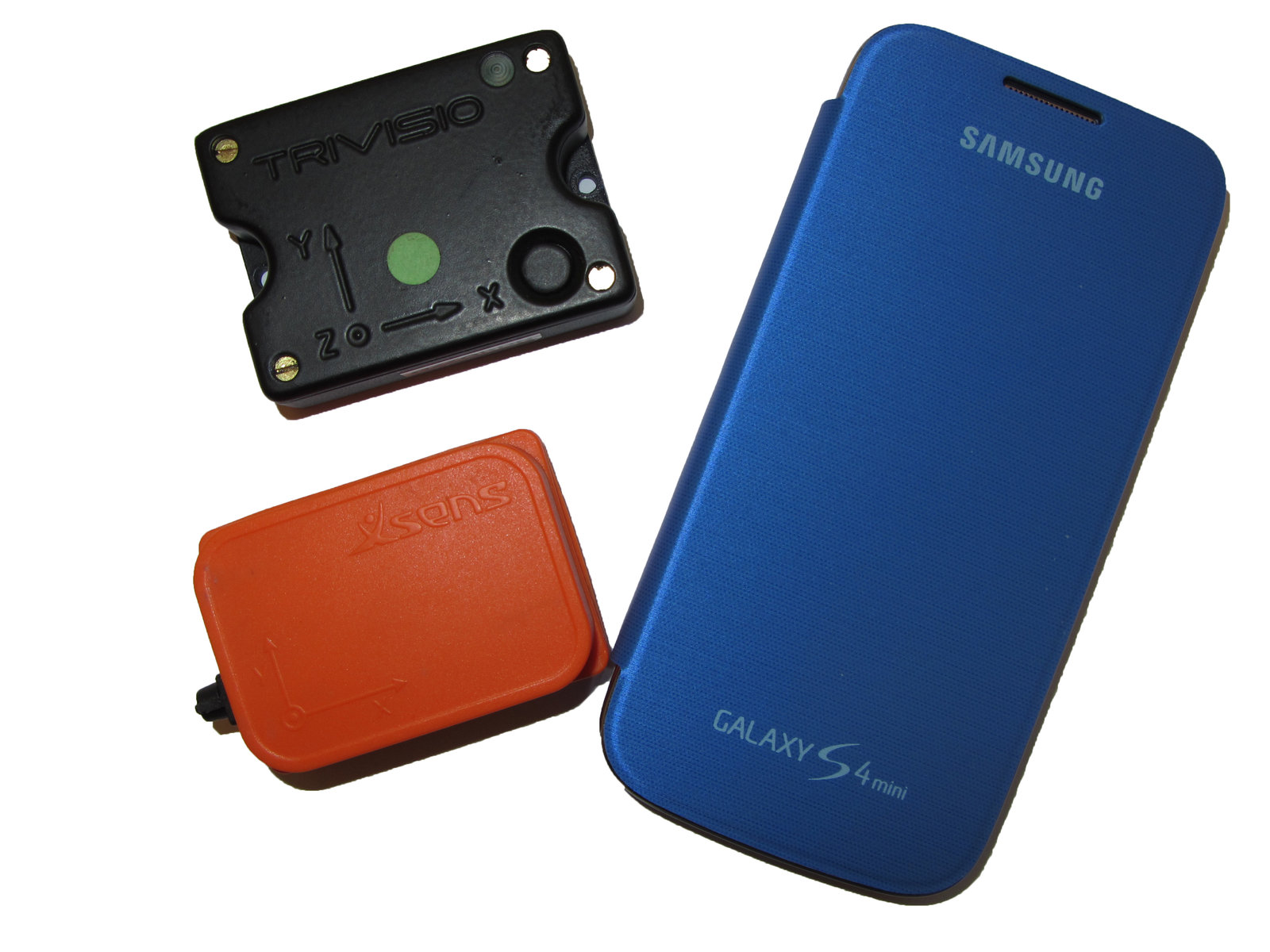}} \\
  \subfigure[A Samsung gear \gls{vr}.$^1$ ]{
  \includegraphics[height = 0.3\textwidth]{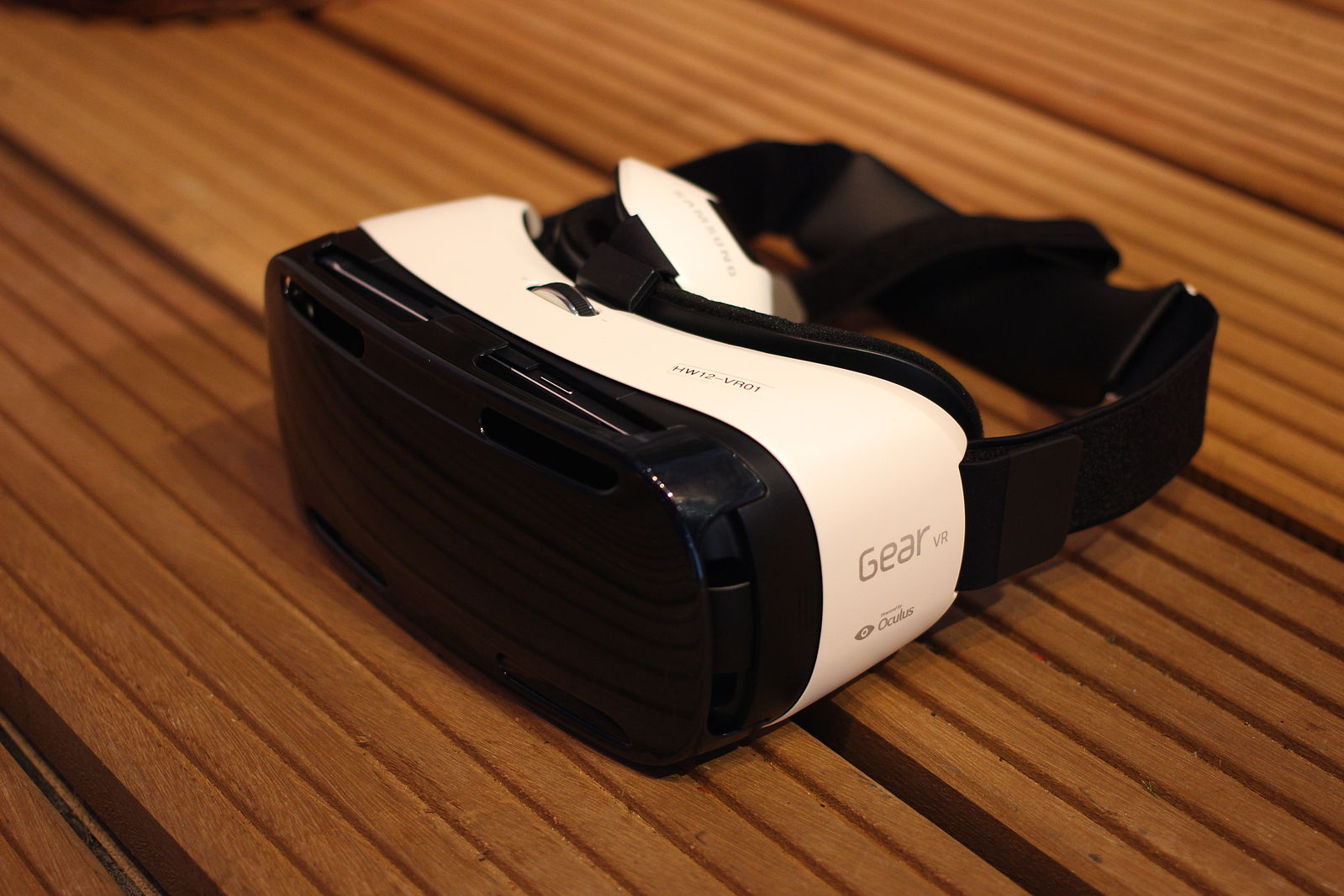}} \hspace{3mm}
    \subfigure[A Wii controller containing an accelerometer and a MotionPlus expansion device containing a gyroscope.$^2$]{
  \includegraphics[height = 0.3\textwidth]{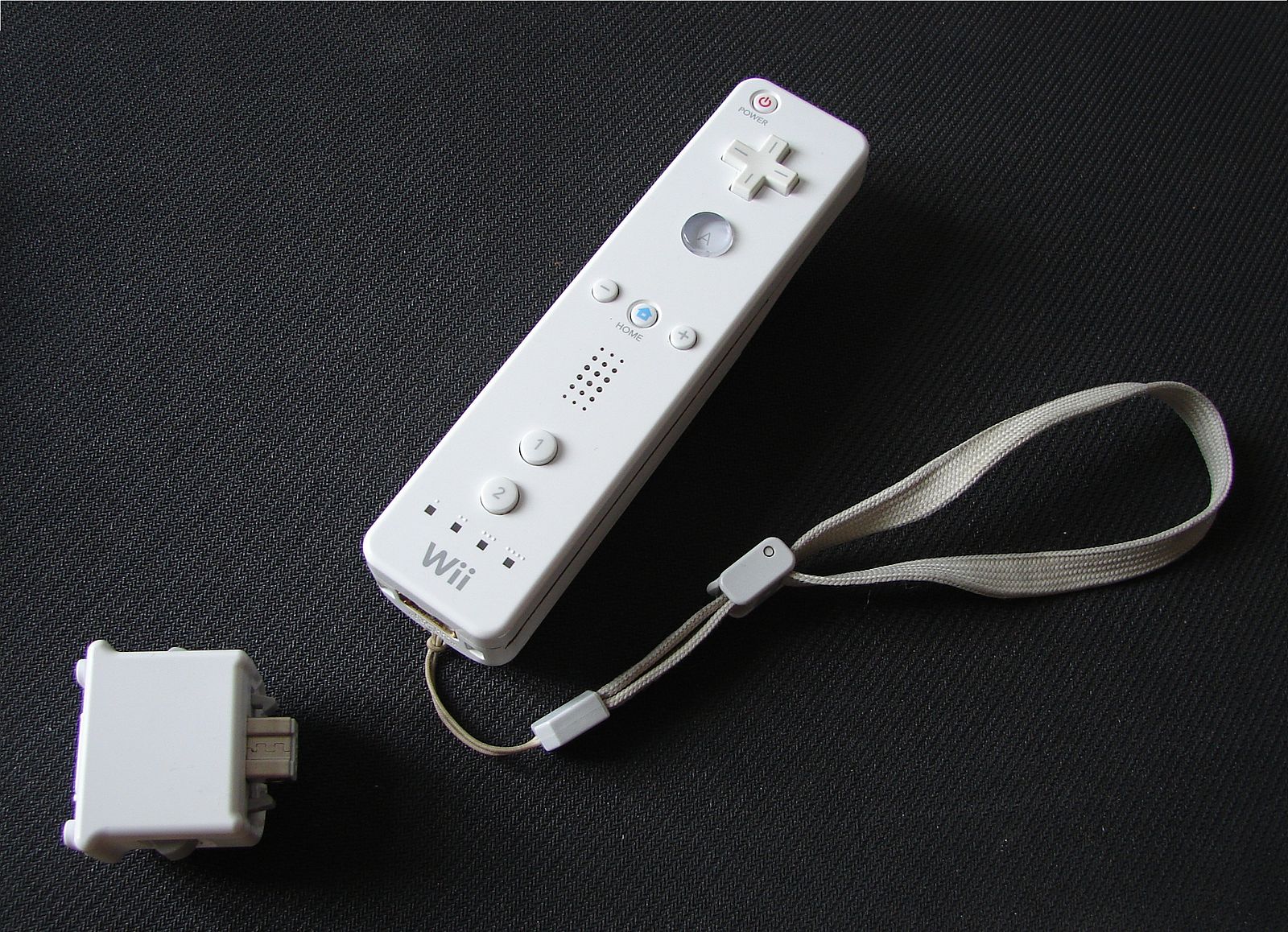}}
  \caption{Examples of devices containing inertial sensors.}
  \vspace{2cm} 
  \footnoterule
  \vspace{-0.3cm}
  \flushleft \setlength{\rightskip}{0pt} 
  \footnotesize{\hspace{1em} $^1$ `Samsung Gear \gls{vr}'  available at \url{flic.kr/photos/pestoverde/15247458515} under CC BY 2.0 (\url{http://creativecommons.org/licenses/by/2.0}).}
  
  \footnotesize{\hspace{1em} $^2$ `WiiMote with MotionPlus' \hspace{0.3pt} by \hspace{0.3pt} Asmodai \hspace{0.3pt} available \hspace{0.3pt} at \hspace{0.3pt} \url{https://commons.wikimedia.org/wiki/File:WiiMote_with_MotionPlus.JPG} under CC BY SA (\url{https://creativecommons.org/licenses/by-sa/3.0/}).}
  \label{fig:intro-inertialSensors}
\end{figure}

The term \textit{inertial sensor} is used to denote the combination of a three-axis accelerometer and a three-axis gyroscope. Devices containing these sensors are commonly referred to as \glspl{imu}. Inertial sensors are nowadays also present in most modern smartphone, and in devices such as Wii controllers and \gls{vr} headsets, as shown in \Figureref{fig:intro-inertialSensors}. 

\setcounter{footnote}{2}

A gyroscope measures the sensor's \emph{angular velocity}, \ie the rate of change of the sensor's orientation. An accelerometer measures the \emph{external specific force} acting on the sensor. The specific force consists of both the \emph{sensor's acceleration} and the \emph{earth's gravity}. Nowadays, many gyroscopes and accelerometers are based on \gls{mems} technology. \Gls{mems} components are small, light, inexpensive, have low power consumption and short start-up times. Their accuracy has significantly increased over the years. 

\begin{figure}
  \centering
  \subfigure[Back pain therapy using serious gaming. \Glspl{imu} are placed on the 
chest-bone and on the pelvis to estimate the movement of the upper body 
and pelvis. This movement is used to control a robot in the game and 
promotes movements to reduce back pain.]{\includegraphics[height = 0.28\textwidth]{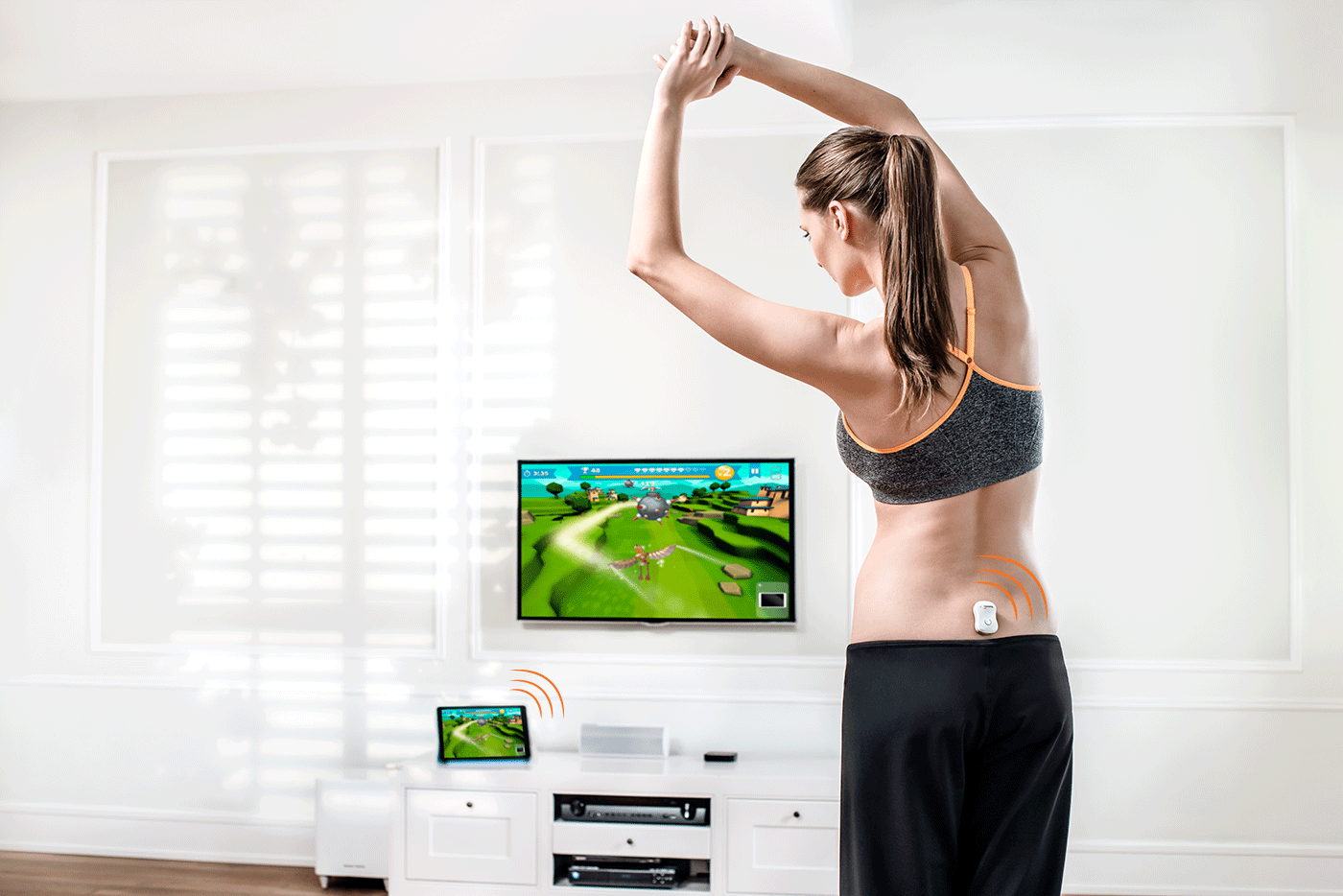}} \hspace{2mm}
  \subfigure[Actor Seth MacFarlane wearing 17 \glspl{imu} to capture his motion and 
animate the teddy bear Ted. The \glspl{imu} are placed on different body segments and provide information about the relative position and orientation of each of these segments.]{\includegraphics[height = 0.28\textwidth]{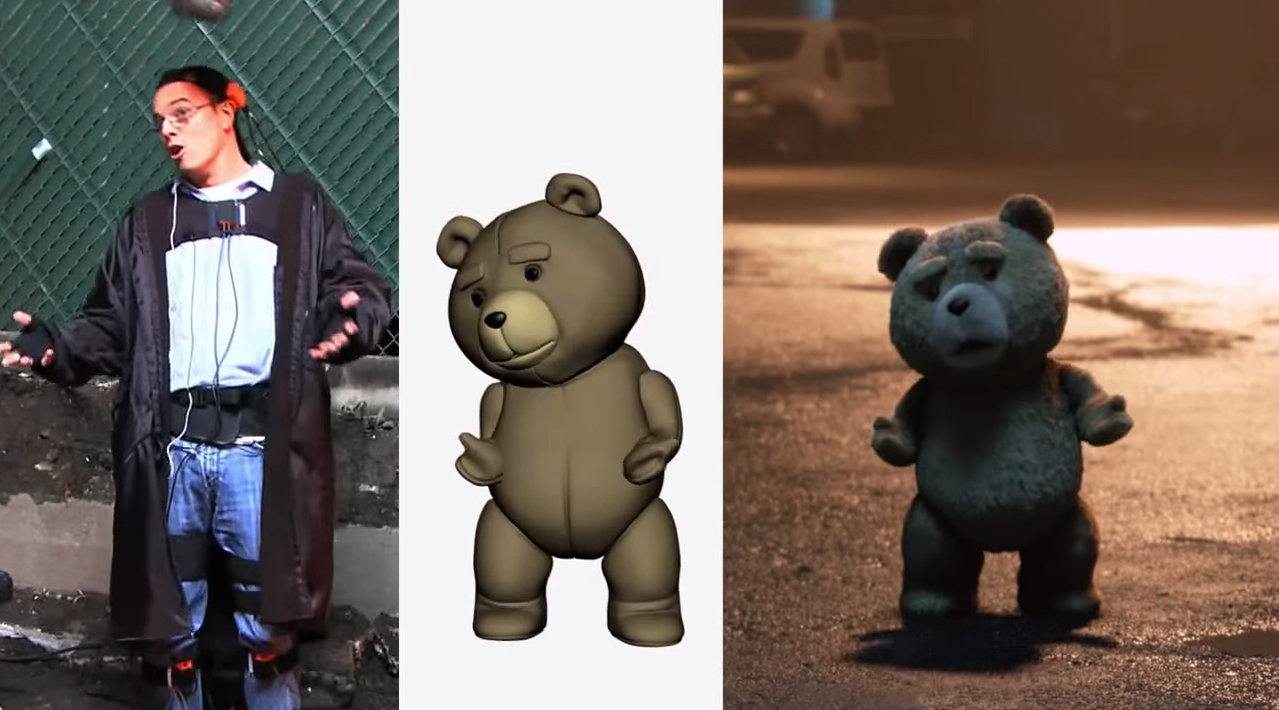} \label{fig:intro-motionCaptureApplications-ted}} \\
  \caption{Examples illustrating the use of multiple \glspl{imu} placed on the 
human body to estimate its pose. Courtesy of Xsens Technologies.}
  \label{fig:intro-motionCaptureApplications}
\end{figure}

\begin{figure}
  \centering
  \subfigure[Inertial sensors are used in combination with \gls{gnss} measurements to estimate 
the position of the cars in a challenge on cooperative and autonomous 
driving.]{\includegraphics[height = 0.3\textwidth]{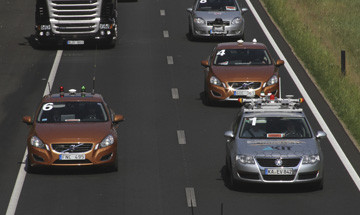} } \hspace{2mm}
  \subfigure[Due to their small size and low weight, \glspl{imu} can be used to estimate the orientation for control of an unmanned helicopter.]{\includegraphics[height = 0.3\textwidth]{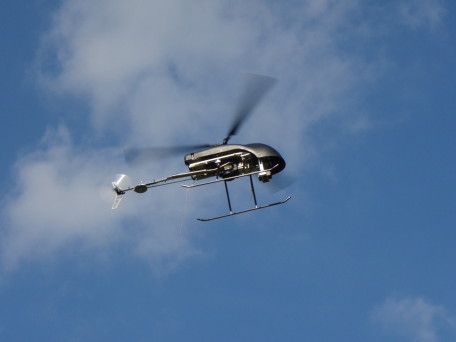}}
  \caption{Examples illustrating the use of a single \gls{imu} placed on a moving 
object to estimate its pose. Courtesy of Xsens Technologies.}
  \label{fig:intro-singleSensorApplications}
\end{figure}

There is a large and ever-growing number of application areas for inertial sensors, see \eg \cite{barbourS:2001,hol:2011,perlmutterR:2012,xsens-tutorial}. Generally speaking, inertial sensors can be used to provide information about the pose of any object that they are rigidly attached to. It is also possible to combine multiple inertial sensors to obtain information about the pose of separate connected objects. Hence, inertial sensors can be used to track human motion as illustrated in \Figureref{fig:intro-motionCaptureApplications}. This is often referred to as motion capture. The application areas are as diverse as robotics, biomechanical analysis and motion capture for the movie and gaming industries. In fact, the use of inertial sensors for pose estimation is now common practice in for instance robotics and human motion tracking, see \eg \cite{luingeV2005,harle:2013,raibertBNPT:2008}. A recent survey~\citep{adlerSWK:2015} shows that 28\% of the contributions to the IEEE International Conference on Indoor Positioning and Indoor Navigation (IPIN) make use of inertial sensors. Inertial sensors are also frequently used for pose estimation of cars, boats, trains and aerial vehicles, see \eg \cite{skogH:2009,chaoCC:2010}. Examples of this are shown in \Figureref{fig:intro-singleSensorApplications}. 

There exists a large amount of literature on the use of inertial sensors for position and orientation estimation. The reason for this is not only the large number of application areas. Important reasons are also that the estimation problems are nonlinear and that different parametrizations of the orientation need to be considered~\citep{grisettiKSB:2010,kurzGJH:2013}, each with its own specific properties. Interestingly, approximative and relatively simple position and orientation estimation algorithms work quite well in practice. However, careful modeling and a careful choice of algorithms do improve the accuracy of the estimates. 

In this tutorial we focus on the signal processing aspects of position and orientation estimation using inertial sensors, discussing different modeling choices and a number of important algorithms. These algorithms will provide the reader with a starting point to implement their own position and orientation estimation algorithms. 

\section{Using inertial sensors for position and orientation estimation}
\label{sec:intro-imusForPose}
\begin{figure}
	\centering
	\includegraphics[scale = 1]{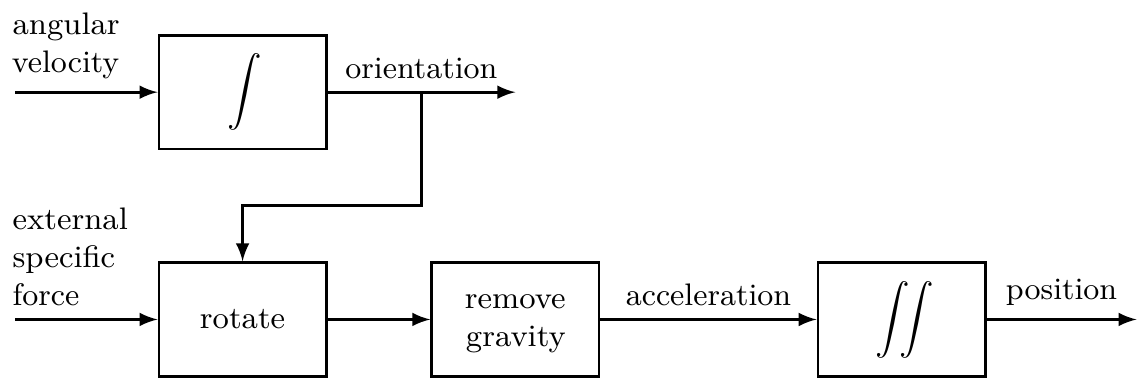}
		\caption{Schematic illustration of dead-reckoning, where the accelerometer measurements (external specific force) and the gyroscope measurements (angular velocity) are integrated to position and orientation.}
	\label{fig:intro-strapdown}
\end{figure}

As illustrated in \Sectionref{sec:intro-background}, inertial sensors are frequently used for navigation purposes where the position and the orientation of a device are of interest. Integration of the gyroscope measurements provides information about the orientation of the sensor. After subtraction of the earth's gravity, double integration of the accelerometer measurements provides information about the sensor's position. To be able to subtract the earth's gravity, the orientation of the sensor needs to be known. Hence, estimation of the sensor's position and orientation are inherently linked when it comes to inertial sensors. The process of integrating the measurements from inertial sensors to obtain position and orientation information, often called \emph{dead-reckoning}, is summarized in \Figureref{fig:intro-strapdown}.

If the initial pose would be known, and if perfect models for the inertial sensor measurements would exist, the process illustrated in \Figureref{fig:intro-strapdown} would lead to perfect pose estimates. In practice, however, the inertial measurements are noisy and biased as will be discussed in more detail in \Sectionref{sec:sensors-errors}. Because of this, the integration steps from angular velocity to rotation and from acceleration to position introduce \textit{integration drift}. This is illustrated in \Exampleref{ex:intro-integrationDrift}. 

\begin{myexample}{Integration drift}%
\label{ex:intro-integrationDrift}%
Let us first focus on the general case of measuring a quantity that is constant and equal to zero. The integrated and double integrated signals are therefore also equal to zero. However, let us now assume that we measure this quantity using a non-perfect sensor. In case our measurements are corrupted by a constant bias, integration of these measurements will lead to a signal which grows linearly with time. Double integration leads to a signal that instead grows quadratically with time. If the sensor instead measures a zero-mean white noise signal, the expected value of the integrated measurements would be zero, but the variance would grow with time. This is illustrated in \Figureref{fig:intro-integrationDrift} for the integration of a signal $y_t = e_t$ with $e_t \sim \mathcal{N}(0,1)$. Hence, integration drift is both due to integration of a constant bias and due to integration of noise. 

\begin{figure}
	\centering
	\includegraphics[scale = 1]{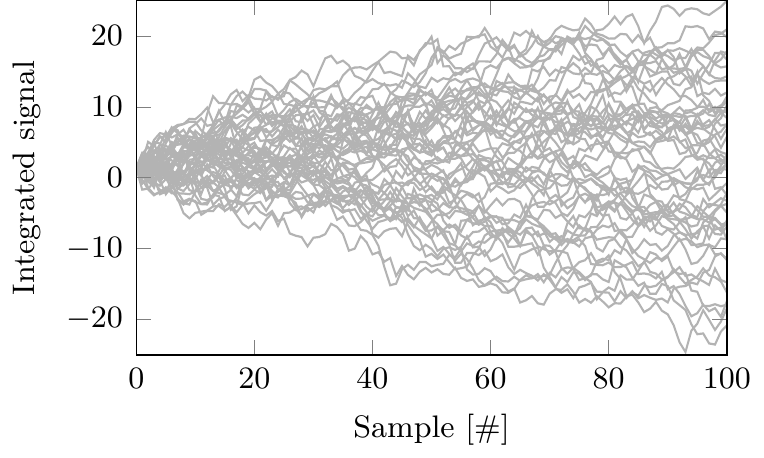}
    	\caption{Integration of a white noise signal $y_t \sim \mathcal{N}(0,1)$ for $50$ noise realizations.}
	\label{fig:intro-integrationDrift}
\end{figure}

To illustrate integration drift using experimental data, a stationary data set is collected with a Sony Xperia Z5 Compact smartphone using the app described in \cite{hendebyGWG:2017}. The smartphone contains accelerometers and gyroscopes produced by Invensense~\citep{invensense-tutorial}. We integrate the inertial measurements to obtain position and orientation estimates. Since the smartphone is kept stationary during the data collection, we expect the position and orientation to remain the same. However, the orientation estimates drift a few degrees over $10$ seconds as shown in \Figureref{fig:intro-oriDrift}. Note that the integration drift is not the same for all axes. This is mainly due to a different sensor bias in the different axes. This will be studied further in \Exampleref{ex:sensors-inertialMeasurements}, where the same data set is used to study the sensor characteristics. As shown in \Figureref{fig:intro-posDrift}, the position drifts several meters over $10 \second$. The reason for this is two-fold. First, the accelerometer measurements need to be integrated twice. Second, the orientation estimates need to be used to subtract the gravity and any errors in this will result in \emph{leakage of gravity} into the other components. 
\begin{figure}[t]
	\centering
	\subfigure[Integrated orientation for the position in $x$- (blue), $y$- (green) and $z$-direction (red).]{
	\includegraphics[scale = 1]{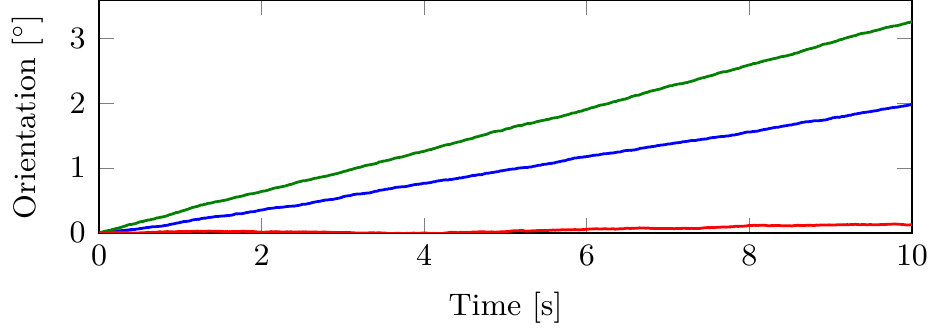}
	\label{fig:intro-oriDrift}}
	\subfigure[Integrated position for rotation around the $x$-axis (blue), the $y$-axis (green) and the $z$-axis (red).]{
	\includegraphics[scale = 1]{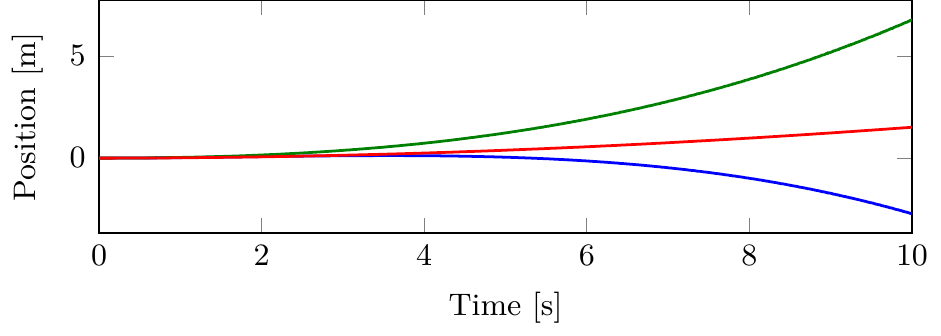}
	\label{fig:intro-posDrift}}
	\caption{Position and orientation estimates based on dead-reckoning of the inertial sensors only. The data is collected with a Sony Xperia Z5 Compact smartphone that is lying stationary on a table.}
\end{figure}
\end{myexample}

From the example above, it can be concluded that errors in the measurements have a large impact on the quality of the estimated position and orientation using inertial sensors only. This is particularly the case for position, which relies both on double integration of the acceleration and on accurate orientation estimates to subtract the earth's gravity. Because of this, inertial sensors need to be supplemented with other sensors and other models to obtain accurate position and orientation estimates. 

Inertial sensors provide pose estimates at high sampling rates which are accurate on a short time scale but drift over longer time scales. They are therefore very suitable for being combined with sensors with a lower sampling rate but with information that does not drift over time. For pose estimation, inertial sensors are often combined with measurements from for instance a \acrfull{gnss} \citep{kaplanH:1996,tittertonW:1997,hol:2011}, an \gls{uwb} system \citep{kokHS:2015,sczysloSGK:2008,pittetRMK:2008,corralesCT:2008,deAngelisNSHC:2010} or cameras \citep{corkeLD:2007,holSLSG:2007,liM:2013,martinelli:2012}. For orientation estimation, they are often used in combination with magnetometers, which measure the direction of the magnetic field \citep{sabatini:2006,roetenbergLBV:2005}. 

This tutorial aims at giving an introduction on how to use inertial sensors for position and orientation estimation, but also on how to combine them with additional information. These additional sensors are not the focus of this paper but simple models will be used for magnetometers and sensors providing position information to illustrate the combined use of these sensors.

\section{Tutorial content and its outline}
\label{sec:intro-outline}
To obtain accurate position and orientation estimates using inertial sensors in combination with additional measurements and models, a number of important things need to be considered. First, the quantities measured by the inertial sensors need to be accurately described and the sources of error need to be characterized. This is the topic of \Chapterref{cha:sensors}. Note that throughout the tutorial, we will focus on \gls{mems} inertial sensors and consider both data from standalone \glspl{imu} and from smartphones. This implies that we do not focus on for instance mechanical or optical gyroscopes and on mechanical or solid-state accelerometers~\citep{tittertonW:1997}. These sensors may have characteristics that are quite different from the \gls{mems} inertial sensors considered here. 

Based on the analysis of the sensors in \Chapterref{cha:sensors} and on additional analysis of the application at hand, models can be constructed. This is the topic of \Chapterref{cha:models}, where we will also discuss different parametrizations of orientation. This will highlight the challenges in parametrizing and estimating orientations and show that the orientation estimation problem is inherently nonlinear. Furthermore, we will present two models that can be used for position and orientation estimation. The first is a model for pose estimation using inertial measurements in combination with position measurements. The second is a model for orientation estimation, using inertial and magnetometer measurements. 

In \Chapterref{cha:orientationEstimation}, different algorithms for position and orientation estimation will be introduced. The general structure of the algorithms will be discussed, after which explicit algorithms for orientation estimation using inertial and magnetometer measurements are given. We will also discuss how the algorithms can be extended to pose estimation when position measurements are available. Some general characteristics of the two estimation problems will be given and the quality of the estimates from the different algorithms will be analyzed. Which algorithm is most suitable for which application depends strongly on the computational power that is available, the accuracy that is required and the characteristics of the problem at hand. 

In \Chapterref{cha:orientationEstimation}, we assume that the sensors are properly calibrated. However, calibration of the sensors is important to for instance estimate the inertial sensor biases. Furthermore, calibration is specifically of concern when combining inertial data with other sensors. In these cases, it is important that the inertial sensor axes and the axes of the additional sensors are aligned. Sensor calibration is the topic of \Chapterref{cha:calibration}. As an illustrative example, we will consider the estimation of an unknown gyroscope bias. 

Our focus in this tutorial is to present models and algorithms that can be used for position and orientation estimation using inertial measurements. Because of this, we assume fairly simple models for the additional information --- the magnetometer and position measurements. In \Chapterref{cha:applications}, we will discuss how the algorithms from \Chapterref{cha:orientationEstimation} can be used in more complex settings. For example, we will consider the cases of more complex position information in the form of images from a camera, the presence of non-Gaussian measurement noise and the availability of additional information that can be exploited. The information provided by the inertial sensors remains one of the main building blocks of algorithms that can be used for these cases. Adaptations of the algorithms presented in \Chapterref{cha:orientationEstimation} can therefore be used to also solve these more complex scenarios. We will end this tutorial with some concluding remarks in \Chapterref{cha:conclusions}.

\chapter{Inertial Sensors}
\label{cha:sensors} 
To combine inertial measurements with additional sensors and models for position and orientation estimation, it is important to accurately describe the quantities measured by the inertial sensors as well as to characterize the typical sensor errors. This will be the topic of this chapter. It will serve as a basis for the probabilistic models discussed in \Chapterref{cha:models}.

As discussed in \Chapterref{cha:introduction}, accelerometers and gyroscopes measure the specific force and the angular velocity, respectively. In~\Sectionref{sec:sensors-angVel} and~\Sectionref{sec:sensors-specForce}, we will discuss these quantities in more detail. To enable a discussion about this, in \Sectionref{sec:sensors-coordFrames} a number of coordinate frames and the transformations between them will be discussed. We assume that we have 3D accelerometers and 3D gyroscopes, \ie that the sensors have three sensitive axes along which these physical quantities are measured. They are measured in terms of an output voltage which is converted to a physical measurement based on calibration values obtained in the factory. Even though the sensors are typically calibrated in the factory, (possibly time-varying) errors can still remain. In \Sectionref{sec:sensors-errors}, the most commonly occurring sensor errors are discussed. 

\section{Coordinate frames}
\label{sec:sensors-coordFrames}
\begin{figure}
	\centering
    	\includegraphics[scale = 1]{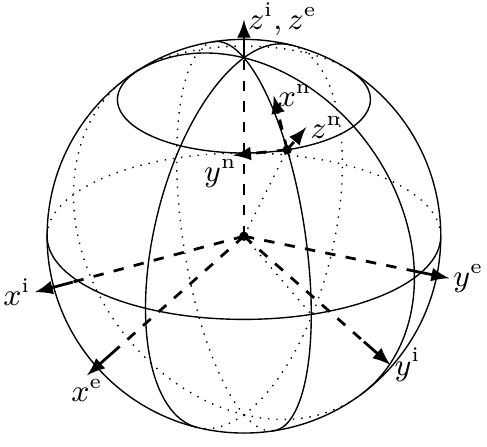}
    	\caption{An illustration of three of the coordinate frames discussed in \Sectionref{sec:sensors-coordFrames}: the $n$-frame at a certain location on the earth, the $e$-frame rotating with the earth
      		and the $i$-frame.}
	\label{fig:sensors-coordinateFrames}
\end{figure}

In order to discuss the quantities measured by the accelerometer and gyroscope in more detail, a number of
coordinate frames need to be introduced:
\begin{description}
\item[The body frame $\boldsymbol{b}$] is the coordinate frame of the moving
  \gls{imu}. Its origin is located in the center of the accelerometer
  triad and it is aligned to the casing. All the inertial measurements
  are resolved in this frame.
\item[The navigation frame $\boldsymbol{n}$] is a local geographic frame in which we
  want to navigate. In other words, we are interested in the position and
  orientation of the $b$-frame with respect to this frame. For most
  applications it is defined stationary with respect to the earth. However,
  in cases when the sensor is expected to move over large distances, it is customary to
  move and rotate the $n$-frame along the surface of the
  earth. The first definition is used throughout this tutorial, unless
  mentioned explicitly.
\item[The inertial frame $\boldsymbol{i}$] is a stationary frame. The
  \gls{imu} measures linear acceleration and angular velocity
  with respect to this frame. Its origin is located at the center of the
  earth and its axes are aligned with respect to the stars.
\item[The earth frame $\boldsymbol{e}$] coincides with the $i$-frame, but rotates
  with the earth. That is, it has its origin at the center of the
  earth and axes which are fixed with respect to the earth.
\end{description}
The $n$, $i$ and $e$ coordinate frames are illustrated in \Figureref{fig:sensors-coordinateFrames}. We use a superscript to indicate in which coordinate frame a vector is expressed. Vectors can be rotated from one coordinate frame to another using a rotation matrix. We use a double superscript to indicate from which coordinate frame to which coordinate frame the rotation is defined. An illustration is given in \Exampleref{ex:sensors-rotationVectors}. 

\begin{myexample}{Rotation of vectors to different coordinate frames}%
\label{ex:sensors-rotationVectors}%
Consider a vector $x$ expressed in the body frame $b$. We denote this vector by $x^\text{b}$. The rotation matrix $R^\text{nb}$ rotates a vector from the body frame $b$ to the navigation frame~$n$. Conversely, the rotation from navigation frame $n$ to body frame $b$ is denoted $R^\text{bn} = ( R^{\text{nb}} )^\Transp$. Hence, the vector $x$ expressed in the body frame ($x^\text{b}$) and expressed in the navigation frame ($x^\text{n}$) are related according to
\begin{align}
x^\text{n} = R^{\text{nb}} x^\text{b}, \qquad x^\text{b} = ( R^{\text{nb}} )^\Transp x^\text{n} = R^{\text{bn}} x^\text{n}.
\end{align}%
\vspace*{-\baselineskip}
\end{myexample}

\section{Angular velocity}
\label{sec:sensors-angVel}
The gyroscope measures the angular velocity of the body frame with respect to the inertial frame, expressed in the body frame~\citep{tittertonW:1997}, denoted by $\omega_{\text{ib}}^\text{b}$. This angular velocity can be expressed as
\begin{align}
\label{eq:sen-imu-gyr-n}
\omega_{\text{ib}}^\text{b}
= R^{\text{bn}} \left(
\omega_{\text{ie}}^\text{n} +
\omega_{\text{en}}^\text{n} \right)
+ \omega_{\text{nb}}^\text{b},
\end{align}
where $R^{\text{bn}}$ is the rotation matrix from the navigation frame to the body frame. The \emph{earth rate}, \ie the angular velocity of the earth frame with respect to the inertial frame is denoted by $\omega_{\text{ie}}$. The earth rotates around its own $z$-axis in $23.9345$ hours with respect to the stars~\citep{nasa:2016}. Hence, the earth rate is approximately $7.29 \cdot 10^{-5}~\radianpersecond$.

In case the navigation frame is not defined stationary with respect to the earth, the angular velocity $\omega_{\text{en}}$, \ie the \emph{transport rate} is non-zero. The angular velocity required for navigation purposes --- in which we are interested when determining the orientation of the body frame with respect to the navigation frame --- is denoted by $\omega_{\text{nb}}$.

\section{Specific force}
\label{sec:sensors-specForce}
The accelerometer measures the specific force $f$ in the body frame $b$~\citep{tittertonW:1997}. This can be expressed as
\begin{align}
\label{eq:sen-imu-acc-i}
f^\text{b} = R^{\text{bn}} 
( a_{\text{ii}}^\text{n} - g^\text{n} ),
\end{align}
where $g$ denotes the gravity vector and $a_{\text{ii}}^\text{n}$ denotes the linear acceleration of the sensor expressed in the navigation frame, which is
\begin{align}
a_{\text{ii}}^\text{n} = R^{\text{ne}} R^{\text{ei}} a_{\text{ii}}^\text{i}.
\end{align}
The subscripts on the linear acceleration $a$ are used to indicate in which frame the differentiation is performed. For navigation purposes, we are interested in the position of the sensor in the navigation frame $p^\text{n}$ and its derivatives as performed in the navigation frame 
\begin{align}
\label{eq:sensors-defva}
\left. \tfrac{\tdiff}{\tdiff t} p^\text{n} \right|_\text{n} = v^\text{n}_\text{n}, \qquad \left. \tfrac{\tdiff}{\tdiff t} v^\text{n} \right|_\text{n} = a^\text{n}_\text{nn}.
\end{align}
A relation between $a_{\text{ii}}$ and $a_{\text{nn}}$ can be derived by using the relation between two rotating coordinate frames. Given a vector $x$ in a coordinate frame $u$,
\begin{align}
\label{eq:sensors-rotatingCF}
\left. \tfrac{\tdiff}{\tdiff t} x^\text{u} \right|_\text{u} = \left. \tfrac{\tdiff}{\tdiff t} R^\text{uv} x^\text{v} \right|_{\text{u}} = R^\text{uv} \left. \tfrac{\tdiff}{\tdiff t} x^\text{v} \right|_{\text{v}} + \omega_{\text{uv}}^\text{u} \times x^\text{u},
\end{align}
where $\omega_{\text{uv}}^\text{u}$ is the angular velocity of the $v$-frame with respect to the $u$-frame, expressed in the $u$-frame. For a derivation of this relation in the context of inertial navigation, see \cite{hol:2011,tittertonW:1997}. For a general introduction, see any textbook on dynamics, \eg \cite{marionT:1995,meriamK:1998}. 

Using the fact that
\begin{align}
p^\text{i} = R^\text{ie} p^\text{e},
\end{align} 
the velocity $v_\text{i}$ and acceleration $a_\text{ii}$ can be expressed as
\begin{subequations}
\label{eq:sensors-pvi-pve}
\begin{align}
v_\text{i}^\text{i} &= \left. \tfrac{\tdiff}{\tdiff t} p^\text{i} \right|_\text{i} = 
\left. \tfrac{\tdiff}{\tdiff t} R^\text{ie} p^\text{e} \right|_\text{i} = 
R^\text{ie} \left. \tfrac{\tdiff}{\tdiff t} p^\text{e} \right|_\text{e} + \omega_{\text{ie}}^\text{i} \times p^\text{i} 
= v^\text{i}_\text{e} + \omega_{\text{ie}}^\text{i} \times p^\text{i}, \\
a_\text{ii}^\text{i} &= \left. \tfrac{\tdiff}{\tdiff t} v^\text{i}_\text{i} \right|_\text{i} = 
\left. \tfrac{\tdiff}{\tdiff t} v^\text{i}_\text{e} \right|_\text{i} + \left. \tfrac{\tdiff}{\tdiff t} \omega_{\text{ie}}^\text{i} \times p^\text{i} \right|_\text{i} \nonumber \\
&= a^\text{i}_\text{ee} + 2 \omega_{\text{ie}}^\text{i} \times v^\text{i}_\text{e} + \omega_{\text{ie}}^\text{i} \times \omega_{\text{ie}}^\text{i} \times p^\text{i},
\end{align} 
\end{subequations}
where we have made use of~\eqref{eq:sensors-defva},~\eqref{eq:sensors-rotatingCF}, and the fact that the angular velocity of the earth is constant, \ie $\tfrac{\tdiff}{\tdiff t} \omega_{\text{ie}}^\text{i} = 0$. Using the relation between the earth frame and the navigation frame 
\begin{align}
p^\text{e} = R^\text{en} p^\text{n} + n_\text{ne}^\text{e},
\end{align} 
where $n_\text{ne}$ is the distance from the origin of the earth coordinate frame to the origin of the navigation coordinate frame, expressions similar to~\eqref{eq:sensors-pvi-pve} can be derived. Note that in general it can not be assumed that $\tfrac{\tdiff}{\tdiff t} \omega_{\text{en}} = 0$. Inserting the obtained expressions into~\eqref{eq:sensors-pvi-pve}, it is possible to derive the relation between $a_\text{ii}$ and $a_\text{nn}$. Instead of deriving these relations, we will assume that the navigation frame is fixed to the earth frame, and hence $R^\text{en}$ and $n_\text{ne}^\text{e}$ are constant and
\begin{subequations}
\label{eq:sensors-pve-pvn}
\begin{align}
v_\text{e}^\text{e} &=
\left. \tfrac{\tdiff}{\tdiff t} p^\text{e} \right|_\text{e} = 
\left. \tfrac{\tdiff}{\tdiff t} R^\text{en} p^\text{n} \right|_\text{e} = 
R^\text{en} \left. \tfrac{\tdiff}{\tdiff t} p^\text{n} \right|_\text{n} = 
v^\text{e}_\text{n}, \\
a_\text{ee}^\text{e} &=
\left. \tfrac{\tdiff}{\tdiff t} v^\text{e}_\text{e} \right|_\text{e} = 
\left. \tfrac{\tdiff}{\tdiff t} v^\text{e}_\text{n} \right|_\text{n} = 
a^\text{e}_\text{nn} .
\end{align} 
\end{subequations}
This is a reasonable assumption as long as the sensor does not travel over significant distances as compared to the size of the earth and it will be one of the model assumptions that we will use in this tutorial. More on the modeling choices will be discussed in \Chapterref{cha:models}. 

Inserting~\eqref{eq:sensors-pve-pvn} into~\eqref{eq:sensors-pvi-pve} and rotating the result, it is possible to express $a_\text{ii}^\text{n}$ in terms of $a_\text{nn}^\text{n}$ as
\begin{align}
\label{eq:sensors-aii-ann} 
a_\text{ii}^\text{n} = 
a_\text{nn}^\text{n} + 
2 \omega_\text{ie}^\text{n} \times v^\text{n}_\text{n} + 
\omega_\text{ie}^\text{n} \times \omega_\text{ie}^\text{n}
\times p^\text{n},
\end{align}
where $a_\text{nn}$ is the acceleration required for navigation
purposes. The term $\omega_\text{ie}^\text{n} \times \omega_\text{ie}^\text{n}
\times p^\text{n}$ is known as the \emph{centrifugal acceleration} and $2 \omega_\text{ie}^\text{n} \times v^\text{n}_\text{n}$ is known as the \emph{Coriolis acceleration}. The centrifugal acceleration is typically absorbed in the (local) gravity vector. In \Exampleref{ex:sensors-magCentrCor}, we illustrate the magnitude of both the centrifugal and the Coriolis acceleration.

\begin{myexample}{Magnitude of centrifugal and Coriolis acceleration}%
\label{ex:sensors-magCentrCor}%
The centrifugal acceleration depends on the location on the earth. It is possible to get a feeling for its magnitude by considering the property of the cross product stating that
\begin{align}
\| \omega_\text{ie}^\text{n} \times \omega_\text{ie}^\text{n} \times p^\text{n} \|_2 \leq 
\| \omega_\text{ie}^\text{n} \|_2 \| \omega_\text{ie}^\text{n} \|_2 \| p^\text{n} \|_2.
\end{align}
Since the magnitude of $\omega_\text{ie}$ is approximately $7.29 \cdot 10^{-5}~\radian\per\second$ and the average radius of the earth is $6371~\kilo\meter$~\citep{nasa:2016}, the magnitude of the centrifugal acceleration is less than or equal to $3.39 \cdot 10^{-2}~\metrepersquaresecond$. 

The Coriolis acceleration depends on the speed of the sensor. Let us consider a person walking at a speed of $5~\kilo\meter\per\hour$. In that case the magnitude of the Coriolis acceleration is approximately $2.03 \cdot 10^{-4}~\metrepersquaresecond$. For a car traveling at $120~\kilo\meter\per\hour$, the magnitude of the Coriolis acceleration is instead $4.86 \cdot 10^{-3}~\metrepersquaresecond$. 
\end{myexample}

\section{Sensor errors}
\label{sec:sensors-errors}
As discussed in \Sectionref{sec:sensors-angVel} and~\Sectionref{sec:sensors-specForce}, the gyroscope measures the angular velocity $\omega^\text{b}_\text{ib}$ and the accelerometer measures the specific force $f^\text{b}$. However, as already briefly mentioned in \Sectionref{sec:intro-imusForPose}, there are several reasons for why this is not exactly the case. Two of these reasons are a slowly time-varying sensor bias and the presence of measurement noise. The sensor errors in the inertial measurements are illustrated in \Exampleref{ex:sensors-inertialMeasurements} using experimental data. 

\newpage 

\begin{myexample}{Inertial sensor measurements and their errors}%
\label{ex:sensors-inertialMeasurements}%
In Figures~\ref{fig:sensors-accgyrMeas}--\ref{fig:sensors-accMeasNoiseHist}, gyroscope and accelerometer measurements are displayed for around $10$ seconds of stationary data collected with a Sony Xperia Z5 Compact smartphone. Since the smartphone is stationary, the gyroscope is expected to only measure the earth's angular velocity. However, as can be seen in \Figureref{fig:sensors-gyrMeas}, the gyroscope measurements are corrupted by noise. As shown in \Figureref{fig:sensors-gyrMeasNoiseHist}, this noise can be seen to be quite Gaussian. Furthermore, the measurements can be seen to be biased. 

During the stationary period, we would expect the accelerometer to measure the gravity, the centrifugal acceleration and the Coriolis acceleration. Note that again the measurements are corrupted by noise, which can be seen to be quite Gaussian in \Figureref{fig:sensors-accMeasNoiseHist}. The $x$- and $y$-components of the accelerometer measurements are not zero-mean. This can be due to the fact that the table on which the smartphone lies is not completely flat, implying that part of the gravity vector is measured in these components. It can also reflect a sensor bias. The $z$-component is actually larger than expected which indicates the presence of an accelerometer bias at least in this axis. 

Note that from the above discussion it can be concluded that it is more straightforward to determine the gyroscope bias than it is to determine the accelerometer bias. To be able to estimate the gyroscope bias, it is sufficient to leave the sensor stationary. For the accelerometer, however, it is in that case difficult to distinguish between a bias and a table that is not completely flat. For more information about accelerometer or general inertial sensor calibration, see~\cite{tedaldiPM:2014,olssonKHS:2016,panahandehSJ:2010}. 

The gyroscope in the smartphone is automatically recalibrated during stationary time periods. The measurements shown in \Figureref{fig:sensors-gyrMeas} have not been corrected for this (so-called uncalibrated or raw data). The reason why the gyroscope is calibrated during stationary periods, is because its bias is slowly time-varying. As an example, let us consider a data set of $55$ minutes. The gyroscope bias during the first minute of the data set was $\begin{pmatrix} 35.67 & 56.22 & 0.30 \end{pmatrix}^\Transp \cdot 10^{-4}~\radianpersecond$, while the gyroscope bias during the last minute of the data set was $\begin{pmatrix} 37.01 & 53.17 & -1.57 \end{pmatrix}^\Transp \cdot 10^{-4}~\radianpersecond$.

\begin{figure}
	\centering
	\subfigure[Gyroscope measurements $y_{\omega,t}$ which we expect to consist only of the earth's angular velocity.]{
	\includegraphics[scale = 1]{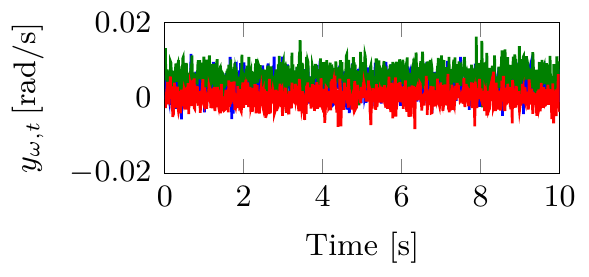}
	\label{fig:sensors-gyrMeas}} 
	\tikzsetnextfilename{sensors-accMeas}
	\subfigure[Accelerometer measurements $y_{\text{a},t}$ which we expect to consist of the gravity vector, the centrifugal acceleration and the Coriolis acceleration.]{
	\includegraphics[scale = 1]{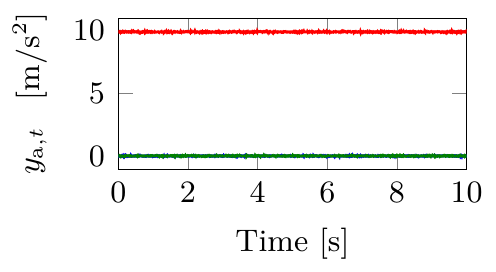}
	\label{fig:sensors-accMeas}} 
    	\caption{Inertial measurements for $10$ seconds of stationary data. As can be seen, the measurements are corrupted by noise and have a bias.}
	\label{fig:sensors-accgyrMeas}
\end{figure}

\begin{figure}
	\centering
    	\includegraphics[scale = 1]{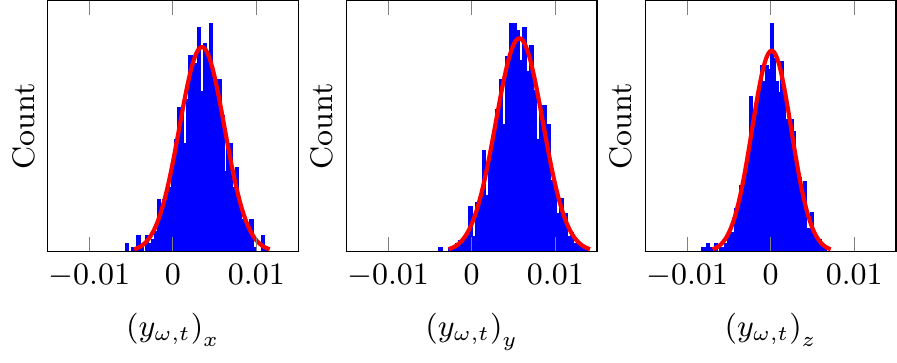}
    	\caption{Histogram (blue) of the gyroscope measurements for $10$ seconds of data from a stationary sensor and a Gaussian fit (red) to the data. As can be seen, the measurement noise looks quite Gaussian.}
	\label{fig:sensors-gyrMeasNoiseHist}
\end{figure}

\begin{figure}
	\centering
    	\includegraphics[scale = 1]{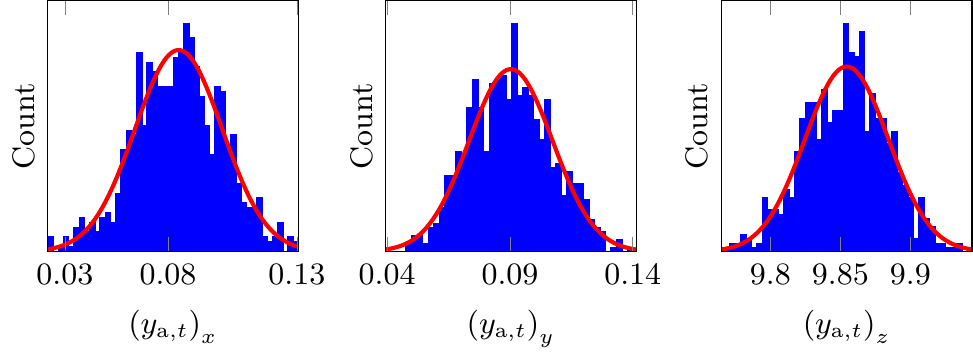}
    	\caption{Histogram (blue) of the accelerometer measurements for $10$ seconds of data from a stationary sensor and a Gaussian fit (red) to the data. As can be seen, the measurement noise looks quite Gaussian. Note the different scales on the horizontal axis.}
	\label{fig:sensors-accMeasNoiseHist}
\end{figure}
\end{myexample}

\begin{figure}
	\centering
    	\includegraphics[scale = 1]{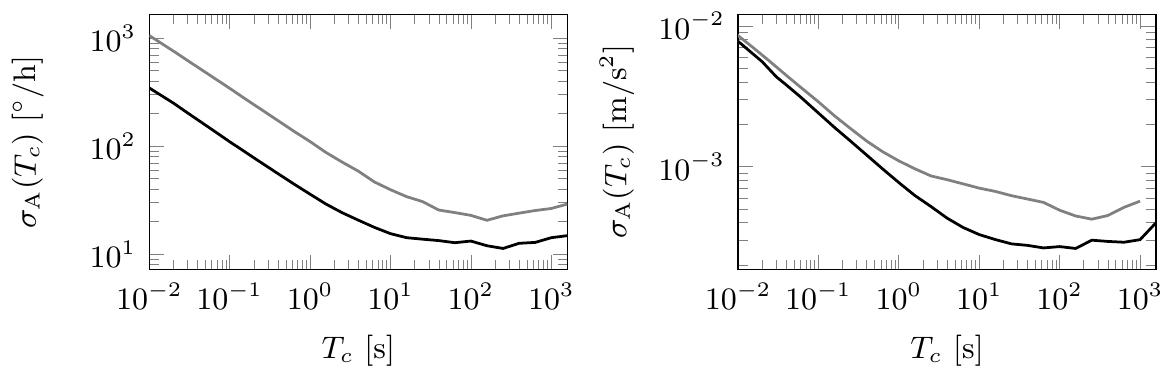}
    	\caption{Left: Allan deviation for two gyroscopes. Right: Allan deviation for two accelerometers. Reproduced with permission from~\cite{vydhyanathanBLS:2015}.}
	\label{fig:sensors-allanVariance}
\end{figure}

The performance of \glspl{imu} is often specified in terms of their \emph{Allan variance} \citep{ieeeStd1559:2009,elsheimy:2008,allan:1966}. The Allan variance gives information about the sensor errors for stationary conditions, \ie in a stable climate without exciting the system. It studies the effect of averaging measurements for different \emph{cluster times} $T_c$. Typically, the Allan \emph{standard deviation} $\sigma_\text{A}(T_c)$ is plotted against the cluster time $T_c$ as illustrated in \Figureref{fig:sensors-allanVariance}. This figure shows the characteristic behavior of the Allan variance for inertial sensors. To study it more in detail, we will discuss two components of the Allan variance that are typically of concern for inertial sensors: the white noise and the bias instability. 

Assume, as in \Exampleref{ex:intro-integrationDrift}, that we have a white noise signal with standard deviation $\sigma$. A longer averaging time would for this signal lead to values closer to zero. The contribution to the Allan standard deviation from the white noise component is given by $\sigma_\text{A}(T_c) = \tfrac{\sigma}{\sqrt{n}}$ where $n$ is the number of samples averaged over. This corresponds to a line with slope $-1/2$ in a $\log$--$\log$ plot. For instance in the Allan deviation for the gyroscope in \Figureref{fig:sensors-allanVariance}, the lines can be seen to have a slope of $-1/2$ until around $10-20~\second$, which indicates that the white noise is the dominating source of error for these short integration times. 

A constant bias does not have any effect on the Allan variance diagram. However, in case the bias changes, longer averaging times will no longer be beneficial. Hence, the Allan variance diagrams in \Figureref{fig:sensors-allanVariance} show a deviation from the slope $-1/2$ for longer averaging times. 

The Allan variance is a useful tool to study and compare the noise characteristics of inertial sensors. However, it only considers stationary conditions. In dynamic conditions, a large number of other error sources potentially come into play, see \eg \cite{tittertonW:1997,woodman:2007}. These are for instance related to the fact that the sensors sample at discrete times. Hence, to capture high-frequency signals, high sampling frequencies are desired~\citep{savage:1998a,savage:1998b}. Furthermore, large dynamics can lead to erroneous or saturated measurements. Other errors that are not included are for instance changes in the sensitivity of the axes due to changes in temperature. We should therefore never just rely on the Allan variance when deciding which sensor to use in a particular application.

\chapter{Probabilistic Models}
\label{cha:models} 
\begin{figure}
  	\centering
    	\includegraphics[scale = 1]{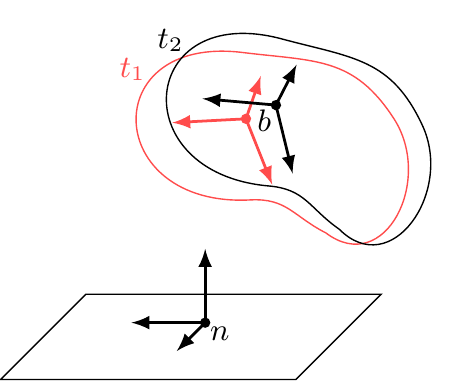}
    	\caption{An illustration of the pose estimation problem. We want to express the position and orientation of the moving body frame $b$ at times $t_1$ and $t_2$ with respect to the navigation frame $n$.}
    	\label{fig:models-poseEstimation}
\end{figure}

Pose estimation is about estimating the position and orientation of the body frame~$b$ in the navigation frame $n$. This problem is illustrated in \Figureref{fig:models-poseEstimation}, where the position and orientation of the body changes from time $t_1$ to time $t_2$. In this chapter, we will introduce the concept of probabilistic models and discuss different modeling choices when using inertial sensors for pose estimation. 

The subject of probabilistic modeling is introduced in \Sectionref{sec:models-probModeling}. Most complexity in pose estimation lies in the nonlinear nature of the orientation and the fact that orientation can be parametrized in different ways. How to parametrize the orientation is therefore a crucial modeling choice in any pose estimation algorithm. Because of this, we will discuss different parametrizations for the orientation in \Sectionref{sec:models-paramOri} and in \Sectionref{sec:models-probModelingOri} we will discuss how these different parametrizations can be used in probabilistic modeling. 

Our probabilistic models consist of three main components. First, in \Sectionref{sec:models-measModels}, we introduce models describing the knowledge about the pose that can be inferred from the measurements. Second, in \Sectionref{sec:models-stateDynamics}, we model how the sensor pose changes over time. Finally, in \Sectionref{sec:models-prior}, models of the initial pose are introduced. 

The chapter will conclude with a discussion on the resulting probabilistic models in \Sectionref{sec:models-resultingProbModel}. The models that will be used in the position and orientation estimation algorithms in \Chapterref{cha:orientationEstimation} will be introduced in this section.

\section{Introduction}
\label{sec:models-probModeling}
Probabilistic models constitute the foundation of the estimation algorithms in \Chapterref{cha:orientationEstimation}. In this section we will introduce the concept of probabilistic modeling and the notation that is used in building our models. Models are used to describe the information about the \emph{dynamics} and the available \emph{measurements}. These models are subsequently used in combination with the measurements to \emph{infer} some knowledge. The knowledge that we are interested in is the pose of the sensor and we use information about the \emph{sensor dynamics} and the available measurements (amongst others, inertial measurements). A simplified case where probabilistic modeling is used to estimate the position of a sensor is given in \Exampleref{ex:models-probModeling}.

\begin{myexample}{Probabilistic modeling}%
\label{ex:models-probModeling}%
Let us estimate the 2D position $p_t$ of a sensor at time $t$ from two position measurements
\begin{align*}
y_t^1 = \begin{pmatrix} 0 & 0 \end{pmatrix}^\Transp, \qquad y_t^2 = \begin{pmatrix} 2 & 0 \end{pmatrix}^\Transp.
\end{align*}
A straightforward suggestion for an estimate of the position would be $\hat{p}_t = \begin{pmatrix} 1 & 0 \end{pmatrix}^\Transp$. Let us now assume that we know the accuracy of the sensors and represent this in terms of the following probabilistic models 
\begin{align*}
y_t^1 &= p_t + e_t^1, &e_t^1 &\sim \mathcal{N}( 0, 0.25 \, \mathcal{I}_2 ), \\
y_t^2 &= p_t + e_t^2, &e_t^2 &\sim \mathcal{N}( 0,\mathcal{I}_2 ),
\end{align*}
where $\mathcal{I}_2$ denotes a $2 \times 2$ identity matrix. A reasonable position estimate would instead be
\begin{align*}
p_t \sim \mathcal{N} \left( \begin{pmatrix} 0.4 \\ 0 \end{pmatrix}, 0.2 \, \mathcal{I}_2 \right) .
\end{align*}
We will not go into details about how this estimate is derived. Instead, we would like to point out two differences between this position estimate and our initial suggestion $\hat{p}_t$. First, based on the knowledge of the accuracy of the sensors, it is sensible to trust the measurement from the first sensor more than the measurement from the second sensor. Our improved position estimate is therefore closer to the measurement of the first sensor than our initial suggestion $\hat{p}_t$. Furthermore, based on the accuracy of the sensors, it is possible to derive the accuracy of our estimate. 

Now consider the case where we are also interested in estimating the position $p_{t+1}$. Knowledge that the sensor is worn by a human or placed in a car, would give us information about how far the sensor can travel from time $t$ to time $t+1$. If the sensor would be placed in for instance a train, the motion would even be constrained to be along the tracks. Incorporating this information about the dynamics of the sensor will improve the estimate of $p_{t+1}$.
\end{myexample}

We split the knowledge that we want to infer into the unknown \emph{time-varying states} $x_t$ for $t = 1, \hdots, N$, or equivalently $x_{1:N}$, and the unknown \emph{constant parameters} $\theta$. We denote the measurements by $y_k$ for $k = 1, \hdots, K$. The times $k$ at which these measurements are obtained do not necessarily correspond with the times $t$ at which the states are defined. It is also not necessary for all sensors to sample at the same frequency. As discussed in \Sectionref{sec:sensors-errors}, the inertial sensors are typically sampled at fairly high rates to capture high-frequency dynamics. In stand-alone, wired \glspl{imu}, all sensors typically have the same, constant sampling frequency. Specifically in the case of wireless sensors and smartphones, however, the sampling frequencies can vary both over sensors and over time. In the remainder, we assume that the times $t$ at which the states are defined coincide with the times $k$ at which the gyroscopes sample. Hence, we denote the gyroscope measurements $y_{\omega,t}$ with $t = 1, \hdots N$. For notational convenience, we will also use the subscript~$t$ for the measurements from other sensors. Note that these are not required to actually sample at each time $t$ for $t = 1, \hdots N$. For instance, magnetometers in smartphones often sample either at equal or at half the sampling frequencies of the inertial sensors, while position aiding sensors like for instance \gls{gnss} or \gls{uwb} typically sample at much lower sampling frequencies. 

Our aim is now to infer information about the states $x_{1:N}$ and the parameters $\theta$ using the measurements $y_{1:N}$ and the probabilistic models. This can be expressed in terms of a \emph{conditional probability distribution}
\begin{align}
\label{eq:models-smoothing}
p(x_{1:N}, \theta \mid y_{1:N} ),
\end{align}
where $p ( a \mid b )$ denotes the conditional probability of $a$ given $b$. In the pose estimation problem, we are interested in obtaining \emph{point estimates} which we denote $\hat{x}_{1:N}$ and $\hat{\theta}$. It is typically also highly relevant to know how \emph{certain} we are about these estimates. This is often expressed in terms of a \emph{covariance}. When the distribution~\eqref{eq:models-smoothing} is Gaussian, the distribution is completely described in terms of its mean 
and covariance. 

In~\eqref{eq:models-smoothing} we assume that all measurements $y_{1:N}$ are used to obtain the posterior distribution of $x_{1:N}$ and $\theta$. This is referred to as \emph{smoothing}. Although it makes sense to use all available information to obtain the best estimates, a downside of smoothing is that we need to wait until all measurements are collected before the pose can be computed. Because of this, in many applications, we are also interested in \emph{filtering}. In filtering we estimate $x_t$ using all measurements up to and including time $t$. One way of dealing with constant parameters in filtering is to treat them as slowly time-varying. In this case, they can be considered to be included in the time-varying states $x_t$. The filtering problem can be expressed in terms of the conditional probability distribution
\begin{align}
\label{eq:models-filtering}
p(x_t \mid y_{1:t} ).
\end{align}
We have now introduced smoothing, where the states $x_{1:N}$ are estimated simultaneously, and filtering, where at each time instance the state $x_t$ is estimated. There is a large range of intermediate methods, where a batch of states $x_{t - L_1 : t + L_2}$, with $L_1$ and $L_2$ being positive integers, is estimated using the measurements $y_{1:t}$. This is related to fixed-lag smoothing and moving horizon estimation \citep{johansen:2011,raoRL:2001}.

\begin{figure}
  	\centering
    	\includegraphics[scale = 1]{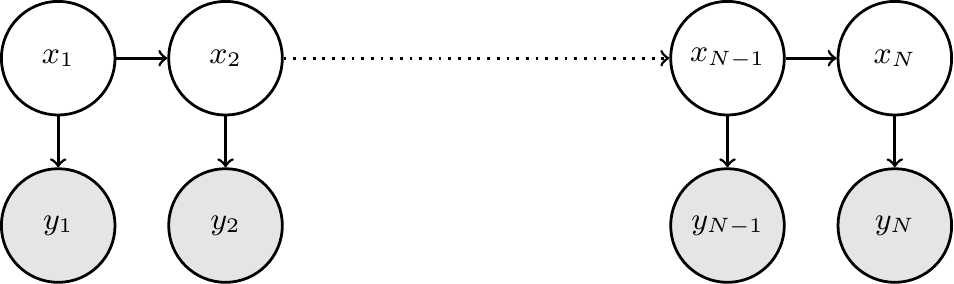}
    	\caption{An illustration of the structure of the pose estimation problem.}
    	\label{fig:models-structureProblem}
\end{figure}

The topic of how to estimate the conditional probability distributions for position and orientation estimation will be introduced in \Chapterref{cha:orientationEstimation}. We will now instead take a closer look at these distributions and their different components. A fundamental assumption here is that we assume that our models possess the \emph{Markov property}, implying that all information up to the current time $t$ is contained in the state $x_t$. This is illustrated in \Figureref{fig:models-structureProblem} in terms of a \emph{probabilistic graphical model}~\citep{bishop:2006}. The state $x_{t+1}$ can be seen to depend on $x_t$ and to result in the measurements $y_{t+1}$. It is conditionally independent of $x_{1:t-1}$ given the state~$x_t$. Using Bayes' rule and the Markov property, the conditional distributions~\eqref{eq:models-smoothing} and~\eqref{eq:models-filtering} can be decomposed as
\begin{subequations}
\label{eq:models-smoothingFiltProbs}
\begin{align}
\hspace{-2mm} p(x_{1:N}, \theta \mid y_{1:N} ) &\propto p(\theta) p(x_1 \mid \theta) \prod_{t=2}^N p(x_t \mid x_{t-1}, \theta) \prod_{t=1}^N p(y_t \mid x_t, \theta) \label{eq:models-smoothingProbs}, \hspace{-1mm} \\
p(x_t \mid y_{1:t} ) &\propto p(y_t \mid x_t) p(x_t \mid y_{1:t-1}) \label{eq:models-filteringProbs}.
\end{align}
\end{subequations}
The predictive distribution $p(x_t \mid y_{1:t-1})$ can be computed by \emph{marginalizing out} the previous state $x_{t-1}$ as
\begin{align}
&p(x_t \mid y_{1:t-1}) = \int p(x_t \mid x_{t-1}) p(x_{t-1} \mid y_{1:t-1}) \, \dint x_{t-1}.
\end{align}
In~\eqref{eq:models-smoothingFiltProbs}, $p(\theta)$ and $p(x_1 \mid \theta)$ encode our \emph{prior} information of $\theta$ and the knowledge of the state $x_1$ given $\theta$, respectively. The \emph{dynamics} are modeled in terms of $p(x_{t+1} \mid x_{t} , \theta)$ and $p(x_{t+1} \mid x_{t})$. The distributions $p(y_t \mid x_{t} , \theta)$ and $p(y_t \mid x_{t})$ model the information given by the measurements about the state and the parameters.

The dynamics of the state can be modeled in terms of a nonlinear function $f_t(\cdot)$ as
\begin{align}
\label{eq:models-dynamics}
x_{t+1} = f_t (x_t, w_t).
\end{align}
The uncertainty of the dynamic model is modeled in terms of $w_t$, which is often referred to as the \emph{process noise}. The model~\eqref{eq:models-dynamics} provides information about the distribution $p(x_{t+1} \mid x_t)$. More explicitly, if $w_t$ is Gaussian additive noise with $w_t \sim \mathcal{N}(0,Q)$, then
\begin{align}
p(x_{t+1} \mid x_t) \sim \mathcal{N}(x_{t+1} \, ; \, f_t(x_t), Q),
\end{align}
where we use the notation $\mathcal{N}(x_{t+1} \, ; \, f_t(x_t), Q)$ to explain that the random variable $x_{t+1}$ is normal distributed with mean $f_t(x_t)$ and covariance $Q$. 

The information given by the measurements about the state $x_t$ can be modeled as
\begin{align}
\label{eq:models-measEqn}
y_{t} = h_t (x_t, e_t),
\end{align}
where $h_t(\cdot)$ is a possibly nonlinear function and $e_t$ is the measurement noise. The measurement model~\eqref{eq:models-measEqn} provides information about the distribution $p(y_t \mid x_t)$. The combination of~\eqref{eq:models-dynamics},~\eqref{eq:models-measEqn} and a model of the prior $p(x_1)$ is referred to as a \emph{state space model} \citep{kailath:1980} which is widely used in a large number of fields.

\section{Parametrizing orientation}
\label{sec:models-paramOri}
Rotating a vector in $\mathbb{R}^3$ changes the \emph{direction} of the vector while retaining its \emph{length}. The group of rotations in $\mathbb{R}^3$ is the special orthogonal group $\SO{3}$. In this section we introduce four different ways of parametrizing orientations. Note that these describe the same quantity and can hence be used interchangeably. The different parametrizations can be converted to one another, see also \Appendixref{app:rotation}. There are differences in for instance the number of parameters used in the representation, the singularities and the uniqueness. 

\subsection{Rotation matrices}
We encountered rotation matrices already in \Chapterref{cha:sensors}. Rotation matrices $R \in \mathbb{R}^{3\times3}$ have the following properties
\begin{align}
  \label{eq:models-rotMatrixProperties}
  R R^\Transp = R^\Transp R = \mathcal{I}_3, \qquad \det R = 1.
\end{align}
The properties~\eqref{eq:models-rotMatrixProperties} provide an interpretation of the name special orthogonal group $\SO{3}$. All orthogonal matrices of dimension $3 \times 3$ have the property $R R^\Transp = R^\Transp R = \mathcal{I}_3$ and are part of the orthogonal group $O(3)$. The notion \emph{special} in $\SO{3}$ specifies that only matrices with $\det R = 1$ are considered rotations.

Consider two coordinate frames denoted $u$ and $v$. As was illustrated in \Exampleref{ex:sensors-rotationVectors}, a vector $x$ expressed in the $v$-frame can be rotated to the $u$-frame as
\begin{subequations}
\begin{align}
\label{eq:models-rotxVtoU}
x^\text{u} &= R^\text{uv} x^\text{v}, \\
\intertext{and conversely we have}
  \label{eq:models-rotxUtoV}
  x^\text{v} &= \left( R^\text{uv} \right)^\Transp x^\text{u} = R^\text{vu} x^\text{u}.
\end{align}
\end{subequations}
A rotation matrix is a unique description of the orientation. It has $9$ components which depend on each other as defined in~\eqref{eq:models-rotMatrixProperties}.

\subsection{Rotation vector}
As described by Leonhard Euler in \cite{euler:1775}, a rotation around a point is always equivalent to a single rotation around some axis through this point, see \cite{palaisPR:2009} for a number of proofs. This is generally referred to as \emph{Euler's rotation theorem}. Hence, it is possible to express the rotation between two coordinate frames in terms of an angle $\alpha$ and a unit vector $n$ around which the rotation takes place. In this section, we will derive a relation between the representation $\alpha, n$ and the rotation matrix parametrization from the previous section. Instead of directly considering the rotation of a coordinate frame, we start by considering the rotation of a vector. Note that a counterclockwise rotation of the coordinate frame is equivalent to a clockwise rotation of a vector, see \Exampleref{ex:models-rotVectorCoordFrame}.

\begin{myexample}{Rotation of a coordinate frame and rotation of a vector}%
\label{ex:models-rotVectorCoordFrame}%
Consider the 2D example in \Figureref{fig:models-rotateVectorCoordFrame}, where on the left, a vector $x$ is rotated clockwise by an angle $\alpha$ to $x_\star$. This is equivalent to (on the right) rotating the coordinate frame $v$ counterclockwise by an angle $\alpha$. Note that $x^\text{v}_\star = x^\text{u}$.

\begin{figure}
      \centering
      \includegraphics[scale = 1]{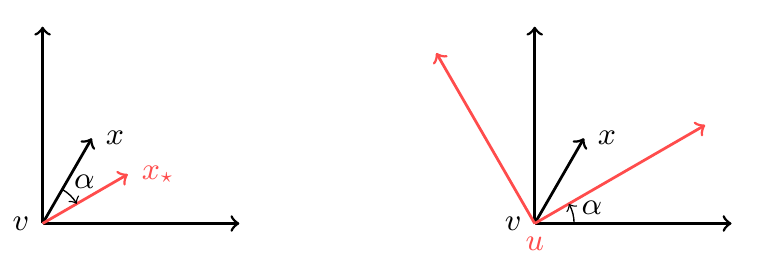}
      \caption[]{Left: clockwise rotation $\alpha$ of the vector $x$ to the vector $x_\star$. Right: counterclockwise rotation $\alpha$ of the coordinate frame $v$ to the coordinate frame~$u$.}
      \label{fig:models-rotateVectorCoordFrame}
\end{figure}
\end{myexample}

\begin{figure}
      \centering
      \includegraphics[scale = 1]{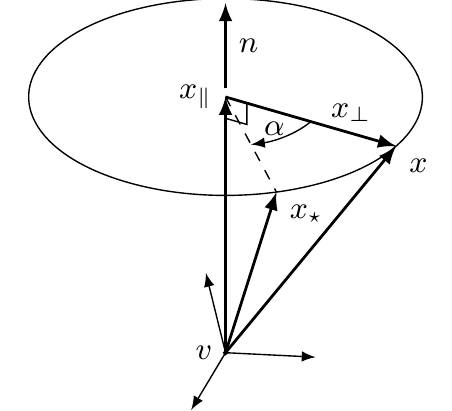}
      \caption[]{Clockwise rotation of a vector $x$ by an angle $\alpha$
        around the unit vector~$n$. The rotated vector is denoted by $x_\star$. The vector $x$ is decomposed in a component $x_\parallel$ that is parallel to the axis $n$, and a component $x_\perp$ that is orthogonal to it.}
      \label{fig:models-axisAngle}
\end{figure}

In \Figureref{fig:models-axisAngle}, a vector $x$ is rotated an angle $\alpha$ around the unit vector~$n$. We denote the rotated vector by $x_\star$. Suppose that $x$ as expressed in the coordinate frame $v$ is known (and denoted $x^\text{v}$) and that we want to express $x_\star^\text{v}$ in terms of $x^\text{v}$, $\alpha$ and $n$. It can first be recognized that the vector $x$ can be decomposed into a component parallel to the axis $n$, denoted $x_\parallel$, and a component orthogonal to it, denoted $x_\perp$, as
\begin{subequations}
\begin{align}
x^\text{v} &= x^\text{v}_\parallel + x^\text{v}_\perp. \\
\intertext{Based on geometric reasoning we can conclude that}
x^\text{v}_\parallel &= \left( x^\text{v} \cdot n^\text{v} \right) n^\text{v},
\end{align} 
\end{subequations} 
where $\cdot$ denotes the inner product. Similarly, $x_\star^\text{v}$ can be decomposed as
\begin{subequations}
\begin{align}
x^\text{v}_\star &= \left( x^\text{v}_\star \right)_\parallel + \left( x^\text{v}_\star \right)_\perp,
\intertext{where}
\left( x^\text{v}_\star \right)_\parallel &= x^\text{v}_\parallel, \\
\left( x^\text{v}_\star \right)_\perp &= x^\text{v}_\perp \cos \alpha + \left( x^\text{v} \times n^\text{v} \right) \sin \alpha.
\end{align}
\end{subequations}
Hence, $x^\text{v}_\star$ can be expressed in terms of $x^\text{v}$ as
\begin{align}
  x_\star^\text{v} 
  &= (x^\text{v} \cdot n^\text{v}) n^\text{v} 
    + (x^\text{v} - (x^\text{v} \cdot n^\text{v}) n^\text{v})\cos\alpha
    + (x^\text{v} \times n^\text{v})\sin\alpha \nonumber \\
  &= x^\text{v} \cos\alpha 
    + n^\text{v} (x^\text{v} \cdot n^\text{v})(1-\cos\alpha)
    - (n^\text{v} \times x^\text{v})\sin\alpha.
\end{align}
Denoting the rotated coordinate frame the $u$-frame and using the equivalence between $x^\text{v}_\star$ and $x^\text{u}$ as shown in \Exampleref{ex:models-rotVectorCoordFrame}, this implies that 
\begin{align}
  \label{eq:kin-rotation-aa}
  x^\text{u} = x^\text{v} \cos\alpha 
    + n^\text{v} (x^\text{v} \cdot n^\text{v})(1-\cos\alpha)
    - (n^\text{v} \times x^\text{v})\sin\alpha.
\end{align}
This equation is commonly referred to as the \emph{rotation formula} or \emph{Euler's formula}
\citep{shuster:1993}. Note that the combination of $n$ and
$\alpha$, or $\oriError = n \alpha$, is denoted as the \emph{rotation vector}
or the \emph{axis-angle parameterization}.  

To show the equivalence between~\eqref{eq:kin-rotation-aa} and the rotation matrix parametrization, we will rewrite~\eqref{eq:kin-rotation-aa}. Here, we make use of the fact that a cross product can be written as a matrix vector product. Given vectors $u$ and $v$ we have, 
\begin{align}
  \label{eq:models-crossproductMatrix}
  u \times v &= [u \times] v = - [v \times] u, 
  \qquad 
  [u \times] \triangleq 
  \begin{pmatrix}
    0 & -u_3 & u_2 \\
    u_3 & 0 & -u_1 \\
    -u_2 & u_1 & 0 
  \end{pmatrix},
\end{align}
where $u_1, u_2, u_3$ denote the three components of the vector $u$. Furthermore, given vectors $u$, $v$ and $w$, multiple cross products can be expanded in terms of the inner product as
\begin{align}
  \label{eq:models-crossproductExp}
  u \times (v \times w)
    = v (w \cdot u) - w (u \cdot v).
\end{align}
Using these relations, \eqref{eq:kin-rotation-aa} can be rewritten as
\begin{align}
  x^\text{u} &= x^\text{v} \cos\alpha 
    + n^\text{v} (x^\text{v} \cdot n^\text{v})(1-\cos\alpha)
    - (n^\text{v} \times x^\text{v})\sin\alpha \nonumber\\
    &= x^\text{v} \cos\alpha 
     + (n^\text{v} \times (n^\text{v} \times x^\text{v}) 
       + x^\text{v})(1-\cos\alpha)
     - (n^\text{v} \times x^\text{v})\sin\alpha \nonumber \\
  \label{eq:models-axisAngle-rotMatrix-der}
  &= \left( \mathcal{I}_3 - \sin\alpha [n^\text{v} \times] 
        + (1-\cos\alpha)[n^\text{v} \times]^2
      \right) x^\text{v}.
\end{align}
Comparing~\eqref{eq:models-axisAngle-rotMatrix-der} and~\eqref{eq:models-rotxVtoU}, it can be seen that a rotation matrix can be parametrized in terms of $\alpha, n$ as
\begin{align}
\label{eq:models-axisAngle-rotMatrix}
R^\text{uv}(n^\text{v},\alpha) = \mathcal{I}_3 - \sin\alpha [n^\text{v} \times] 
        + (1-\cos\alpha)[n^\text{v} \times]^2.
\end{align}
Note that equivalently, $R^\text{uv}(n^\text{v},\alpha)$ can also be written as
\begin{align}
\label{eq:models-rotMatrixAxisAngleExp}
R^\text{uv}(n^\text{v},\alpha) = \exp \left( - \alpha [n^\text{v} \times] \right), 
\end{align}
since 
\begin{align}
\label{eq:models-rotMatrixAxisAngle}
&\exp \left( - \alpha [n^\text{v} \times] \right) = \sum_{k = 0}^\infty \tfrac{1}{k!} \left( - \alpha [n^\text{v} \times] \right)^k \nonumber \\
&\qquad = \mathcal{I}_3 - \alpha [n^\text{v} \times] + \tfrac{1}{2!} \alpha^2 [n^\text{v} \times]^2 + 
\tfrac{1}{3!} \alpha^3 [n^\text{v} \times] - 
\tfrac{1}{4!} \alpha^4 [n^\text{v} \times]^2 - \hdots \nonumber \\
&\qquad= \mathcal{I}_3 - \left( \alpha - \tfrac{1}{3!} \alpha^3 + \hdots \right) [n^\text{v} \times] + 
\left( \tfrac{1}{2!} \alpha^2 - \tfrac{1}{4!} \alpha^4 + \hdots \right) [n^\text{v} \times]^2 \nonumber \\
&\qquad= \mathcal{I}_3 - \sin\alpha [n^\text{v} \times] + (1-\cos\alpha)[n^\text{v} \times]^2.
\end{align}
The rotation vector introduced in this section parametrizes the orientation in only three parameters. It is, however, not a unique parametrization since adding $2 \pi$ to any angle $\alpha$ results in the same orientation. This is called \emph{wrapping}. As shown in~\eqref{eq:models-axisAngle-rotMatrix} and~\eqref{eq:models-rotMatrixAxisAngleExp}, the rotation matrix can straightforwardly be expressed in terms of the axis-angle representation. 

\subsection{Euler angles}
\begin{figure}
	\centering
	\includegraphics[scale = 1]{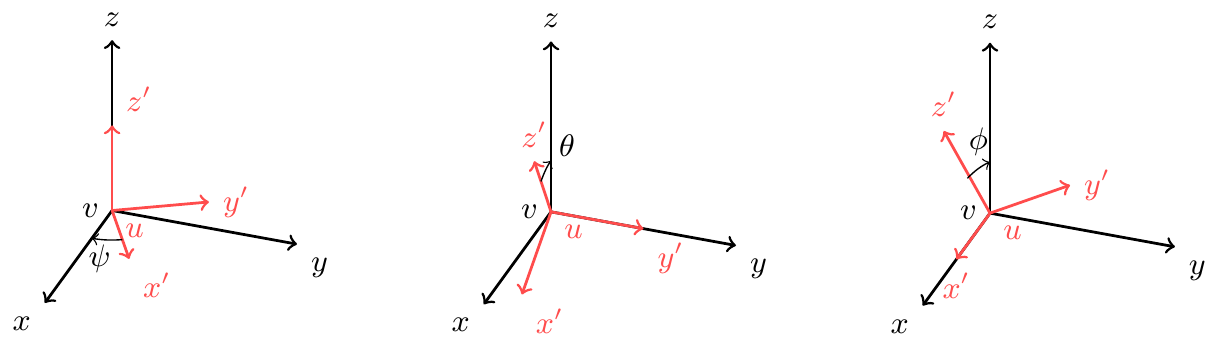}
		\caption{Definition of Euler angles as used in this work with left: rotation~$\psi$ around the $z$-axis, middle: rotation~$\theta$ around the $y$-axis and right: rotation~$\phi$ around the $x$-axis.}
	\label{fig:models-eulerAngles}
\end{figure}

Rotation can also be defined as a consecutive rotation around three axes in terms of so-called \emph{Euler angles}. We use the convention $(z,y,x)$ which first rotates an angle $\psi$ around the $z$-axis, subsequently an angle $\theta$ around the $y$-axis and finally an angle $\phi$ around the $x$-axis. These angles are illustrated in \Figureref{fig:models-eulerAngles}. Assuming that the $v$-frame is rotated by $(\psi, \theta, \phi)$ with respect to the $u$-frame as illustrated in this figure, the rotation matrix $R^\text{uv}$ is given by 

{\footnotesize{
\begin{align}
\label{eq:models-rotMatrix}
R^\text{uv} &= R^\text{uv}(e_1, \phi) R^\text{uv}(e_2, \theta) R^\text{uv}(e_3, \psi) \\
&= \begin{pmatrix} 1 & 0 & 0 \\ 0 & \cos \phi & \sin \phi \\ 0 & -\sin \phi & \cos \phi \end{pmatrix}
\begin{pmatrix} \cos \theta & 0 & -\sin \theta \\ 0 & 1 & 0 \\ \sin \theta & 0 & \cos \theta \end{pmatrix}
\begin{pmatrix} \cos \psi & \sin \psi & 0 \\ -\sin \psi & \cos \psi & 0 \\ 0 & 0 & 1 \end{pmatrix} \nonumber \\
&= \begin{pmatrix} \cos \theta \cos \psi & \cos \theta \sin \psi & -\sin \theta \\ 
\sin \phi \sin \theta \cos \psi - \cos \phi \sin \psi & \sin \phi \sin \theta \sin \psi + \cos \phi \cos \psi & \sin \phi \cos \theta \\ 
\cos \phi \sin \theta \cos \psi + \sin \phi \sin \psi & \cos \phi \sin \theta \sin \psi - \sin \phi \cos \psi & \cos \phi \cos \theta \end{pmatrix},\nonumber
\end{align}}}%
where we make use of the notation introduced in~\eqref{eq:models-axisAngle-rotMatrix} and the following definition of the unit vectors 
\begin{align}
e_1 = \begin{pmatrix} 1 & 0 & 0 \end{pmatrix}^\Transp, \quad e_2 = \begin{pmatrix} 0 & 1 & 0 \end{pmatrix}^\Transp, \quad e_3 = \begin{pmatrix} 0 & 0 & 1 \end{pmatrix}^\Transp.
\end{align}
The $\psi, \theta, \phi$ angles are also often referred to as yaw (or heading), pitch and roll, respectively. Furthermore, roll and pitch together are often referred to as inclination. 

Similar to the rotation vector, Euler angles parametrize orientation as a three-dimensional vector. Euler angle representations are not unique descriptions of a rotation for two reasons. First, due to wrapping of the Euler angles, the rotation $(0, 0, 0)$ is for instance equal to $(0, 0, 2 \pi k)$ for any integer $k$.  Furthermore, setting $\theta = \tfrac{\pi}{2}$ in~\eqref{eq:models-rotMatrix}, leads to
\begin{align}
R^\text{uv} &= \begin{pmatrix} 0 & 0 & -1 \\ 
\sin \phi \cos \psi - \cos \phi \sin \psi & \sin \phi \sin \psi + \cos \phi \cos \psi & 0 \\ 
\cos \phi \cos \psi + \sin \phi \sin \psi & \cos \phi \sin \psi - \sin \phi \cos \psi & 0 \end{pmatrix} \nonumber \\
&= \begin{pmatrix} 0 & 0 & -1 \\ 
\sin (\phi - \psi) & \cos (\phi - \psi) & 0 \\ 
\cos (\phi - \psi) & - \sin (\phi - \psi) & 0 \end{pmatrix}.
\end{align}
Hence, only the rotation $\phi - \psi$ can be observed. Because of this, for example the rotations $(\tfrac{\pi}{2}, \tfrac{\pi}{2}, 0)$, $(0, \tfrac{\pi}{2}, -\tfrac{\pi}{2})$, $(\pi, \tfrac{\pi}{2},\tfrac{\pi}{2})$ are all three equivalent. This is called \emph{gimbal lock}~\citep{diebel:2006}. 

\subsection{Unit quaternions}
A commonly used parametrization of orientation is that of unit quaternions. Quaternions were first introduced by~\cite{hamilton:1844} and are widely used in orientation estimation algorithms, see \eg \cite{kuipers:1999,hol:2011}. A unit quaternion use a $4$-dimensional representation of the orientation according to
\begin{align}
q = \begin{pmatrix} q_0 & q_1 & q_2 & q_3 \end{pmatrix}^\Transp = \begin{pmatrix} q_0 \\ q_v \end{pmatrix}, \qquad q \in \mathbb{R}^4, \qquad \| q\|_2 = 1.
\end{align}
A unit quaternion is not a unique description of an orientation. The reason for this is that if $q$ represents a certain orientation, then $-q$ describes the same orientation. 

A rotation can be defined using unit quaternions as
\begin{align}
  \label{eq:models-rotatex_q}
  \bar x^\text{u} = 
    q^\text{uv} \odot \bar x^\text{v} \odot \left( q^\text{uv} \right)^\conj,
\end{align}
where $\left( q^\text{uv} \right)^\conj = q^\text{vu}$ denotes the quaternion conjugate, defined as
\begin{align}
\label{eq:models-quatConj}
q^\conj = \begin{pmatrix} q_0 & -q_v^\Transp \end{pmatrix}^\Transp,
\end{align}
and $\bar x^\text{v}$ denotes the quaternion representation of $x^\text{v}$ as 
\begin{align}
\label{eq:models-quatVector}
\bar x^\text{v} = \begin{pmatrix} 0 & (x^\text{v})^\Transp \end{pmatrix}^\Transp.
\end{align}
Note that~\eqref{eq:models-quatVector} is typically not a unit quaternion. The notation $\odot$ denotes the quaternion multiplication given by 
\begin{align}
\label{eq:models-quatMult}
p \odot q = \begin{pmatrix} p_0 q_0 - p_v \cdot q_v \\ p_0 q_v + q_0 p_v + p_v \times q_v \end{pmatrix} = p^\leftMult q = q^\rightMult p,
\end{align}
where 
\begin{align}
\label{eq:models-leftRightQuatMult}
p^\leftMult &\triangleq \begin{pmatrix} p_0 & -p_v^\Transp \\ p_v & p_0 \mathcal{I}_3 + [p_v \times] \end{pmatrix}, \qquad q^\rightMult \triangleq \begin{pmatrix} q_0 & -q_v^\Transp \\ q_v & q_0 \mathcal{I}_3 - [q_v \times] \end{pmatrix}.
\end{align}
Using~\eqref{eq:models-quatConj}--\eqref{eq:models-leftRightQuatMult},~\eqref{eq:models-rotatex_q} can be written as
\begin{align}
\label{eq:models-quatRotInterm}
\bar x^\text{u} &= \left( q^\text{uv} \right)^\leftMult \left( q^\text{vu} \right)^\rightMult \bar x^\text{v} \nonumber \\
&= \begin{pmatrix} q_0 & -q_v^\Transp \\ q_v & q_0 \mathcal{I}_3 + [q_v \times] \end{pmatrix} \begin{pmatrix} q_0 & q_v^\Transp \\ -q_v & q_0 \mathcal{I}_3 + [q_v \times] \end{pmatrix} \begin{pmatrix} 0 \\ x^\text{v} \end{pmatrix} \nonumber \\
&= \begin{pmatrix} 1 & 0_{1 \times 3} \\ 0_{3 \times 1} & q_v q_v^\Transp + q_0^2 \mathcal{I}_3 + 2 q_0 [q_v \times] + [q_v \times]^2 \end{pmatrix} \begin{pmatrix} 0 \\ x^\text{v} \end{pmatrix}.
\end{align}
Comparing~\eqref{eq:models-quatRotInterm} to~\eqref{eq:models-axisAngle-rotMatrix}, it can be recognized that if we choose 
\begin{align}
\label{eq:models-axisAngle-quat}
q^\text{uv}(n^\text{v},\alpha) = \begin{pmatrix} \cos \tfrac{\alpha}{2} \\ - n^\text{v} \sin \tfrac{\alpha}{2} \end{pmatrix}, 
\end{align}
the two rotation formulations are equivalent since
\begin{align}
\bar x^\text{u} &= \begin{pmatrix} 1 & 0_{1 \times 3} \\ 0_{3 \times 1} & \mathcal{I}_3 - 2 \cos \tfrac{\alpha}{2} \sin \tfrac{\alpha}{2} [n^\text{v} \times] + 2 \sin^2 \tfrac{\alpha}{2} [n^\text{v} \times]^2 \end{pmatrix} \begin{pmatrix} 0 \\ x^\text{v} \end{pmatrix} \nonumber \\
&= \begin{pmatrix} 1 & 0_{1 \times 3} \\ 0_{3 \times 1} & \mathcal{I}_3 - \sin \alpha [n^\text{v} \times] + \left( 1 - \cos \alpha \right) [n^\text{v} \times]^2 \end{pmatrix} \begin{pmatrix} 0 \\ x^\text{v} \end{pmatrix}.
\end{align}
Here, we made use of standard trigonometric relations and the fact that since $\| n^\text{v} \|_2= 1$, $n^\text{v} \left( n^\text{v} \right)^\Transp = \mathcal{I}_3 + [n^\text{v} \times]^2$. Hence, it can be concluded that $q^\text{uv}$ can be expressed in terms of $\alpha$ and $n^\text{v}$ as in~\eqref{eq:models-axisAngle-quat}.

Equivalently, $q^\text{uv}(n^\text{v},\alpha)$ can also be written as 
\begin{align}
q^\text{uv}(n^\text{v},\alpha) = \exp( - \tfrac{\alpha}{2} \bar n^\text{v} ) = \sum_{k = 0}^\infty \tfrac{1}{k!} \left( - \tfrac{\alpha}{2} \bar n^\text{v} \right)^k,
\end{align}
where 
\begin{subequations}
\begin{align}
\left( \bar n^\text{v} \right)^0 &= \begin{pmatrix} 1 & 0 & 0 & 0 \end{pmatrix}^\Transp, \\
\left( \bar n^\text{v} \right)^1 &= \begin{pmatrix} 0 & \left( n^\text{v} \right)^\Transp \end{pmatrix}^\Transp, \\
\left( \bar n^\text{v} \right)^2 &= \bar n^\text{v} \odot \bar n^\text{v} = \begin{pmatrix} - \| n^\text{v} \|_2^2 & 0_{3 \times 1} \end{pmatrix}^\Transp = \begin{pmatrix} - 1 & 0_{3 \times 1} \end{pmatrix}^\Transp, \\
\left( \bar n^\text{v} \right)^3 &= \begin{pmatrix} 0 & - \left( n^\text{v} \right)^\Transp \end{pmatrix}^\Transp, 
\end{align}
\end{subequations}
This leads to 
\begin{align}
\label{eq:models-quatAxisAngle}
q^\text{uv}(n^\text{v},\alpha) &= \exp( - \tfrac{\alpha}{2} \bar n^\text{v} ) = \sum_{k = 0}^\infty \tfrac{1}{k!} \left( - \tfrac{\alpha}{2} \bar n^\text{v} \right)^k \nonumber \\
&= \begin{pmatrix} 1 - \tfrac{1}{2!} \tfrac{\alpha^2}{4} + \tfrac{1}{4!} \tfrac{\alpha^4}{16} - \hdots \\
- \tfrac{\alpha}{2} n^\text{v} + \tfrac{1}{3!} \tfrac{\alpha^3}{8} n^\text{v} - \tfrac{1}{5!} \tfrac{\alpha^5}{32} n^\text{v} + \hdots
\end{pmatrix} \nonumber \\
&= \begin{pmatrix} \cos \tfrac{\alpha}{2} \\ - n^\text{v} \sin \tfrac{\alpha}{2}
\end{pmatrix}.
\end{align}
Note the similarity to~\eqref{eq:models-rotMatrixAxisAngleExp} and~\eqref{eq:models-rotMatrixAxisAngle}. The reason why both rotation matrices and unit quaternions can be described in terms of an exponential of a rotation vector will be discussed in \Sectionref{sec:models-linearization}.

\section{Probabilistic orientation modeling}
\label{sec:models-probModelingOri}
The four parametrizations of orientation discussed in \Sectionref{sec:models-paramOri} can be used interchangeably. However, the choice of which parametrization to use as states $x_t$ in the filtering and smoothing problems introduced in \Sectionref{sec:models-probModeling} has significant impact on the workings of the algorithm. An important reason for this is that estimation algorithms typically assume that the unknown states and parameters are represented in Euclidean space. For instance, they assume that the subtraction of two orientations gives information about the difference in orientation and that the addition of two orientations is again a valid orientation. For the four parametrizations discussed in \Sectionref{sec:models-paramOri}, this is generally not true. For instance, due to wrapping and gimbal lock, subtraction of Euler angles and rotation vectors can result in large numbers even in cases when the rotations are similar. Also, addition and subtraction of unit quaternions and rotation matrices do not in general result in a valid rotation. The equality constraints on the norm of unit quaternions and on the determinant and the orthogonality of rotation matrices are typically hard to include in the estimation algorithms. In \Sectionref{sec:models-linearization}, we will discuss a method to represent orientation in estimation algorithms that deals with the issues described above. It is frequently used in the algorithms that will be described in \Chapterref{cha:orientationEstimation}. In \Sectionref{sec:models-altProbOriModels}, we will also discuss some alternative methods to parametrize orientation for estimation purposes. 

\subsection{Linearization}
\label{sec:models-linearization}
As mentioned in \Sectionref{sec:models-paramOri}, the group of rotations in three dimensions is the special orthogonal group $\SO{3}$. More specifically, $\SO{3}$ is a so-called \emph{matrix Lie group}. For a discussion on the properties of matrix Lie groups and on the reasons why $\SO{3}$ is indeed such a group we refer the reader to \eg \cite{barfoot:2016}. Since rotations are a matrix Lie group, there exists an \emph{exponential map} from a corresponding Lie algebra. Using this property, it is possible to represent orientations on $\SO{3}$ using unit quaternions or rotation matrices, while orientation deviations are represented using rotation vectors on $\mathbb{R}^3$, see \eg \cite{bloeschEtAl:2016}. Hence, we encode an orientation $q^\text{nb}_t$ in terms of a \emph{linearization point} parametrized either as a unit quaternion $\tilde{q}^\text{nb}_t$ or as a rotation matrix $\tilde{R}^\text{nb}_t$ and an \emph{orientation deviation} using a rotation vector $\oriError_t$. Assuming that the orientation deviation is expressed in the body frame~$n$,\footnote{A similar derivation can be done by assuming an orientation deviation in the navigation frame $b$.}
	\begin{align}
	\label{eq:models-oriDev}
	q^\text{nb}_t = \exp \left( \tfrac{\bar \oriError_t^\text{n}}{2} \right) \odot \tilde{q}^\text{nb}_t, \qquad
	R^\text{nb}_t = \exp \left( [\oriError_t^\text{n} \times] \right) \tilde{R}^\text{nb}_t,
	\end{align}
where analogously to~\eqref{eq:models-quatAxisAngle} and~\eqref{eq:models-rotMatrixAxisAngle}, 
\begin{subequations}
\begin{align}
\exp (\bar \oriError) &= \begin{pmatrix} \cos \| \oriError \|_2 \\ \tfrac{\oriError}{\| \oriError \|_2} \sin \| \oriError \|_2 \end{pmatrix}, \\
\exp ([\oriError \times]) &= \mathcal{I}_3 + \sin\left( \| \oriError \|_2 \right) \left[ \tfrac{\oriError}{\| \oriError \|_2} \times \right] + \nonumber \\
& \qquad \quad \left(1-\cos \left( \| \oriError \|_2\right) \right) \left[ \tfrac{\oriError}{\| \oriError \|_2} \times \right]^2.
\label{eq:models-expqR}
\end{align}
\end{subequations}
For notational convenience, in the remainder we will use the mappings 
\begin{subequations}
\begin{align}
q &= \expq(\oriError), 
 \qquad \expq : \mathbb{R}^3 \rightarrow \{ q \in \mathbb{R}^{4}: \| q \|_2 = 1 \}, \label{eq:models-expq-map} \\
R &= \expR(\oriError), \nonumber \\
&\qquad \expR : \mathbb{R}^3 \rightarrow \{ R \in \mathbb{R}^{3 \times 3}: R R^\Transp = \mathcal{I}_3,~\det R = 1\}, \label{eq:models-expR-map}
\end{align} 
\label{eq:models-expqR-map}%
\end{subequations}%
which allow us to rewrite~\eqref{eq:models-oriDev} as 
    \begin{align}
    q^\text{nb}_t = \expq \left( \tfrac{\oriError_t^\text{n}}{2} \right) \odot \tilde{q}^\text{nb}_t, \qquad
    R^\text{nb}_t = \expR \left( \oriError_t^\text{n} \right) \tilde{R}^\text{nb}_t.
    \label{eq:models-oriDev_expqR}
    \end{align}
The reverse mappings are defined as
\begin{subequations}
\begin{align}
\oriError &= \logq (q) = \tfrac{\arccos q_0}{\sin \arccos q_0} q_v = \tfrac{\arccos q_0}{\| q_v\|_2} q_v, \nonumber \\
&\qquad \logq : \{ q \in \mathbb{R}^{4}: \| q \|_2 = 1 \} \rightarrow \mathbb{R}^3, \label{eq:models-logq-map} \\
\oriError &= \logR (R) = \begin{pmatrix} (\log R)_{32} \\ (\log R)_{13} \\ (\log R)_{21} \end{pmatrix}, \nonumber \\ 
&\qquad \logR : \{ R \in \mathbb{R}^{3 \times 3}: R R^\Transp = \mathcal{I}_3,~\det R = 1\} \rightarrow \mathbb{R}^3, \label{eq:models-logR-map}
\end{align}
\label{eq:models-logqR}%
\end{subequations}%
where $\log R$ is the standard matrix logarithm. Since we typically assume that $\oriError_t^\text{n}$ is small, we will frequently make use of the following approximations
\begin{subequations}
\begin{align}
\expq (\oriError) &\approx \begin{pmatrix} 1 \\ \oriError \end{pmatrix}, \qquad &\logq (q) &\approx q_v, \\
\expR (\oriError) &\approx \mathcal{I}_3 + [\oriError \times], \qquad
&\logR (R) &\approx \begin{pmatrix} R_{32} & R_{13} & R_{21} \end{pmatrix}^\Transp.
\end{align}
\label{eq:models-logexpq-approx}
\end{subequations}

The idea briefly outlined in this section is closely related to approaches used to estimate orientation in robotics, see \eg \cite{grisettiKSB:2010,grisettiKSFH:2010,bloeschEtAl:2016,barfoot:2016,forsterCDS:2016}. It is also related to the so-called \gls{mekf} frequently used in aeronautics, see \eg \cite{markley:2003,crassidisMC:2007}. 

\subsection{Alternative methods}
\label{sec:models-altProbOriModels}
An alternative method to estimate orientation assumes that the states representing the orientation lie on a manifold. This can be done by modeling the orientation and its uncertainty using a \emph{spherical distribution} which naturally restricts the orientation estimates and their uncertainties to be in $\SO{3}$. In recent years, a number of approaches have been proposed to estimate the orientation using these kinds of distributions. For instance, in \cite{kurzGJH:2013,gilitschenskiKJH:2016,gloverK:2013} algorithms are presented to estimate orientation by modeling it using a Bingham distribution. 

The difficulties caused by directly using one of the four orientation parametrizations introduced in \Sectionref{sec:models-paramOri} in orientation estimation algorithms is widely recognized. Nevertheless, a large number of approaches directly uses these parametrizations in estimation algorithms. For instance, it is common practice to use unit quaternions in estimation algorithms and to normalize the resulting quaternions each time they loose their normalization, see \eg \cite{sabatini:2006,marinsYBMZ:2001,madgwickHV:2011}. Different approaches to handle the normalization of the quaternions in these algorithms are discussed in~\cite{julierV:2007}.

\section{Measurement models}
\label{sec:models-measModels}
In the past two sections, we have focused on how orientations can be parametrized. In this section, we will go back to the probabilistic models for the position and orientation estimation problems introduced in \Sectionref{sec:models-probModeling} and provide different measurement models $p(y_t \mid x_t, \theta)$. 

\subsection{Gyroscope measurement models}
\label{sec:models-gyrMeasModel}
As discussed in \Sectionref{sec:sensors-angVel}, the gyroscope measures the angular velocity $\omega_{\text{ib}}^\text{b}$ at each time instance $t$. However, as shown in \Sectionref{sec:sensors-errors}, its measurements are corrupted by a slowly time-varying bias $\delta_{\omega,t}$ and noise $e_{\omega,t}$. Hence, the gyroscope measurement model is given by
\begin{align}
\label{eq:models-gyrMeasModelGeneral}
y_{\omega,t} = \omega_{\text{ib},t}^\text{b} + \delta_{\omega,t}^\text{b} + e_{\omega,t}^\text{b}.
\end{align}
As was shown in \Figureref{fig:sensors-gyrMeasNoiseHist}, the gyroscope measurement noise is quite Gaussian. Because of this, it is typically assumed that $e_{\omega,t}^\text{b} \sim \mathcal{N}(0, \Sigma_\omega)$. If the sensor is properly calibrated, the measurements in the three gyroscope axes are independent. In that case, it can be assumed that
\begin{align}
\Sigma_\omega = \begin{pmatrix} \sigma_{\omega,x}^2 & 0 & 0 \\ 0 & \sigma_{\omega,y}^2 & 0 \\ 0 & 0 & \sigma_{\omega,z}^2 \end{pmatrix}.
\end{align}

The gyroscope bias $\delta_{\omega,t}^\text{b}$ is slowly time-varying, as discussed in \Sectionref{sec:sensors-errors}. There are two conceptually different ways to treat this slowly time-varying bias. One is to treat the bias as a constant parameter, assuming that it typically changes over a longer time period than the time of the experiment. The bias can then either be pre-calibrated in a separate experiment, or it can be considered to be part of the unknown parameters~$\theta$ as introduced in \Sectionref{sec:models-probModeling}. Alternatively, it can be assumed to be slowly time-varying. This can be justified either by longer experiment times or by shorter bias stability. In the latter case, $\delta_{\omega,t}^\text{b}$ can instead be considered as part of the state vector $x_t$ and can for instance be modeled as a random walk 
\begin{align}
\label{eq:models-gyrBiasRandomWalk}
\delta_{\omega,t+1}^\text{b} = \delta_{\omega,t}^\text{b} + e^\text{b}_{\delta_\omega,t},
\end{align}
where $e^\text{b}_{\delta_\omega,t} \sim \mathcal{N}(0, \Sigma_{\delta_{\omega},t})$ represents how constant the gyroscope bias actually is. 

Modeling the sensor noise and bias is related to the sensor properties. However, there are also modeling choices related to the experiments that can be made. 
As described in \Sectionref{sec:sensors-angVel}, the angular velocity $\omega_{\text{ib}}^\text{b}$ can be expressed as
\begin{align}
\omega_{\text{ib},t}^\text{b}
= R^{\text{bn}}_t \left(
\omega_{\text{ie},t}^\text{n} +
\omega_{\text{en},t}^\text{n} \right)
+ \omega_{\text{nb},t}^\text{b}.
\end{align}
If the sensor does not travel over significant distances as compared to the size of the earth --- which is often the case for the applications discussed in \Chapterref{cha:introduction} --- the navigation frame $n$ can safely be assumed to be stationary. In that case, the transport rate $\omega_{\text{en},t}^\text{n}$ is zero. Although the earth rotation $\omega_{\text{ie}}$ as expressed in the body frame $b$ is not constant, its magnitude as compared to the magnitude of the actual measurements is fairly small (see \Sectionref{sec:sensors-angVel} and the experimental data presented in \Exampleref{ex:sensors-inertialMeasurements}). Assuming that the earth rotation is negligible and the navigation frame is stationary leads to the following simplified measurement model
\begin{align}
\label{eq:models-gyrMeasModel}
y_{\omega,t} = \omega_{\text{nb},t}^\text{b} + \delta_{\omega,t}^\text{b} + e_{\omega,t}^\text{b}.
\end{align}

\subsection{Accelerometer measurement models}
\label{sec:models-accMeasModel}
The accelerometer measures the specific force $f^\text{b}_t$ at each time instance~$t$, see also \Sectionref{sec:sensors-specForce}. As shown in \Sectionref{sec:sensors-errors}, the accelerometer measurements are typically assumed to be corrupted by a bias $\delta_{\text{a},t}$ and noise $e_{\text{a},t}$ as
\begin{align}
\label{eq:models-accMeasModelGeneral}
y_{\text{a},t} 
= f^\text{b}_t 
+ \delta_{\text{a},t}^\text{b}
+ e_{\text{a},t}^\text{b}.
\end{align}
The accelerometer noise is typically quite Gaussian as was illustrated in \Figureref{fig:sensors-accMeasNoiseHist} and can hence be modeled as $e_{\text{a},t}^\text{b} \sim \mathcal{N}(0, \Sigma_\text{a})$. For a properly calibrated sensor, the covariance matrix $\Sigma_\text{a}$ can often be assumed to be diagonal. 

The accelerometer bias $\delta_{\text{a},t}^\text{b}$ is slowly time-varying. Similar to the gyroscope bias, the accelerometer bias can either be modeled as a constant parameter, or as part of the time-varying state, for instance using a random walk model as in~\eqref{eq:models-gyrBiasRandomWalk}.

As introduced in \Sectionref{sec:sensors-specForce}, the specific force measured by the accelerometer is given by
\begin{align}
\label{eq:models-specForce}
f^\text{b} = R^{\text{bn}} 
( a_{\text{ii}}^\text{n} - g^\text{n} ).
\end{align}
Assuming that the navigation frame is fixed to the earth frame, we derived a relation for $a_\text{ii}^\text{n}$ as
\begin{align}
\label{eq:models-aii-ann} 
a_\text{ii}^\text{n} = 
a_\text{nn}^\text{n} + 
2 \omega_\text{ie}^\text{n} \times v^\text{n}_\text{n} + 
\omega_\text{ie}^\text{n} \times \omega_\text{ie}^\text{n}
\times p^\text{n}. 
\end{align}
The centrifugal acceleration $\omega_\text{ie}^\text{n} \times \omega_\text{ie}^\text{n}
\times p^\text{n}$ is typically absorbed in the local gravity vector. The magnitude of the Coriolis acceleration is small compared to the magnitude of the accelerometer measurements (see \Exampleref{ex:sensors-magCentrCor} and the experimental data presented in \Exampleref{ex:sensors-inertialMeasurements}). Neglecting this term leads to the following simplified measurement model
\begin{align}
\label{eq:models-accMeasModel}
y_{\text{a},t} 
= R^{\text{bn}}_t  ( a_{\text{nn}}^\text{n} - g^\text{n} )
+ \delta_{\text{a},t}^\text{b}
+ e_{\text{a},t}^\text{b}.
\end{align} 

Since the accelerometer measures both the local gravity vector and the linear acceleration of the sensor, it provides information both about the change in position and about the inclination of the sensor. For orientation estimation, only the information about the inclination is of concern. Hence, a model for the linear acceleration needs to be made to express the relation between the inclination and the measurements. To model this, it can be recognized that in practice, most accelerometer measurements are dominated by the gravity vector, as illustrated in \Exampleref{ex:models-accMagnitude}. 

\begin{myexample}{Magnitude of a sensor's linear acceleration}%
\label{ex:models-accMagnitude}%
Let us consider a 1D example where a sensor has an initial velocity $v_1 = 0~\metrepersecond$ and accelerates with $a_\text{nn}^\text{n} = 9.82~\metrepersquaresecond$. After $4.51$ seconds, the sensor will have traveled $100$ meters. This is about twice as fast as the world record currently held by Usain Bolt. In fact, humans can reach fairly high accelerations but can only accelerate for a short time. Naturally, cars can accelerate to higher velocities than humans. The sensor in this example has reached a final velocity of $160~\kilo\metre\per\hour$. Even in the case of a car it is therefore unlikely that it can have an acceleration this high for a long period of time. 
\end{myexample}

Since the accelerometer measurements are typically dominated by the gravity vector, a commonly used model assumes the linear acceleration to be approximately zero 
\begin{align}
\label{eq:models-accMeasModelZeroAcc}
y_{\text{a},t} 
= - R^{\text{bn}}_t g^\text{n}
+ \delta_{\text{a},t}^\text{b}
+ e_{\text{a},t}^\text{b}.
\end{align} 
Naturally, the model~\eqref{eq:models-accMeasModelZeroAcc} is almost never completely true. However, it can often be used as a sufficiently good approximation of reality. Note that the noise term $e_{\text{a},t}^\text{b}$ in this case does not only represent the measurement noise, but also the model uncertainty. The model~\eqref{eq:models-accMeasModelZeroAcc} can for instance be used in combination with \emph{outlier rejection} where measurements that clearly violate the assumption that the linear acceleration is zero are disregarded. It is also possible to adapt the noise covariance matrix $\Sigma_\text{a}$, depending on the sensor's acceleration~\citep{foxlin:1996,rehbinderH:2004}. Furthermore, it is possible to instead model the acceleration based on physical reasoning~\citep{luinge:2002}. 

\subsection{Modeling additional information}
\label{sec:models-addMeasModels}
In this section we will discuss models for the measurements we use to complement the inertial sensor measurements. For orientation estimation we use magnetometers, while for pose estimation we use position measurements.

\paragraph{Magnetometer models}
Magnetometers measure the local magnetic field, consisting of both the earth magnetic field and the magnetic field due to the presence of magnetic material. The (local) earth magnetic field is denoted $m^\text{n}$ and it is illustrated in \Figureref{fig:models-magneticField}. Its horizontal component points towards the earth's magnetic north pole. The ratio between the horizontal and vertical component depends on the location on the earth and can be expressed in terms of the so-called dip angle~$\delta$. The dip angle and the magnitude of the earth magnetic field are accurately known from geophysical studies, see \eg \cite{ncei-tutorial}. 

Assuming that the sensor does not travel over significant distances as compared to the size of the earth, the local earth magnetic field can be modeled as being constant. In case no magnetic material is present in the vicinity of the sensor, orientation information can be deduced from the magnetometer. More specifically, magnetometers are typically used to complement accelerometers to provide information about the sensor heading, \ie about the orientation around the gravity vector which can not be determined from the accelerometer measurements. Magnetometers provide information about the heading in all locations on the earth except on the magnetic poles, where the local magnetic field $m^\text{n}$ is vertical. Orientation can be estimated based on the \emph{direction} of the magnetic field. The \emph{magnitude} of the field is irrelevant. Because of this, without loss of generality we model the earth magnetic field as 
\begin{align}
\label{eq:models-localMagField}
m^\text{n}  = \begin{pmatrix} \cos \delta & 0 & \sin \delta \end{pmatrix}^\Transp,
\end{align}
\ie we assume that $\| m^\text{n} \|_2 = 1$. Assuming that the magnetometer only measures the local magnetic field, its measurements $y_{\text{m},t}$ can be modeled as
\begin{align}
\label{eq:models-magMeasModel}
y_{\text{m},t} &= R^\text{bn}_t m^\text{n} + e_{\text{m},t}, 
\end{align}
where $e_{\text{m},t} \sim \mathcal{N}(0,\Sigma_\text{m})$. The noise $e_{\text{m},t}$ represents the magnetometer measurement noise as well as the model uncertainty. Note that using the models~\eqref{eq:models-localMagField} and~\eqref{eq:models-magMeasModel}, we define the heading with respect to the magnetic north, which is sufficient for most applications. In case we would be interested in navigation with respect to the true north instead, the magnetic declination needs to be taken into account. Since the magnetic declination depends on the location on the earth it is necessary to know the location of the sensor to correct for this. 

\afterpage{
\begin{figure}
  \centering
  	\subfigure[]{
  	\includegraphics[scale = 1]{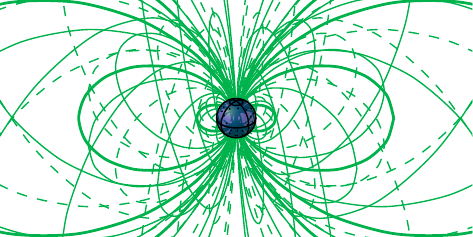}
  	} 
 	\subfigure[]{
	\includegraphics[scale = 1]{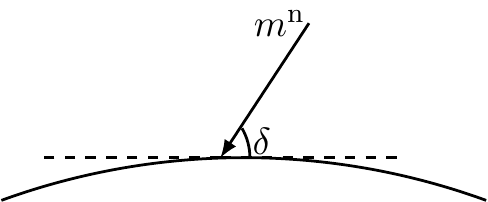}
	} 
  \caption[]{(a) Schematic of the earth magnetic field lines (green) around the earth (blue).\footnotemark~(b) Schematic of a part of the earth where the local earth magnetic field $m^{\text{n}}$ makes an angle $\delta$ with the horizontal plane. This angle is called the dip angle.} 
  \label{fig:models-magneticField}
\end{figure}
\footnotetext{Adapted version of `Dipolar magnetic field'  by Cyril Langlois available at \url{http://texample.net} under CC BY 2.5 (\url{http://creativecommons.org/licenses/by/2.5}).}
}

In practice, the actual magnetic field can differ significantly from the earth magnetic field. In indoor environments, for instance, presence of magnetic material in the structure of buildings and in furniture influences the magnetic field that is measured by the magnetometer. Furthermore, the magnetic field is affected in applications where the magnetometer is mounted in \eg a vehicle, train or on a robot. In case the magnetic material is rigidly attached to the sensor, the magnetometer can be calibrated for its presence \citep{kokS:2016,vasconcelosESOC:2011,renaudinAL:2010,salehiMB:2012}. The presence of magnetic material in the vicinity of the sensor that can not be calibrated for is of major concern for practical applications. Because of this, there is a vast amount of literature on the topic, see \eg \cite{callmer:2013,ligorioS:2016,roetenbergLBV:2005}. 

Note that when using magnetometers for orientation estimation, the presence of magnetic material is typically considered to be an undesired disturbance. However, the presence of magnetic material can also be considered to be a property which can be exploited. This is done in approaches which use the magnetic field as a source of position information \citep{solinKWSS:2015,haverinenK:2009,robertsonFADJPKLB:2013}.

\paragraph{Position information}
Position information can be obtained from for instance \gls{gnss} or \gls{uwb} measurements. In this tutorial, we will consider a very basic measurement model where the sensors directly measure the position as
\begin{align}
\label{eq:models-posMeasModel}
y_{\text{p},t} &= p_t^\text{n} + e_{\text{p},t}, 
\end{align}
with $e_{\text{p},t} \sim \mathcal{N}(0,\Sigma_\text{p})$. 

Many sensors do not measure the position directly. Their measurements can, however, be pre-processed to obtain position estimates and their corresponding covariances~\citep{gustafssonG:2005}. For example, time of arrival measurements can be pre-processed using multilateration techniques. Measurements of this type often contain a significant amount of outliers. The reason is that the signals can be delayed due to multipath or \gls{nlos} conditions. Possible solutions deal with this by doing outlier rejection, by using robust algorithms, see \eg \cite{zoubirKCM:2012}, or by modeling the noise distribution as a non-Gaussian distribution, see \eg \cite{kokHS:2015,gustafssonG:2005,nurminenAPG:2015}. We will discuss an example of this in more detail in~\Sectionref{sec:appl-uwb}.

\section{Choosing the state and modeling its dynamics}
\label{sec:models-stateDynamics}
The fundamental continuous-time relations that form the basis of our dynamic models are the fact that the position $p^\text{n}$, velocity $v^\text{n}_\text{n}$ and acceleration $a^\text{n}_\text{nn}$ are related as 
\begin{align}
v^\text{n}_\text{n} = \left. \tfrac{\tdiff p^\text{n}}{\tdiff t} \right|_\text{n}, \qquad
a^\text{n}_\text{nn} = \left. \tfrac{\tdiff v^\text{n}}{\tdiff t} \right|_\text{n},
\label{eq:models-contTimePosVel}
\end{align}
and that the orientation and angular velocity $\omega_{\text{nb},t}^\text{b}$ are related as
\begin{align}
\tfrac{\tdiff q^\text{nb}}{\tdiff t} = q^\text{nb} \odot \tfrac{1}{2}\bar{\omega}_{\text{nb}}^\text{b}, \qquad
\tfrac{\tdiff R^\text{nb}}{\tdiff t} = R^\text{nb} [\omega_{\text{nb}}^\text{b} \times],
\label{eq:models-contTimeOri}
\end{align}
depending on orientation parametrization. For a derivation of~\eqref{eq:models-contTimeOri}, see \eg \cite{hol:2011}. Using an Euler discretization of~\eqref{eq:models-contTimePosVel} assuming that the acceleration is constant between samples, the dynamics of the position and velocity can be expressed in terms of the acceleration as
\begin{subequations}
\label{eq:models-dynModelPosVel}
\begin{align}
p^\text{n}_{t+1} &=  p_t^\text{n} + T v_{\text{n},t}^\text{n} + \tfrac{T^2}{2} a^\text{n}_{\text{nn},t}, \\
v^\text{n}_{\text{n},t+1} &= v_{\text{n},t}^\text{n} + T a_{\text{nn},t}^\text{n}, 
\end{align}
\end{subequations}
where $T$ is the time between two samples. Similarly, the dynamics of the orientation can be expressed in terms of unit quaternions or rotation matrices as
\begin{subequations}
\label{eq:models-dynModelOri}
\begin{align}
q^\text{nb}_{t+1} &= q_t^\text{nb} \odot \expq \left( \tfrac{T}{2} \omega^\text{b}_{\text{nb},t} \right), \\
R^\text{nb}_{t+1} &= R_t^\text{nb} \expR \left( T \omega^\text{b}_{\text{nb},t} \right).
\end{align}
\end{subequations}

Dynamic models describe how the state changes over time. For the problem of position and orientation estimation using inertial sensors, there are two commonly used modeling alternatives for the dynamics~\citep{gustafsson:2012}. In the first, the state vector $x_t$ is chosen to consists of  
\begin{align}
x_t = \begin{pmatrix} \left(p^\text{n}_t \right)^\Transp & \left(v^\text{n}_{\text{n},t} \right)^\Transp & \left(a^\text{n}_{\text{nn,t}} \right)^\Transp & \left(q^\text{nb}_t \right)^\Transp & \left(\omega^\text{b}_{\text{nb},t} \right)^\Transp \end{pmatrix}^\Transp.
\end{align}
The change in position, velocity and orientation states can then be described in terms of the velocity, acceleration and angular velocity states, respectively. The dynamics of the acceleration and the angular velocity can be described in terms of a \emph{motion model}. Examples of motion models that can be used are a constant acceleration model, which assumes that the dynamics of the acceleration can be described as 
\begin{align}
\label{eq:models-constAcc}
a^\text{n}_{\text{nn},t+1} = a^\text{n}_{\text{nn},t} + w_{\text{a},t},
\end{align}
with $w_{\text{a},t} \sim \mathcal{N}(0,\Sigma_\text{w,a} )$, and a constant angular velocity model, which describes the dynamics of the angular velocity as
\begin{align}
\label{eq:models-constAngVel}
\omega^\text{b}_{\text{nb},t+1} = \omega^\text{b}_{\text{nb},t} + w_{\omega,t},
\end{align}
with $w_{\omega,t} \sim \mathcal{N}(0,\Sigma_{\text{w},\omega})$. The process noise terms $w_{\text{a},t}$ and $w_{\omega,t}$ model the assumptions on how constant the acceleration and angular velocity actually are. 

Alternatively, the state vector $x_t$ can be chosen as 
\begin{align}
x_t = \begin{pmatrix} \left(p^\text{n}_t \right)^\Transp & \left(v^\text{n}_{\text{n},t} \right)^\Transp & \left(q^\text{nb}_t \right)^\Transp \end{pmatrix}^\Transp.
\end{align}
To describe the dynamics of the states, the inertial measurements can then be used as an \emph{input} to the dynamic equation~\eqref{eq:models-dynamics}. Hence, the change in position, velocity and orientation is directly modeled in terms of the inertial measurements. In this case, expressions for $a^\text{n}_{\text{nn},t}$ and $\omega^\text{b}_{\text{nb},t}$ in~\eqref{eq:models-dynModelPosVel} and~\eqref{eq:models-dynModelOri} are obtained from the accelerometer measurement model and the gyroscope measurement model, see \Sectionref{sec:models-measModels}. The process noise can explicitly be modeled in terms of the accelerometer measurement noise $e_{\text{a},t}$ and the gyroscope measurement noise $e_{\omega,t}$. 

The benefit of using a motion model for the state dynamics is that knowledge about the motion of the sensor can be included in this model. However, it comes at the expense of having a larger state vector. The benefit of using the inertial measurements as an input to the dynamics is that the process noise has the intuitive interpretation of representing the inertial measurement noise. Hence, the latter approach is often used for applications where it is difficult to obtain sensible motion models. Another difference between the two approaches is that changes in the acceleration and angular velocity will have a slightly faster effect on the state when the inertial measurements are used as an input to the dynamics. The reason for this is that the constant acceleration and angular velocity models delay the effects of changes in the dynamics. Because of this, their estimates typically look slightly more smooth. 

\section{Models for the prior}
\label{sec:models-prior}
Looking back at \Sectionref{sec:models-probModeling}, to solve the smoothing and filtering problems~\eqref{eq:models-smoothingProbs} and~\eqref{eq:models-filteringProbs}, we have now discussed different models for the dynamics $p(x_t \mid x_{t-1},\theta)$ in \Sectionref{sec:models-stateDynamics} and for the measurements $p(y_t \mid x_{t},\theta)$ in \Sectionref{sec:models-measModels}. The remaining distributions to be defined are $p(x_1 \mid \theta)$ and $p(\theta)$. Note that we typically assume that $x_1$ is independent of $\theta$. Hence, in this section we focus on the prior distributions $p(x_1)$ and $p(\theta)$.

In many cases, there is fairly little prior information available about the parameters $\theta$. It is, however, often possible to indicate reasonable values for the parameters. For example, it is reasonable to assume that the gyroscope bias is fairly small but can be both positive and negative. For instance, for the data presented in \Exampleref{ex:sensors-inertialMeasurements}, the average bias over the 55-minute data set was $\begin{pmatrix} 35.67 & 54.13 & -1.07 \end{pmatrix}^\Transp \cdot 10^{-4}~\radianpersecond$. If we would assume that in $68\%$ of the cases, the bias is within the bounds $-\sigma_{\delta_\omega} \leq \delta_{\omega,t}^\text{b} \leq \sigma_{\delta_\omega}$ with $\sigma_{\delta_\omega} = 5 \cdot 10^{-3}$, a reasonable prior would be
\begin{align}
\delta_{\omega,t}^\text{b} \sim \mathcal{N}(0,\sigma^2_{\delta_\omega} \mathcal{I}_3).
\end{align}

For the prior $p(x_1)$, it is typically possible to get a reasonable estimate from data. For the position and velocity, this can be modeled as
\begin{subequations}
\begin{align}
p^\text{n}_1 &= \initEst{p}_1^\text{n} + e_{\text{p,i}}, \quad & e_{\text{p,i}} &\sim \mathcal{N}(0, \sigma^2_{\text{p,i}} \mathcal{I}_3), \\
v^\text{n}_1 &= \initEst{v}_1^\text{n} + e_{\text{v,i}}, \quad & e_{\text{v,i}} &\sim \mathcal{N}(0, \sigma^2_{\text{v,i}} \mathcal{I}_3).
\end{align}
\end{subequations}
Here, the estimate $\initEst{p}_1^\text{n}$ can for instance be determined based on the first position measurement. In that case, the uncertainty $\sigma_{\text{p,i}}$ can also be chosen equal to the uncertainty of the position measurements. In case no additional information is available, the estimates $\initEst{p}_1^\text{n}$ and $\initEst{v}_1^\text{n}$ can be set to zero with an appropriate standard deviation instead.

A commonly used method to determine the initial orientation is to use the first accelerometer and magnetometer samples. This method is based on the fact that given two (or more) linearly independent vectors in two coordinate frames, the rotation between the two coordinate frames can be determined. The implicit assumption is that the accelerometer only measures the gravity vector and the magnetometer only measures the local magnetic field. Hence, the four vectors are given by the measurements $y_{\text{a},t}$ and $y_{\text{m},t}$, the local gravity vector $g^\text{n}$ and the local magnetic field $m^\text{n}$. These vectors are linearly independent except when the measurements are obtained on the magnetic north or south poles where the dip angle is $\delta = 0$ and the magnetic field does not contain any heading information. 

The accelerometer provides information about the sensor's inclination. Heading information is provided by the magnetometer. However, at all locations except on the equator, the magnetometer also provides information about the inclination due to its non-zero vertical component, see \eqref{eq:models-localMagField}. In practice, the accelerometer typically provides more accurate inclination information. Hence, we choose to use the magnetometer only to provide heading information by projecting the magnetic field and the magnetometer measurement on the horizontal plane. Furthermore, we normalize the vectors. Because of this, an adapted model uses the four normalized vectors
\begin{subequations}
\label{eq:models-oriAccMagVectors}
\begin{align}
\hat{g}^\text{n} &= \begin{pmatrix} 0 & 0 & 1 \end{pmatrix}^\Transp,  \qquad
\hat{g}^\text{b} = \tfrac{y_{\text{a},1}}{\| y_{\text{a},1} \|_2}, \\
\hat{m}^\text{n} &= \begin{pmatrix} 1 & 0 & 0 \end{pmatrix}^\Transp, \qquad 
\hat{m}^\text{b} = \hat{g}^\text{b} \times \left( \tfrac{y_{\text{m},1}}{\| y_{\text{m},1} \|_2}  \times \hat{g}^\text{b}\right).
\end{align}
\end{subequations}
A number of algorithms are available to estimate the orientation from these vectors. Well-known examples are the TRIAD algorithm, the QUEST algorithm, see \eg \cite{shusterO:1981}, and the method presented in~\cite{horn:1987}. For our problem at hand, these methods give equivalent results, even though they use slightly different solution strategies. Generally speaking, they solve the problem of determining the rotation $q^\text{nb}$ from 
\begin{align}
\label{eq:models-oriAccMag}
&\argmin_{q^\text{nb}} &&\| \bar{\hat{g}}^\text{n} - q^\text{nb} \odot \bar{\hat{g}}^\text{b} \odot q^\text{bn} \|^2_2 +  
\| \bar{\hat{m}}^\text{n} - q^\text{nb} \odot \bar{\hat{m}}^\text{b} \odot q^\text{bn} \|^2_2 \nonumber \\
& \st &&\| q^\text{nb} \|_2 = 1,
\end{align}
Recall from~\eqref{eq:models-rotatex_q} that $q^\text{nb} \odot \bar{x}^\text{b} \odot q^\text{bn}$ is the rotation of the vector $x^\text{b}$ to the $n$-frame. The optimization problem~\eqref{eq:models-oriAccMag} therefore determines the orientation $q^\text{nb}$ that minimizes the distance between the normalized magnetic field and gravity vectors measured in the first sample and the normalized magnetic field and gravity vectors in the navigation frame. These four vectors were defined in~\eqref{eq:models-oriAccMagVectors}.

Defining 
\begin{align}
A = - \left( \bar{\hat{g}}^\text{n} \right)^\leftMult \left( \bar{\hat{g}}^\text{b} \right)^\rightMult - \left( \bar{\hat{m}}^\text{n} \right)^\leftMult \left( \bar{\hat{m}}^\text{b} \right)^\rightMult,
\end{align}
where the left and right quaternion multiplications are defined in~\eqref{eq:models-leftRightQuatMult}, \eqref{eq:models-oriAccMag} can equivalently be written as 
\begin{align}
\initEst{q}^\text{nb}_1 = &\argmin_{q^\text{nb}} \quad \left( q^\text{nb} \right)^\Transp A q^\text{nb} \nonumber \\
&\st \quad \| q^\text{nb} \|_2 = 1.
\label{eq:models-oriAccMagReform}
\end{align}
For a derivation, see \cite{hol:2011}. The solution to this problem is given by the eigenvector corresponding to the largest eigenvalue of $A$. Note that although this method can be used to compute the orientation from any two linearly independent vectors in two coordinate frames, we only use it to compute a prior on the orientation. 

Based on the estimate $\initEst{q}_1^\text{nb}$ from~\eqref{eq:models-oriAccMagReform}, we can model the orientation at time $t = 1$ in terms of an orientation deviation
\begin{subequations}
\begin{align}
q_1^\text{nb} &= \expq \left( \tfrac{e_{\oriError,\text{i}}}{2} \right) \odot \initEst{q}_1^\text{nb},  & e_{\oriError,\text{i}} &\sim \mathcal{N}(0, \Sigma_{\oriError,\text{i}}), \\
\intertext{or in terms of a quaternion as}
q_1^\text{nb} &= \initEst{q}_1^\text{nb} + e_\text{q,i},  & e_\text{q,i} &\sim \mathcal{N}(0, \Sigma_\text{q,i}) .
\end{align}
\end{subequations}
Explicit formulations for the covariance of the orientation estimates from the TRIAD and QUEST algorithms are discussed by \cite{shuster:2006}. In practice, however, the accuracy of the estimates from~\eqref{eq:models-oriAccMagReform} highly depends on the validity of the model assumptions, \ie on whether the sensor is indeed far from magnetic material and whether the linear acceleration is indeed zero. Because this has such a significant influence on the quality of the estimates, we choose $\Sigma_{\oriError,\text{i}}$ and $\Sigma_\text{q,i}$ somewhat conservatively. Modeling that in $68 \%$ of the cases the orientation error is less than $20^\circ$, 
\begin{subequations}
\begin{align}
\Sigma_{\oriError,\text{i}} &= \sigma_{\oriError, \text{i}}^2 \mathcal{I}_3, \qquad \qquad \sigma_{\oriError,\text{i}} = \tfrac{20}{180} \pi, \\
\Sigma_{\text{q,i}} &= \tfrac{1}{4} \left( \initEst{q}_1^\text{nb} \right)^\rightMult \tfrac{\diff \expq (e_{\oriError,\text{i}})}{\diff e_{\oriError,\text{i}}} \Sigma_{\oriError,\text{i}} \left( \tfrac{\diff \expq (e_{\oriError,\text{i}})}{\diff e_{\oriError,\text{i}}} \right)^\Transp \left( \initEst{q}_1^\text{bn} \right)^\rightMult,
\end{align}
\end{subequations}
where we use the fact that $( q^\rightMult )^\Transp = (q^\conj)^\rightMult$. Note that the covariance $\Sigma_{\text{q,i}}$ is defined in terms of the covariance $\Sigma_{\oriError,\text{i}}$ to allow for explicit comparison between different algorithms in \Chapterref{cha:orientationEstimation}.

\section{Resulting probabilistic models}
\label{sec:models-resultingProbModel}
The information from the previous sections can now be combined into one model which will be used in the algorithms in \Chapterref{cha:orientationEstimation}. In this section, we describe our modeling choices for the pose estimation problem and for the orientation estimation problem. 

We assume that the sensor does not travel over significant distances as compared to the size of the earth and hence keep the navigation frame $n$ fixed with respect to the earth frame $e$. Furthermore, we assume that the magnitude of the earth rotation and of the Coriolis acceleration are negligible. Our gyroscope and accelerometer models are hence given by 
\begin{subequations}
\begin{align}
y_{\text{a},t} &= R^{\text{bn}}_t  ( a_{\text{nn}}^\text{n} - g^\text{n} ) + \delta_{\text{a},t}^\text{b} + e_{\text{a},t}^\text{b}, \label{eq:models-accMeasModelRes} \\
y_{\omega,t} &= \omega_{\text{nb},t}^\text{b} + \delta_{\omega,t}^\text{b} + e_{\omega,t}^\text{b}. \label{eq:models-gyrMeasModelRes}
\end{align}
\end{subequations}
In the remainder, for notational convenience we drop the subscripts $n$ which indicate in which frame the differentiation is performed, see \Sectionref{sec:sensors-specForce}, and use the shorthand notation $a^\text{n}$ for $a_{\text{nn}}^\text{n}$. Furthermore, we will denote $\omega_{\text{nb}}^\text{b}$ simply by $\omega$ and omit the superscript $b$ on the noise terms $e_{\text{a},t}$ and $e_{\omega,t}$ and the bias terms $\delta_{\text{a},t}$ and $\delta_{\omega,t}$. We assume that the inertial measurement noise is given by
\begin{align}
e_{\text{a},t} \sim \mathcal{N}(0, \sigma_\text{a}^2 \, \mathcal{I}_3 ), \qquad e_{\omega,t} \sim \mathcal{N}(0, \sigma_\omega^2 \, \mathcal{I}_3 ),
\label{eq:models-assumptionNoiseDiag}
\end{align}
\ie we assume that the three sensor axes are independent and have the same noise levels. 

\subsection{Pose estimation}
For pose estimation, we model the accelerometer and gyroscope measurements as inputs to the dynamics. Hence, the state vector consists of the position $p^\text{n}_{t}$, the velocity $v^\text{n}_{t}$ and a parametrization of the orientation. We use the inertial measurements in combination with position measurements to estimate the pose. 

Using the accelerometer measurement model \eqref{eq:models-accMeasModelRes} in~\eqref{eq:models-dynModelPosVel}, the dynamics of the position and velocity is given by 
\begin{subequations}
\begin{align}
p_{t+1}^\text{n} &= p_t^\text{n} + T v_t^\text{n} + \tfrac{T^2}{2} \left( R^\text{nb}_t (y_{\text{a},t} - \delta_{\text{a},t}) + g^\text{n} + e_{\text{a},t} \right), \\
v_{t+1}^\text{n} &= v_t^\text{n} + T \left( R^\text{nb}_t (y_{\text{a},t} - \delta_{\text{a},t}) + g^\text{n} + e_{\text{a},t} \right),
\end{align}
\end{subequations}
where without loss of generality, we switch the sign on the noise. Note that the noise term $e_{\text{a},t}$ should be rotated to the navigation frame $n$ by multiplying it with the rotation matrix $R^\text{nb}_t$. However, because of the assumption~\eqref{eq:models-assumptionNoiseDiag}, the rotation matrix can be omitted without loss of generality. The dynamics of the orientation parametrized using quaternions is given by
\begin{align}
q^\text{nb}_{t+1} &= q_t^\text{nb} \odot \expq \left( \tfrac{T}{2} (y_{\omega,t} - \delta_{\omega,t} - e_{\omega,t} ) \right).
\end{align}
Equivalent dynamic models can be obtained for the other parametrizations. 

The position measurements are modeled as in~\eqref{eq:models-posMeasModel}. In summary, this leads to the following state space model for pose estimation
\begin{subequations}
\label{eq:models-ssPose}
\begin{align}
\begin{pmatrix} 
p^\text{n}_{t+1} \\
v^\text{n}_{t+1} \\ 
q^\text{nb}_{t+1} 
\end{pmatrix}
&= \begin{pmatrix} 
p_t^\text{n} + T v_t^\text{n} + \tfrac{T^2}{2} \left( R^\text{nb}_t (y_{\text{a},t} - \delta_{\text{a},t}) + g^\text{n} + e_{\text{p,a},t} \right) \\
v_t^\text{n} + T \left( R^\text{nb}_t (y_{\text{a},t}  - \delta_{\text{a},t}) + g^\text{n} + e_{\text{v,a},t} \right) \\
q_t^\text{nb} \odot \expq \left( \tfrac{T}{2} (y_{\omega,t} - \delta_{\omega,t} - e_{\omega,t} ) \right)
\end{pmatrix}, \label{eq:models-ssPose-dyn} \\
y_{\text{p},t} &= p_t^\text{n} + e_{\text{p},t}, \label{eq:models-ssPose-meas}
\intertext{where}
e_{\text{p,a},t} &\sim {\mathrlap{\mathcal{N}(0, \Sigma_\text{a} ),}\phantom{\bar{q}_1^\text{nb} \odot \expq ( \tfrac{e_{\oriError,\text{i}}}{2} )}} \qquad \, e_{\text{v,a},t} \sim \mathcal{N}(0, \Sigma_\text{a} ), \\
e_{\text{p},t} &\sim {\mathrlap{\mathcal{N}(0, \Sigma_\text{p} ),}\phantom{\bar{q}_1^\text{nb} \odot \expq ( \tfrac{e_{\oriError,\text{i}}}{2} ), \, \qquad}} \, e_{\omega,t} \sim \mathcal{N}(0, \Sigma_\omega ), 
\intertext{with $\Sigma_\text{a} = \sigma_\text{a}^2 \, \mathcal{I}_3$ and $\Sigma_\omega = \sigma_\omega^2 \, \mathcal{I}_3$. Note that we model the process noise on the position and velocity states in terms of the accelerometer noise. However, we do not enforce these to have the same noise realization. Hence, we use the notation $e_{\text{p,a},t}$ and $e_{\text{v,a},t}$ for the two process noise terms. The covariance of both is equal to the covariance of the accelerometer noise. The initial position is assumed to be given by the first position measurement as}
p^\text{n}_1 &= {\mathrlap{y_{\text{p},1} + e_{\text{p,1}},}\phantom{\bar{q}_1^\text{nb} \odot \expq ( \tfrac{e_{\oriError,\text{i}}}{2} ),}} \, \qquad \, e_{\text{p,1}} \sim \mathcal{N}(0, \Sigma_\text{p,i}), 
\intertext{while the initial velocity is assumed to be approximately zero as}
v^\text{n}_1 &= {\mathrlap{e_{\text{v,i}},}\phantom{\bar{q}_1^\text{nb} \odot \expq ( \tfrac{e_{\oriError,\text{i}}}{2} ),}} \, \qquad \, \, e_{\text{v,i}} \sim \mathcal{N}(0, \Sigma_{\text{v,i}}).
\intertext{The orientation at time $t = 1$ is given by the QUEST algorithm described in \Sectionref{sec:models-prior}, parametrized in terms of quaternions or rotation vectors as }
q_1^\text{nb} &= \expq \left( \tfrac{e_{\oriError,\text{i}}}{2} \right) \odot \initEst{q}_1^\text{nb}, \qquad e_{\oriError,\text{i}} \sim \mathcal{N}(0, \Sigma_{\oriError,\text{i}}), \label{eq:models-ssPose-priorOriDev} \\
q_1^\text{nb} &= { \mathrlap{\initEst{q}_1^\text{nb} + e_\text{q,i},}\phantom{\bar{q}_1^\text{nb} \odot \expq ( \tfrac{e_{\oriError,\text{i}}}{2} ),} } \qquad \, \, e_\text{q,i} \sim \mathcal{N}(0, \Sigma_\text{q,i}), \label{eq:models-ssPose-priorQuat}
\end{align}
\end{subequations}
where the initial orientation uncertainty is given in terms of a standard deviation of $20^\circ$.

In \Chapterref{cha:orientationEstimation} we assume that the inertial measurement are properly calibrated. Hence, we assume that their biases $\delta_{\text{a},t}$ and $\delta_{\omega,t}$ are zero. Calibration is the topic of \Chapterref{cha:calibration} where we will also introduce possible extensions of the state space model in which the bias terms are included either as states or as unknown parameters.

\subsection{Orientation estimation}
For orientation estimation, the state vector only consists of a parametrization of the orientation. We use the inertial sensors in combination with the magnetometer measurements to estimate the orientation. The magnetometer measurements are modeled as in~\eqref{eq:models-magMeasModel}. Instead of using the accelerometer measurements model~\eqref{eq:models-accMeasModelRes}, we use the model~\eqref{eq:models-accMeasModelZeroAcc} where it is assumed that the linear acceleration is approximately zero. This leads to the following state space model for orientation estimation,
\begin{subequations}
\label{eq:models-ssOri}
\begin{align}
q^\text{nb}_{t+1} &= q_t^\text{nb} \odot \expq \left( \tfrac{T}{2} (y_{\omega,t} - \delta_{\omega} - e_{\omega,t} ) \right), \label{eq:models-ssOri-dyn} \\
y_{\text{a},t} &= - R^\text{bn}_t g^\text{n} + e_{\text{a},t} , \label{eq:models-ssOri-measAcc} \\
y_{\text{m},t} &= R^\text{bn}_t m^\text{n} + e_{\text{m},t}, \label{eq:models-ssOri-measMag}
\intertext{where~\eqref{eq:models-ssOri-dyn} describes the dynamics while~\eqref{eq:models-ssOri-measAcc} and~\eqref{eq:models-ssOri-measMag} describe the measurement models and} 
e_{\omega,t} &\sim \mathcal{N}(0,\Sigma_\omega), \qquad e_{\text{a},t} \sim \mathcal{N}(0, \Sigma_\text{a}), \qquad e_{\text{m},t} \sim \mathcal{N}(0,\Sigma_\text{m}),
\end{align}
\end{subequations}
with $\Sigma_\omega = \sigma_\omega^2 \, \mathcal{I}_3$ and $\Sigma_\text{a} = \sigma_\text{a}^2 \, \mathcal{I}_3$. The initial orientation is given by the QUEST algorithm described in \Sectionref{sec:models-prior} and is modeled as in~\eqref{eq:models-ssPose-priorOriDev} or~\eqref{eq:models-ssPose-priorQuat}. Also for orientation estimation, in \Chapterref{cha:orientationEstimation} we assume that the inertial measurements are properly calibrated. Hence, we assume that the bias $\delta_{\omega,t}$ is zero.

\chapter{Estimating Position and Orientation}
\label{cha:orientationEstimation} 
In this chapter we will focus on position and orientation estimation using the models~\eqref{eq:models-ssPose} and~\eqref{eq:models-ssOri} derived in \Chapterref{cha:models}. In \Sectionref{sec:oriEst-smoothingOpt}, we will first describe a method to solve the smoothing problem~\eqref{eq:models-smoothingProbs}. Subsequently, in \Sectionref{sec:oriEst-filteringOpt}--\Sectionref{sec:oriEst-compl}, we will derive different methods for solving the filtering problem~\eqref{eq:models-filteringProbs}. In each section, after a general introduction of the estimation method, we will illustrate the method by explicitly deriving algorithms to estimate the orientation using the state space model~\eqref{eq:models-ssOri}. The orientation estimation problem also illustrates the most important parts of the pose estimation problem, since most complexities lie in the parametrization of the orientation and in the nonlinear nature of  the orientation. In \Sectionref{sec:oriEst-orientationEstimation}, we show some characteristics of the different algorithms for the orientation estimation problem. In \Sectionref{sec:oriEst-poseEstimation}, we will discuss how the algorithms for orientation estimation can be extended to also estimate the position. Throughout this section, we assume that the sensors are calibrated, \ie we assume that we do not have any unknown parameters $\theta$ in our models. Because of this, the models that we use are the most basic models that can be used for position and orientation estimation using inertial sensors. 

\section{Smoothing in an optimization framework}
\label{sec:oriEst-smoothingOpt}
Perhaps the most intuitive way to solve the smoothing problem is by posing it as an optimization problem, where a \emph{\gls{map}} estimate is obtained as
\begin{align}
\hat{x}_{1:N} &= \argmax_{x_{1:N}} p(x_{1:N} \mid y_{1:N} ) \label{eq:oriEst-optLik} \nonumber \\
&= \argmax_{x_{1:N}} p(x_1) \prod_{t = 2}^N p(x_{t} \mid x_{t-1} ) p(y_t \mid x_t ). 
\end{align}
Here, we use the notation in terms of probability distributions as introduced in \Sectionref{sec:models-probModeling} and model the measurements and the state dynamics as described in~\Sectionref{sec:models-measModels} and~\Sectionref{sec:models-stateDynamics}, respectively. Furthermore, we assume that a prior on the initial state is obtained using the measurements at $t = 1$ as described in \Sectionref{sec:models-prior}. Because of this, the measurement model $p(y_1 \mid x_1)$ from~\eqref{eq:models-smoothingProbs} is explicitly omitted in~\eqref{eq:oriEst-optLik}. Note that in practice, we typically minimize $-\log p(x_{1:N} \mid y_{1:N})$ instead of maximizing $p(x_{1:N} \mid y_{1:N})$ itself, resulting in the optimization problem
\begin{align}
\argmin_{x_{1:N}} - \log p(x_1) - \sum_{t = 2}^N \log p(x_{t} \mid x_{t-1} ) - \sum_{t = 2}^N \log p(y_t \mid x_t ).\label{eq:oriEst-optNegLogLik}
\end{align}

There are various ways to solve problems of this kind, for instance particle smoothers \citep{lindstenS:2013}, an extended \gls{rts} smoother \citep{sarkka:2013} and optimization methods, see \eg \cite{nocedalW:2006,mattingleyB:2010}. The latter approach is closely related to iterated Kalman smoothers~\citep{bell:1994,jazwinski:1970}. We will solve the problem using an optimization method. Compared to extended \gls{rts} smoothers, optimization methods allow for more flexibility in the models that are being used. For instance, additional information outside of the standard state space model can straightforwardly be included. Optimization approaches are typically computationally heavier than extended \gls{rts} smoothers but less heavy than particle smoothers. The latter are capable of capturing the whole distribution, which is a clear advantage when the distributions are multi-modal. Optimization instead gives a point estimate and an associated measure of uncertainty. This is typically sufficient for position and orientation estimation using inertial sensors.

\subsection{Gauss-Newton optimization}
\label{sec:oriEst-GNopt}
To obtain a smoothing estimate of the position and orientation using optimization, we first recognize that for our models~\eqref{eq:models-ssPose} and~\eqref{eq:models-ssOri}, all probability distributions in~\eqref{eq:oriEst-optNegLogLik} are Gaussian. Let us therefore consider a slightly more general problem where the objective function consists of the product of Gaussian probability functions $p(e_i(x_{1:N}))$, $i = 1, \hdots, M$. Hence, the optimization problem can be written as
\begin{align}
\hat{x}_{1:N} = \argmin_{x_{1:N}} - \sum_{i = 1}^M \log p \left(e_i (x_{1:N}) \right). \label{eq:oriEst-optNegLogLikGaussian}
\end{align}
The probability distribution of $e_i(x)$ is given by
\begin{align}
p \left(e_i(x_{1:N}) \right) = \tfrac{1}{\sqrt{ (2 \pi)^{n_e} \det \Sigma_i}} \exp \left( - \tfrac{1}{2} e_i^\Transp(x_{1:N}) \Sigma_i^{-1} e_i(x_{1:N}) \right).
\end{align}
Omitting the terms independent of $x_{1:N}$, the optimization problem~\eqref{eq:oriEst-optNegLogLikGaussian} reduces to
\begin{align}
\label{eq:oriEst-generalNLS}
\hat x_{1:N} = \argmin_{x_{1:N}} \tfrac{1}{2} \sum_{i = 1}^M \| e_i(x_{1:N}) \|_{\Sigma_i^{-1}}^2, 
\end{align}
with $\| e_i(x_{1:N}) \|_{\Sigma_i^{-1}}^2 = e^\Transp_i(x_{1:N}) \Sigma_i^{-1} e_i(x_{1:N})$. The function that is being minimized in optimization problems, is often referred to as the \emph{objective function}.

The solution to~\eqref{eq:oriEst-generalNLS} can be found by studying the shape of the objective function as a function of $x_{1:N}$. This can be characterized in terms of the \textit{gradient} $\mathcal{G}(x_{1:N})$ and \textit{Hessian} $\mathcal{H}(x_{1:N})$, which provide information about the slope and curvature of the function, respectively. Defining 
\begin{align*}
e^\Transp_i(x_{1:N}) \Sigma_i^{-1} e_i(x_{1:N}) = \varepsilon_i^\Transp \varepsilon_i, \qquad \varepsilon_i = \Sigma_i^{-1/2} e_i(x_{1:N}),
\end{align*}
and the stacked variables 
\begin{align*}
\varepsilon = \begin{pmatrix} \varepsilon_1^\Transp & \cdots & \varepsilon^\Transp_M \end{pmatrix}^\Transp,
\end{align*}
the gradient and the Hessian are given by
\begin{subequations}
\label{eq:oriEst-gradHess}
\begin{align}
\mathcal{G}(x_{1:N}) &= \sum_{i = 1}^{M} \left( \tfrac{\diff \varepsilon_i}{\diff x_{1:N}} \right)^\Transp \varepsilon_i= \mathcal{J}^\Transp(x_{1:N}) \varepsilon, \label{eq:oriEst-gradient} \\
\mathcal{H}(x_{1:N}) &= \sum_{i = 1}^{M} \left( \left( \tfrac{\diff \varepsilon_i}{\diff x_{1:N}} \right)^\Transp \tfrac{\diff \varepsilon_i}{\diff x_{1:N}} + \varepsilon_i^\Transp \tfrac{\diff^2 \varepsilon_i}{\diff x_{1:N}^2} \right) \nonumber \\
&= \mathcal{J}^\Transp(x_{1:N}) \mathcal{J}(x_{1:N}) + \sum_{i = 1}^{M} \varepsilon_i^\Transp \tfrac{\diff^2 \varepsilon_i}{\diff x_{1:N}^2}. \label{eq:oriEst-hessian}
\end{align}
\end{subequations}
Note that for notational convenience, we have omitted the explicit dependence of~$\varepsilon$ on $x_{1:N}$. In~\eqref{eq:oriEst-gradHess}, we introduced the notation $\mathcal{J}(x_{1:N})$, which is the \emph{Jacobian} of the vector $\varepsilon$ with respect to $x_{1:N}$ as
\begin{align}
\mathcal{J}(x_{1:N}) = \begin{pmatrix} \tfrac{\diff \varepsilon_1}{\diff x_1} & \hdots & \tfrac{\diff \varepsilon_1}{\diff x_N} \\
\vdots & & \vdots \\
\tfrac{\diff \varepsilon_{M n_\varepsilon}}{\diff x_1} & \hdots & \tfrac{\diff \varepsilon_{ M n_\varepsilon}}{\diff x_N}
\end{pmatrix},
\end{align}
where $n_\varepsilon$ is the length of the vector $\varepsilon_i$. Instead of computing the true Hessian~\eqref{eq:oriEst-hessian}, we compute an approximation of it \citep{nocedalW:2006}, given by
\begin{align}
\label{eq:oriEst-approxHessian}
\mathcal{\hat{H}}(x_{1:N}) = \mathcal{J}^\Transp(x_{1:N}) \mathcal{J}(x_{1:N}).
\end{align}
This has the benefit of not having to compute second derivatives, at the same time as it guarantees that the Hessian is positive semidefinite. The downside of using~\eqref{eq:oriEst-approxHessian} is that it introduces an approximation.

The gradient and the (approximate) Hessian can be used to find the minimum of the objective function. For our models~\eqref{eq:models-ssPose} and~\eqref{eq:models-ssOri}, in which the functions $e_i(x_{1:N})$ are nonlinear, an estimate $\hat{x}_{1:N}$ can iteratively be computed as
\begin{align}
\label{eq:oriEst-GNiteration}
\hat x_{1:N}^{(k+1)} = \hat x_{1:N}^{(k)} - \beta^{(k)} \left( \mathcal{\hat{H}}(\hat x_{1:N}^{(k)}) \right)^{-1} \mathcal{G}(\hat x_{1:N}^{(k)}),
\end{align}
where $k$ denotes the iteration number. The \emph{step length} $\beta^{(k)}$ is computed for instance using a backtracking line search \citep{nocedalW:2006,boydV:2004}. The \emph{search direction} is computed as $\left( \mathcal{\hat{H}}(\hat x_{1:N}^{(k)}) \right)^{-1} \mathcal{G}(\hat x_{1:N}^{(k)})$. Note that an initial point $\hat{x}_{1:N}^{(0)}$ needs to be chosen close enough to the desired minimum to ensure convergence to this minimum. 

\begin{figure}
	\centering
	\[ \begin{pmatrix}\,
 \includegraphics[scale = 1]{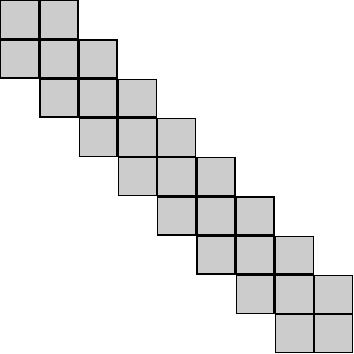} \end{pmatrix} \, \]
    	\caption{An illustration of the sparsity pattern that is present in the smoothing problem.}
	\label{fig:oriEst-sparsity}
\end{figure}

In case the functions $e_i(x_{1:N})$ would be linear, the problem~\eqref{eq:oriEst-generalNLS} would be a \emph{\gls{ls}} problem for which the update~\eqref{eq:oriEst-GNiteration} directly leads to the minimum of the objective function, irrespective of the initial point. In our case where the functions $e_i(x_{1:N})$ are nonlinear due to the nonlinear nature of the orientation, the problem~\eqref{eq:oriEst-generalNLS} is instead a \emph{\gls{nls}} problem. Each iteration in~\eqref{eq:oriEst-GNiteration} can be interpreted as solving a \gls{ls} problem around a linearization point. The linearization point is updated after each iteration, bringing it closer to the minimum of the objective function. Computing an estimate $\hat x_{1:N}$ by iterating~\eqref{eq:oriEst-GNiteration} until convergence, making use of the approximate Hessian from~\eqref{eq:oriEst-approxHessian}, is called \emph{Gauss-Newton optimization}.

Note that since the inertial sensors sample at high sampling rates, the length of the vector $x_{1:N}$ quickly becomes fairly large. For instance, for inertial sensors sampling at $100\hertz$ for $10$ seconds, $N = 1 \thinspace 000$ and the size of the (approximate) Hessian $\mathcal{\hat{H}}(x)$ is $1 \thinspace 000 n_x \times 1 \thinspace 000 n_x$, where $n_x$ is the length of the vector $x_t$. However, as can be seen in~\eqref{eq:oriEst-optNegLogLik}, the components of the objective function only depend on the current and next time steps $x_t$ and $x_{t+1}$. Hence, the structure of the Hessian~\eqref{eq:oriEst-approxHessian} is of the form given in \Figureref{fig:oriEst-sparsity}. There exist efficient algorithms to compute search directions for problems with this sparsity pattern, which can be exploited using sparse matrix packages, see \eg \cite{davis:2006}, or by using tools like dynamic programming and message passing~\citep{bertsekas:1995,golubvL:2013,saad:2003}.

\subsection{Smoothing estimates of the orientation using optimization}
In this section, we will illustrate the use of Gauss-Newton optimization to obtain smoothing estimates of the orientation. As discussed in the previous section, the crucial part is to identify the objective function and its Jacobian. From this, the gradient and approximate Hessian can be computed using~\eqref{eq:oriEst-gradient} and~\eqref{eq:oriEst-approxHessian}, which can be used to iteratively update the estimates using~\eqref{eq:oriEst-GNiteration}. 

Combining the general formulation of the smoothing problem~\eqref{eq:oriEst-optNegLogLik} and using the model for orientation estimation~\eqref{eq:models-ssOri}, the orientation smoothing problem is given by 
\begin{align}
\label{eq:oriEst-oriSmoothing}
\hat{x}_{1:N} = \argmin_{x_{1:N}} &\underbrace{\vphantom{\sum_{t = 2}^N} \| e_{\oriError,\text{i}} \|_{\Sigma_{\oriError,\text{i}}^{-1}}^2}_{\text{Prior}} + \underbrace{\sum_{t = 2}^N \| e_{\omega,t} \|_{\Sigma_\omega^{-1}}^2}_{\text{Dynamics}} + 
\underbrace{\sum_{t = 2}^N \left( \| e_{\text{a},t} \|_{\Sigma_\text{a}^{-1}}^2 + \| e_{\text{m},t} \|_{\Sigma_\text{m}^{-1}}^2\right)}_{\text{Measurement models}},
\end{align}
with
\begin{subequations}
\label{eq:oriEst-smoothingTerms}
\begin{align}
e_{\oriError,\text{i}} &=  2 \logq \left( q_1^\text{nb} \odot \initEst{q}_1^\text{bn} \right), \quad & e_{\oriError,\text{i}} &\sim \mathcal{N}(0,\Sigma_{\oriError,\text{i}}), \\
e_{\omega,t} &= \tfrac{2}{T} \logq \left( q_t^\text{bn} \odot q^\text{nb}_{t+1}\right) - y_{\omega,t}, \quad & e_{\omega,t} &\sim \mathcal{N}(0,\Sigma_\omega),  \\
e_{\text{a},t} &= y_{\text{a},t} + R^\text{bn}_t g^\text{n}, \quad & e_{\text{a},t} &\sim \mathcal{N}(0,\Sigma_\text{a}), \\
e_{\text{m},t} &= y_{\text{m},t} - R^\text{bn}_t m^\text{n}, \quad & e_{\text{m},t} &\sim \mathcal{N}(0,\Sigma_\text{m}). 
\end{align}
\end{subequations}

Convex equality constraints can straightforwardly be incorporated in optimization problems. However, using $N$ equality constraints to preserve the unit norm of the quaternions, severely complicates the optimization problem since these norm constraints are non-convex. Instead, we encode an orientation in terms of a linearization point parametrized as a unit quaternion $\tilde{q}^\text{nb}_t$ and an orientation deviation parametrized as a rotation vector $\oriError_t^\text{n}$ as discussed in \Sectionref{sec:models-linearization}. Hence, we model the orientation as 
\begin{align}
q^\text{nb}_t = \expq \left( \tfrac{\oriError_t^\text{n}}{2} \right) \odot \tilde{q}^\text{nb}_t.
\label{eq:oriEst-linSmoothing}
\end{align}
At each Gauss-Newton iteration~\eqref{eq:oriEst-GNiteration}, we estimate the state vector $\oriError_{1:N}^\text{n}$. Before starting the next iteration, the linearization points $\tilde{q}^\text{nb}_{1:N}$ are updated and the state vector $\oriError_{1:N}^\text{n}$ is reset to zero.

Using the notation introduced in \Sectionref{sec:models-linearization}, the objective function~\eqref{eq:oriEst-oriSmoothing} can be expressed in terms of the orientation deviation $\oriError_t^\text{n}$ by rewriting~\eqref{eq:oriEst-smoothingTerms} as
\begin{subequations}
\label{eq:oriEst-cost}
\begin{align}
e_{\oriError,\text{i}} &=  2 \logq \left( \expq (\tfrac{\eta_{1}^\text{n}}{2}) \odot \tilde{q}_1^\text{nb} \odot \initEst{q}_1^\text{bn} \right), \\
e_{\omega,t} &= \tfrac{2}{T} \logq \left( (\tilde{q}_t^\text{bn} \odot \expq( {\tfrac{\oriError_t^\text{n}}{2}}) )^\conj \odot \expq ({\tfrac{\oriError_{t+1}^\text{n}}{2}}) \odot \tilde{q}^\text{nb}_{t+1} \right) - y_{\omega,t}, \\
e_{\text{a},t} &= y_{\text{a},t} + \tilde{R}^\text{bn}_t \left( \expR( \oriError_t^\text{n} ) \right)^\Transp g^\text{n}, \label{eq:oriEst-costAcc} \\
e_{\text{m},t} &= y_{\text{m},t} - \tilde{R}^\text{bn}_t \left( \expR( \oriError_t^\text{n} ) \right)^\Transp m^\text{n}. \label{eq:oriEst-costMag}
\end{align}
\end{subequations}
Here, $\tilde{R}^\text{bn}_t$ is the rotation matrix representation of the linearization point $\tilde{q}^\text{bn}_t$. This leads to the following derivatives
\begin{subequations}
\label{eq:oriEst-der}
\begin{align}
\tfrac{\diff e_{\oriError,\text{i}} }{\diff \oriError_{1}^\text{n}}  &=  \tfrac{\diff \logq (q)}{\diff q} \left( \tilde{q}^\text{nb}_{1} \odot \initEst{q}_1^\text{bn}\right)^\rightMult \tfrac{\diff \expq (\oriError_{1}^\text{n})}{\diff \oriError_{1}^\text{n}}, \label{eq:oriEst-derInit} \\
\tfrac{\diff e_{\omega,t} }{\diff \oriError_{t+1}^\text{n}} &= \tfrac{1}{T} \tfrac{\diff \logq (q)}{\diff q} \left( \tilde{q}_t^\text{bn} \right)^\leftMult \left( \tilde{q}^\text{nb}_{t+1}\right)^\rightMult \tfrac{\diff \expq (\oriError_{t+1}^\text{n})}{\diff \oriError_{t+1}^\text{n}}, \label{eq:oriEst-derDynModelNow} \\
\tfrac{\diff e_{\omega,t} }{\diff \oriError_t^\text{n}} &= \tfrac{1}{T} \tfrac{\diff \logq (q)}{\diff q} \left( \tilde{q}_t^\text{bn} \right)^\leftMult \left( \tilde{q}^\text{nb}_{t+1} \right)^\rightMult \tfrac{\diff \left( \expq  (\oriError_{t}^\text{n}) \right)^\conj}{\diff \expq (\oriError_{t}^\text{n})} \tfrac{\diff \expq (\oriError_{t}^\text{n})}{\diff \oriError_{t}^\text{n}}, \label{eq:oriEst-derDynModelPrev} \\
\tfrac{\diff e_{\text{a},t} }{\diff \oriError_t^\text{n}}&\approx \tilde{R}^\text{bn}_t [g^\text{n} \times], \label{eq:oriEst-derAcc} \\
\tfrac{\diff e_{\text{m},t} }{\diff \oriError_t^\text{n}} &\approx - \tilde{R}^\text{bn}_t [m^\text{n} \times], \label{eq:oriEst-derMag}
\end{align}
\end{subequations}
where, using~\eqref{eq:models-logexpq-approx} and the definition of the quaternion conjugate~\eqref{eq:models-quatConj}, 
\begin{align}
\tfrac{\diff \logq (q)}{\diff q} &\approx \begin{pmatrix} 0_{1 \times 3} \\ \mathcal{I}_3 \end{pmatrix}^\Transp, \qquad
\tfrac{\diff \left( \expq (\oriError) \right)^\conj}{\diff \exp \oriError} = \begin{pmatrix} 1 & 0_{1 \times 3} \\ 0_{3 \times 1} & - \mathcal{I}_3 \end{pmatrix}, \nonumber \\
\tfrac{\diff \expq \oriError}{\diff \oriError} &\approx \begin{pmatrix} 0_{1 \times 3} \\ \mathcal{I}_3 \end{pmatrix}.
\end{align}
Using the approximate derivatives~\eqref{eq:oriEst-der}, the gradient and approximate Hessian can be computed for the Gauss-Newton iterations. The resulting solution is summarized in \Algorithmref{alg:oriEst-smoothingOpt}. One of the inputs to the algorithm is an initial estimate of the orientation $\tilde{q}^{\text{nb},(0)}_{1:N}$, which will aid the convergence to the desired minimum. There are (at least) two ways to obtain good initial estimates $\tilde{q}^{\text{nb},(0)}_{1:N}$. First, they can be obtained by direct integration of the gyroscope measurements. As discussed in \Sectionref{sec:intro-imusForPose}, these estimates will suffer from integration drift. However, they still form a good initial estimate for the optimization problem. Second, one of the other, computationally cheaper estimation algorithms that will be discussed in the remainder of this section can be used to obtain initial estimates of the orientation.

\begin{algorithm}[htp]
\caption{\textsf{Smoothing estimates of the orientation using optimization}}
\label{alg:oriEst-smoothingOpt}
\small
\textsc{Inputs:} An initial estimate of the orientation $\tilde{q}^{\text{nb},(0)}_{1:N}$, inertial data $\left\{ y_{\text{a},t}, y_{\omega,t} \right\}_{t=1}^N$, magnetometer data $\left\{ y_{\text{m},t}\right\}_{t=1}^N$ and covariance matrices $\Sigma_\omega$, $\Sigma_\text{a}$ and $\Sigma_\text{m}$. \\
\textsc{Outputs:} An estimate of the orientation $\hat{q}^\text{nb}_{1:N}$ and optionally its covariance $\cov (\hat{\oriError}^\text{n}_{1:N})$.
\algrule[.4pt]
\begin{enumerate}
\item Set $\hat \oriError_t^{\text{n},(0)} = 0_{3 \times 1}$ for $t = 1, \hdots, N$, set $k = 0$ and compute $\initEst{q}_1^\text{nb}$ and $\Sigma_{\oriError,\text{i}}$ as described in \Sectionref{sec:models-prior}.
\item \textbf{while} \textit{termination condition is not satisfied} \textbf{do}
\begin{enumerate}
\item Compute the gradient~\eqref{eq:oriEst-gradient} and the approximate Hessian~\eqref{eq:oriEst-approxHessian} of the orientation smoothing problem~\eqref{eq:oriEst-oriSmoothing} using the expressions for the different parts of the cost function and their Jacobians~\eqref{eq:oriEst-cost} and~\eqref{eq:oriEst-der}.
\item Apply the update~\eqref{eq:oriEst-GNiteration} to obtain $\hat \oriError_{1:N}^{\text{n},(k+1)}$.
\item Update the linearization point as
\begin{align}
\tilde q^{\text{nb},(k+1)}_t = \expq \left( \tfrac{\hat{\oriError}_t^{\text{n},(k+1)}}{2} \right) \odot \tilde{q}^{\text{nb},(k)}_t,
\label{eq:oriEst-relinSmoothing}
\end{align}
and set $\hat \oriError_t^{\text{n},(k+1)} = 0_{3 \times 1}$ for $t = 1, \hdots, N$.
\item Set $k = k+1$.
\end{enumerate}
\item[] \textbf{end while}
\item Set $\hat{q}^\text{nb}_{1:N} = \tilde{q}^{\text{nb},(k)}_{1:N}$.
\item Optionally compute
\begin{align}
\cov (\hat{\oriError}^\text{n}_{1:N}) = \left( \mathcal{J}^\Transp(\hat{\oriError}^\text{n}_{1:N}) \mathcal{J} (\hat{\oriError}^\text{n}_{1:N}) \right)^{-1}.
\end{align}
\end{enumerate}
\normalsize
\end{algorithm}

\subsection{Computing the uncertainty}
\label{sec:oriEst-smoothingUncertainty}
As discussed in \Sectionref{sec:models-probModeling}, we are not only interested in obtaining point estimates of the position and orientation, but also in estimating their uncertainty. In our case of Gaussian noise, this is characterized by the covariance. As shown in \eg \cite{ljung:1999,verhaegenV:2007}, if our position and orientation estimation problems would be \gls{ls} problems, the covariance of the estimates would be given by the inverse of the Hessian of the objective function~\eqref{eq:oriEst-hessian}. Instead, we solve an \gls{nls} problem, for which a number of \gls{ls} problems are solved around linearization points closer and closer to the minimum of the objective function. Hence, when the algorithm has converged, the problem can locally well be described by the quadratic approximation around its resulting linearization point. We can therefore approximate the covariance of our estimates as
\begin{align}
\cov{ (\hat{x}_{1:N}) } = \left( \mathcal{J}^\Transp(\hat{x}_{1:N}) \mathcal{J}(\hat{x}_{1:N}) \right)^{-1}.
\end{align}
An intuition behind this expression is that the accuracy of the estimates is related to the sensitivity of the objective function with respect to the states. The covariance of the estimate will play a crucial role in the filtering approaches discussed in \Sectionref{sec:oriEst-filteringOpt} and~\Sectionref{sec:oriEst-ekf}.

The matrix $\mathcal{J}^\Transp(\hat{x}_{1:N}) \mathcal{J}(\hat{x}_{1:N})$ quickly becomes fairly large due to the high sampling rates of the inertial sensors. Hence, computing its inverse can be computationally costly. We are, however, typically only interested in a subset of the inverse. For instance, we are often only interested in diagonal or block diagonal elements representing $\cov (x_t)$. It is therefore not necessary to explicitly form the complete inverse, which makes the computation tractable also for larger problem sizes.

\section{Filtering estimation in an optimization framework}
\label{sec:oriEst-filteringOpt}
One of the benefits of using a smoothing formulation is that all measurements $y_{1:N}$ are used to get the best estimates of the states $x_{1:N}$. However, both the computational cost and the memory requirements grow with the length of the data set. Furthermore, it is a post-processing solution in which we have to wait until all data is available. Alternatively, the filtering problem can be formulated as 
\begin{align}
\label{eq:oriEst-filtOpt}
\hat{x}_{t+1} &= \argmin_{x_{t+1}} - \log p(x_{t+1} \mid y_{1:t+1}) \nonumber \\
&= \argmin_{x_{t+1}} - \log p(y_{t+1} \mid x_{t+1}) - \log p(x_{t+1} \mid y_{1:t}).
\end{align}
Note that without loss of generality, we have shifted our time indices as compared to the notation in \Sectionref{sec:models-probModeling}. The probability distribution $p(x_{t+1} \mid y_{1:t})$ can now be seen as a \emph{prior} and can be obtained by marginalizing out the previous state $x_{t}$ as
\begin{align}
\label{eq:oriEst-filteringOptMargDistr}
p(x_{t+1}  \mid y_{1:t}) &= \int p(x_{t+1} , x_{t} \mid y_{1:t} ) \dint x_{t} \nonumber \\
&= \int p(x_{t+1} \mid x_{t}) p(x_{t} \mid y_{1:t}) \dint x_{t}.
\end{align}
Assuming that
\begin{subequations}
\begin{align}
p(x_{t+1} \mid x_{t}) &\sim \mathcal{N}( x_{t+1} \, ; \, f(x_{t}), Q),  \\
p(x_{t} \mid y_{1:t}) &\sim \mathcal{N}( x_{t} \, ; \, \hat{x}_{t}, P_{t \mid t}),
\end{align}
\end{subequations}
the integral in~\eqref{eq:oriEst-filteringOptMargDistr} can be approximated as
\begin{align}
\label{eq:oriEst-filteringOptMargDistr-approx}
p(x_{t+1} \mid y_{1:t}) &\approx \mathcal{N}\left( x_{t+1} \, ; \,  f(\hat{x}_{t}), F_{t} P_{t \mid t} F_{t}^\Transp + G_{t} Q G_{t}^\Transp \right).
\end{align}
Here, $F_{t} = \givenThat{\tfrac{\partial f(x_{t})}{\partial x_{t}}}{x_t = \hat{x}_t}{w_t = 0}$, $\givenThat{G_{t} = \tfrac{\partial f(x_{t})}{\partial w_{t}}}{x_t = \hat{x}_t}{w_t = 0}$ and $w_t$ is the process noise defined in~\eqref{eq:models-dynamics}. Using~\eqref{eq:oriEst-filteringOptMargDistr-approx}, the filtering problem~\eqref{eq:oriEst-filtOpt} can again be written as a nonlinear least squares problem and can be solved using Gauss-Newton optimization as described in \Sectionref{sec:oriEst-GNopt}. Instead of optimizing over the whole data set at once as in \Sectionref{sec:oriEst-smoothingOpt}, we solve a small optimization problem at each time instance $t$. This is closely related to what is often called an iterated extended Kalman filter~\cite{jazwinski:1970,skoglundHA:2015}.

\subsection{Filtering estimates of the orientation using optimization}
\label{sec:oriEst-filteringOpt-ori}
In this section, we will derive an algorithm to obtain filtering estimates of the orientation using Gauss-Newton optimization. The resulting algorithm is closely related to the iterated multiplicative extended Kalman filter presented in \cite{skoglundSK:2017}.

To derive the approximation~\eqref{eq:oriEst-filteringOptMargDistr-approx} for the orientation estimation problem, we first derive an expression for the function $f$ that expresses $\oriError_{t+1}^\text{n}$ in terms of $\oriError^\text{n}_{t}$. Using~\eqref{eq:oriEst-linSmoothing} in combination with~\eqref{eq:models-ssOri-dyn}, 
\begin{align}
\oriError^\text{n}_{t+1} &= f(\oriError^\text{n}_{t}, y_{\omega,t}, e_{\omega,t}) \\ 
&= 2 \logq \left( \expq(\tfrac{\oriError^\text{n}_{t}}{2}) \odot \tilde{q}_{t}^\text{nb} \odot \expq \left(\tfrac{T}{2} (y_{\omega,t} + e_{\omega,t} ) \right) \odot \tilde{q}_{t+1}^\text{bn} \right). \nonumber
\label{eq:oriEst-filtOptCost}
\end{align}
Here, $\oriError^\text{n}_{t}$ and $\tilde{q}_{t}^\text{nb}$ provide information about the previous time step. We denote the orientation estimate from the previous time step as $\hat{q}_{t}^\text{nb}$ and assume that $\hat{\oriError}_t^\text{n}$ is zero since the linearization point is updated after each iteration of the optimization algorithm. To start with a fairly good linearization point, we choose $\tilde{q}_{t+1}^{\text{nb},(0)}$ at iteration $k = 0$ as
\begin{align}
\tilde{q}_{t+1}^{\text{nb},(0)} = \hat{q}_{t}^\text{nb} \odot \expq \left( \tfrac{T}{2} y_{\omega,t} \right).
\end{align}
Following~\eqref{eq:oriEst-filteringOptMargDistr-approx}, the distribution $p(\oriError^\text{n}_{t+1} \mid y_{1:t})$ around this linearization point can now be written as
\begin{align}
\label{eq:oriEst-filtOptDist}
p(\oriError^\text{n}_{t+1} \mid y_{1:t}) &\approx \mathcal{N}\left( \oriError^\text{n}_{t+1} \, ; \, 0, F_{t} P_{t \mid t} F_{t}^\Transp + G_{t} \Sigma_\omega G_{t}^\Transp \right),
\end{align}
with 
\begin{subequations}
\label{eq:oriEst-filtOptFG}
\begin{align}
F_{t} &= \givenThat{\tfrac{\partial f(\oriError^\text{n}_{t},y_{\omega,t},e_{\omega,t})}{\partial \oriError^\text{n}_{t}}}{\oriError^\text{n}_{t} = 0, \tilde{q}_{t}^\text{nb} = \hat{q}_{t}^\text{nb}}{e_{\omega,t} = 0} \nonumber \\
&= 2 \left. \tfrac{\partial}{\partial \oriError^\text{n}_{t}}  \logq \left( \expq(\tfrac{\oriError^\text{n}_{t}}{2}) \right) \right|_{\oriError^\text{n}_{t} = 0} \nonumber \\
&= \mathcal{I}_3, \label{eq:oriEst-filtOptDynF} \\
G_{t} &= \givenThat{\tfrac{\partial f(\oriError^\text{n}_{t},y_{\omega,t},e_{\omega,t})}{\partial e_{\omega,t}}}{\oriError^\text{n}_{t} = 0, \tilde{q}_{t}^\text{nb} = \hat{q}_{t}^\text{nb}}{e_{\omega,t} = 0} \nonumber \\
&= 2 \left. \tfrac{\partial}{\partial e_{\omega,t}}   \logq \left( \hat{q}_{t}^\text{nb} \odot \expq \left(\tfrac{T}{2} (y_{\omega,t} + e_{\omega,t} ) \right) \right. \right. \nonumber \\
& \qquad \qquad \left. \left. \odot \expq \left(-\tfrac{T}{2} y_{\omega,t} \right) \odot \hat{q}_{t}^\text{bn} \right) \right|_{e_{\omega,t} = 0} \nonumber \\
&\approx T \tilde{R}_{t+1}^{\text{nb},(0)}.
\end{align}
\end{subequations}

The covariance $P_{t \mid t}$ in~\eqref{eq:oriEst-filtOptDist} can be approximated as the inverse of the Hessian of the objective function from the previous time step, see also \Sectionref{sec:oriEst-smoothingUncertainty}. We make use of the shorthand notation $P_{t+1 \mid t} = F_{t} P_{t \mid t} F_{t}^\Transp + G_{t} Q G_{t}^\Transp$ and define
\begin{align}
\label{eq:oriEst-costDerFiltMarg}
e_{\text{f},t} = \oriError^\text{n}_{t}- 2 \logq \left(\hat{q}_{t-1}^\text{nb} \odot \expq (\tfrac{T}{2} y_{\omega,t-1} ) \odot  \tilde{q}_{t}^\text{bn} \right), 
\qquad \tfrac{\diff e_{\text{f},t}}{\diff \oriError^\text{n}_{t}} &= \mathcal{I}_3.
\end{align}
Note that $e_{\text{f},t}$ is equal to zero for iteration $k = 0$ but can be non-zero for subsequent iterations. Using this notation, the filtering problem~\eqref{eq:oriEst-filtOpt} results in the following optimization problem
\begin{align}
\label{eq:oriEst-filtOptNLS}
\hat{x}_t &= \argmin_{x_{t}} - \log p(x_{t} \mid y_{1:t}) \nonumber \\
&= \argmin_{x_{t}} \underbrace{\| e_{\text{f},t} \|_{P_{t \mid t-1}^{-1}}}_{\begin{subarray}{c}\text{Dynamics and}\\
    \text{knowledge about $x_{t-1}$}\end{subarray}} + \underbrace{\vphantom{\| e_{\text{f},t} \|_{P_{t \mid t-1}^{-1}}} \| e_{\text{a},t} \|_{\Sigma_\text{a}^{-1}} + \| e_{\text{m},t} \|_{\Sigma_\text{m}^{-1}}}_{\text{Measurement models}}.
\end{align}
Note the similarity of this optimization problem to the smoothing formulation in~\eqref{eq:oriEst-oriSmoothing}. The term $\| e_{\text{f},t} \|_{P_{t \mid t-1}^{-1}}$ takes into account both the knowledge about the previous state $x_t$ and the dynamics. Furthermore, due to the fact that $P_{t \mid t-1}^{-1}$ is time-varying, the uncertainty and cross-correlation of the states at the previous time instance is taken into consideration. Including this term is similar to the inclusion of an arrival cost in moving horizon estimation approaches~\citep{raoRM:2003}.

After each Gauss-Newton iteration, we need to recompute the linearization point as
\begin{align}
\label{eq:oriEst-filtOptRelinCov}
\tilde{q}_{t}^{\text{nb},(k+1)} &= \expq \left( \tfrac{\hat{\oriError}_t^{\text{n},(k+1)}}{2} \right) \odot \tilde{q}_{t}^{\text{nb},(k)}.
\end{align}
We also need to compute the covariance around this updated linearization point as
\begin{align}
P_{t \mid t-1}^{(k+1)} &= J_{t}^{(k)} P_{t \mid t-1}^{(k)} \left( J_{t}^{(k)} \right)^\Transp,
\end{align}
where $J_t^{(k)}$ can be derived similarly to the derivation of~$F_t$ in~\eqref{eq:oriEst-filtOptDynF}. Since $J_t^{(k)} = \mathcal{I}_3$, in practice the relinearization of the covariance can be omitted. The process of estimating orientation using filtering is summarized in \Algorithmref{alg:oriEst-filteringOpt}.

\begin{algorithm}[ht]
\caption{\textsf{Filtering estimates of the orientation using optimization}}
\label{alg:oriEst-filteringOpt}
\small
\textsc{Inputs:} Inertial data $\left\{ y_{\text{a},t}, y_{\omega,t} \right\}_{t=1}^N$, magnetometer data $\left\{ y_{\text{m},t}\right\}_{t=1}^N$ and covariance matrices $\Sigma_\omega$, $\Sigma_\text{a}$ and $\Sigma_\text{m}$. \\
\textsc{Outputs:} An estimate of the orientation $\hat{q}^\text{nb}_{t}$ and its covariance $P_{t \mid t}$ for $t = 1, \hdots N$.
\algrule[.4pt]
\begin{enumerate}
\item Compute $\initEst{q}_1^\text{nb}$ and $\Sigma_\text{i}$ as described in \Sectionref{sec:models-prior} and set $\hat{q}_{1}^\text{nb} = \initEst{q}_{1}^\text{nb}$ and $P_{1 \mid 1} = \Sigma_{\oriError,\text{i}}$.
\item \textbf{for} $t = 2, \hdots N$ \textbf{do}
\begin{enumerate}
\item Set $\hat \oriError_t^{\text{n},(0)} = 0_{3 \times 1}$, set $k = 0$, choose the linearization point $\tilde q^{\text{nb},(0)}_t$ as
\begin{subequations}
\begin{align}
\label{eq:oriEst-filteringOpt-updateLinPointIt0}
\tilde q^{\text{nb},(0)}_t &= \hat{q}^{\text{nb}}_{t-1} \odot \expq \left( \tfrac{T}{2} y_{\omega,t-1} \right), \\
\intertext{and compute $P_{t \mid t-1}$ as}
P_{t \mid t-1} &= P_{t-1 \mid t-1} + G_{t-1} \Sigma_\omega G_{t-1}^\Transp,
\label{eq:oriEst-timeUpdateFiltOpt}
\end{align}
\end{subequations}
with $G_{t-1} = T \tilde{R}^{\text{nb},(0)}_{t}$.
\item \textbf{while} \textit{termination condition is not satisfied} \textbf{do}
\begin{enumerate}
\item Compute the gradient~\eqref{eq:oriEst-gradient} and the approximate Hessian~\eqref{eq:oriEst-approxHessian} of the filtering problem~\eqref{eq:oriEst-filtOptNLS} using the expressions for the different parts of the cost function and their Jacobians~\eqref{eq:oriEst-costDerFiltMarg},~\eqref{eq:oriEst-costAcc},~\eqref{eq:oriEst-costMag},~\eqref{eq:oriEst-derAcc} and~\eqref{eq:oriEst-derMag}.
\item Apply the update~\eqref{eq:oriEst-GNiteration} to obtain $\hat \oriError_{t}^{\text{n},(k+1)}$.
\item Update the linearization point as 
\begin{align}
\tilde q^{\text{nb},(k+1)}_t &= \expq \left( \tfrac{\hat{\oriError}_t^{\text{n},(k+1)}}{2} \right) \odot \tilde{q}^{\text{nb},(k)}_t, 
\label{eq:oriEst-relinFiltering}
\end{align}
$\hat \oriError_t^{\text{n},(k+1)} = 0_{3 \times 1}$.
\item Set $k = k+1$.
\end{enumerate}
\item[] \textbf{end while}
\item Set $\hat{q}^\text{nb}_{t} = \tilde{q}^{\text{nb},(k)}_{t}$, $\hat{\oriError}^\text{n}_{t} = \hat{\oriError}^{\text{n},(k)}_{t}$ and compute $P_{t \mid t}$ as
\begin{align}
P_{t \mid t} = \left( \mathcal{J}^\Transp(\hat{\oriError}^\text{n}_{t}) \mathcal{J}(\hat{\oriError}^\text{n}_{t}) \right)^{-1}.
\end{align}
\end{enumerate}
\end{enumerate}
\normalsize
\end{algorithm}

\section{Extended Kalman filtering}
\label{sec:oriEst-ekf}
Instead of using optimization for position and orientation estimation, it is also possible to use extended Kalman filtering. \Glspl{ekf} compute filtering estimates in terms of the conditional probability distribution~\eqref{eq:models-filtering}. Hence, the approach is similar to the one discussed in \Sectionref{sec:oriEst-filteringOpt}. In fact, extended Kalman filtering can be interpreted as Gauss-Newton optimization of the filtering problem using only one iteration~\eqref{eq:oriEst-GNiteration} with a step length of one, see \eg \cite{skoglundHA:2015}. In this section we first introduce how an \gls{ekf} can be used to compute state estimates in a general nonlinear state space model. Subsequently, we illustrate the use of \glspl{ekf} for position and orientation estimation by focusing on the orientation estimation problem. Two different implementations will be discussed. First, we introduce an \gls{ekf} implementation that uses unit quaternions as states. Subsequently, we discuss an \gls{ekf} which parametrizes the orientation in terms of an orientation deviation from a linearization point, similar to the approach used in \Sectionref{sec:oriEst-smoothingOpt} and~\Sectionref{sec:oriEst-filteringOpt}. 

An \gls{ekf} makes use of a nonlinear state space model. Assuming that the measurement noise is additive and that both the process and the measurement noise are zero-mean Gaussian with constant covariance, the state space model is given by
\begin{subequations}
\begin{align}
x_{t+1} &= f_t(x_t,u_t,w_t), \label{eq:oriEst-generalSSmodel-dyn} \\
y_t &= h_t(x_t) + e_t, \label{eq:oriEst-generalSSmodel-meas}
\end{align}
\label{eq:oriEst-generalSSmodel}%
\end{subequations}%
with $w_t \sim \mathcal{N}(0,Q)$ and $e_t \sim \mathcal{N}(0,R)$.

The state is estimated recursively by performing a \emph{time update} and a \emph{measurement update}. The time update uses the model~\eqref{eq:oriEst-generalSSmodel-dyn} to ``predict'' the state to the next time step according to 
\begin{subequations}
\begin{align}
\hat{x}_{t+1 \mid t} &= f_t(\hat{x}_{t \mid t},u_t), \label{eq:ekfTxupdate} \\
P_{t+1 \mid t} &= F_t P_{t \mid t} F_t^\Transp + G_t Q G_t^\Transp, \label{eq:ekfTPupdate}
\end{align}
\label{eq:oriEst-ekfTimeUpdate}%
\end{subequations}%
with 
\begin{align}
F_t = \givenThat{\tfrac{\partial f_t(x_t,u_t,w_t)}{\partial x_t}}{x_t=\hat{x}_{t \mid t}}{w_t=0}, \qquad G_t = \givenThat{\tfrac{\partial f_t(x_t,u_t,w_t)}{\partial v_t}}{x_t=\hat{x}_{t \mid t}}{w_t=0}.
\label{eq:oriEst-defFG}
\end{align}
Here, the matrix $P$ denotes the state covariance. The double subscripts on $\hat{x}_{t+1 \mid t}$ and $P_{t+1 \mid t}$ denote the state estimate and the state covariance at time $t+1$ given measurements up to time $t$. Similarly, $\hat{x}_{t \mid t}$ and $P_{t \mid t}$ denote the state estimate and the state covariance at time $t$ given measurements up to time $t$. 

The measurement update makes use of the measurement model \eqref{eq:oriEst-generalSSmodel-meas} in combination with the measurements $y_t$ to update the ``predicted'' state estimate as 
\begin{subequations}
\begin{align}
\hat{x}_{t \mid t} &= \hat{x}_{t \mid t-1} + K_t \varepsilon_t, \label{eq:ekfMxupdate} \\
P_{t \mid t} &= P_{t \mid t-1} - K_t S_t K_t^\Transp, \label{eq:ekfMPupdate}
\end{align}
\label{eq:oriEst-ekfMeasUpdate}%
\end{subequations}%
with
\begin{align}
\label{eq:oriEst-defEpsKS}
\varepsilon_t \triangleq y_t - \hat{y}_{t \mid t-1}, \, \, \, \, S_t \triangleq H_t P_{t \mid t-1} H_t^\Transp + R, \, \, \, \, K_t \triangleq P_{t \mid t-1} H_t^\Transp S_t^{-1},
\end{align}
and
\begin{align}
\label{eq:oriEst-defYhatH}
\hat{y}_{t \mid t-1} = h(\hat{x}_{t \mid t-1}), \qquad H_t = \left. \tfrac{\partial h_t(x_t)}{\partial x_t} \right|_{x_t=\hat{x}_{t \mid t-1}}.
\end{align}
Note that in~\eqref{eq:oriEst-ekfMeasUpdate} we have shifted our notation one time step compared to the notation in~\eqref{eq:oriEst-ekfTimeUpdate} to avoid cluttering the notation. The \gls{ekf} iteratively performs a time update and a measurement update to estimate the state and its covariance.

\subsection{Estimating orientation using quaternions as states}
\label{sec:oriEst-quat-ekf}
In this section, we will illustrate the use of an \gls{ekf} to compute filtering estimates of the orientation. As discussed in the previous section, the crucial part is to compute the matrices $F_t$, $G_t$ and $H_t$ to perform the \gls{ekf} time and measurement updates. Using the state space model~\eqref{eq:models-ssOri} and using unit quaternions as states in the \gls{ekf}, the dynamic model is given by 
\begin{align}
q^\text{nb}_{t+1} &= f_t(q^\text{nb}_t,y_{\omega,t},e_{\omega,t}) = q^\text{nb}_t \odot \expq \left(\tfrac{T}{2} ( y_{\omega,t} - e_{\omega,t} ) \right) \nonumber \\
&= \left( \expq \left(\tfrac{T}{2} ( y_{\omega,t} - e_{\omega,t} ) \right) \right)^\rightMult q^{\text{nb}}_t \nonumber \\
&= \left( q^\text{nb}_t \right)^\leftMult \expq \left(\tfrac{T}{2} ( y_{\omega,t} - e_{\omega,t} ) \right).
\label{eq:oriEst-quatEKF-dyn}
\end{align}
The following derivatives of the dynamic model~\eqref{eq:oriEst-quatEKF-dyn} can be obtained
\begin{subequations}
\begin{align}
F_t &= \givenThat{\tfrac{\partial f_t(q^\text{nb}_t,y_{\omega,t},e_{\omega,t})}{\partial q^\text{nb}_t}}{q^\text{nb}_{t}=\hat{q}^\text{nb}_{t \mid t}}{e_{\omega,t}=0} = \left( \expq (\tfrac{T}{2} y_{\omega,t} ) \right)^\rightMult, \\
G_t &= \givenThat{\tfrac{\partial f_t(q^\text{nb}_t,y_{\omega,t},e_{\omega,t})}{\partial e_{\omega,t}}}{q^\text{nb}_{t}=\hat{q}^\text{nb}_{t \mid t}}{e_{\omega,t}=0} \nonumber \\
&= \givenThat{\tfrac{\partial}{\partial e_{\omega,t}} \left( q^{\text{nb}}_t \right)^\leftMult \expq ( \tfrac{T}{2} ( y_{\omega,t} - e_{\omega,t} )) }{q^\text{nb}_{t}=\hat{q}^\text{nb}_{t \mid t}}{e_{\omega,t}=0} \nonumber \\
&= -\tfrac{T}{2} \left( \hat{q}^{\text{nb}}_{t \mid t} \right)^\leftMult \tfrac{\diff \expq (e_{\omega,t})}{\diff e_{\omega,t}}.
\end{align}
\end{subequations}

In the measurement update of the \gls{ekf}, the state is updated using the accelerometer and magnetometer measurements. Using the measurement models
\begin{align}
y_{\text{a},t} = - R^\text{bn}_t g^\text{n} + e_{\text{a},t}, \qquad y_{\text{m},t} = R^\text{bn}_t m^\text{n} + e_{\text{m},t},
\end{align}
the matrix $H_t$ in~\eqref{eq:oriEst-defYhatH} is given by
\begin{align}
H_t &= \left. \tfrac{\partial}{\partial q^\text{nb}_t} \begin{pmatrix} R^\text{bn}_t g^\text{n} \\ R^\text{bn}_t m^\text{n} \end{pmatrix} \right|_{q^\text{nb}_t=\hat{q}^\text{nb}_{t \mid t-1}} 
= \begin{pmatrix} - \left. \tfrac{\partial R^\text{bn}_{t \mid t-1}}{\partial q^\text{nb}_{t \mid t-1}} \right|_{q^\text{nb}_{t \mid t-1}=\hat{q}^\text{nb}_{t \mid t-1}} g^\text{n} \\ \left. \tfrac{\partial R^\text{bn}_{t \mid t-1}}{\partial q^\text{nb}_{t \mid t-1}} \right|_{q^\text{nb}_{t \mid t-1}=\hat{q}^\text{nb}_{t \mid t-1}} m^\text{n} \end{pmatrix}.
\end{align}
The derivative $\tfrac{\partial R^\text{bn}_{t \mid t-1}}{\partial q^\text{nb}_{t \mid t-1}}$ can be computed from the definition of the relation between and the rotation matrix and the quaternion representation given in~\eqref{eq:app-defRq}.

Note that the quaternion obtained from the measurement update \eqref{eq:oriEst-ekfMeasUpdate} is no longer normalized. We denote this unnormalized quaternion and its covariance by $\tilde{q}_{t \mid t}^\text{nb}$ and $\tilde{P}_{t \mid t}$, respectively. The $\tilde{\cdot}$ instead of the $\hat{\cdot}$ is meant to explicitly indicate that the quaternion still needs to be updated. This is done by an additional renormalization step as compared to a standard \gls{ekf} implementation. A possible interpretation of the renormalization is as an additional measurement update without measurement noise. Hence, we adjust both the quaternion and its covariance as
\begin{subequations}
\label{eq:oriEst-ekfQuatRenorm}
\begin{align}
\hat{q}_{t \mid t}^\text{nb} &= \tfrac{\tilde{q}^\text{nb}_{t \mid t}}{\| \tilde{q}^\text{nb}_{t \mid t}\|_2},  \qquad
P_{t \mid t} = J_t \tilde{P}_{t \mid t} J_t^\Transp, 
\intertext{with}
J_t &= \tfrac{1}{\| \tilde{q}^\text{nb}_{t \mid t} \|_2^3} \tilde{q}^\text{nb}_{t \mid t} \left( \tilde{q}^\text{nb}_{t \mid t} \right)^\Transp.
\end{align}
\end{subequations}
Here, $q^\Transp$ is the standard vector transpose. The resulting \gls{ekf} is summarized in \Algorithmref{alg:oriEst-ekfQuat}. 

\begin{algorithm}[ht]
\caption{\textsf{Orientation estimation using an EKF with quaternion states}}
\label{alg:oriEst-ekfQuat}
\small
\textsc{Inputs:} Inertial data $\left\{ y_{\text{a},t}, y_{\omega,t} \right\}_{t=1}^N$, magnetometer data $\left\{ y_{\text{m},t}\right\}_{t=1}^N$ and covariance matrices $\Sigma_\omega$, $\Sigma_\text{a}$ and $\Sigma_\text{m}$. \\
\textsc{Outputs:} An estimate of the orientation $\hat{q}^\text{nb}_{t \mid t}$ and its covariance $P_{t \mid t}$ for $t = 1, \hdots N$.
\algrule[.4pt]
\begin{enumerate}
\item Compute $\initEst{q}^\text{nb}_{1}$ and $\Sigma_\text{i}$ as described in \Sectionref{sec:models-prior} and set $\hat{q}_{1 \mid 1}^\text{nb} = \initEst{q}_{1}^\text{nb}$ and $P_{1 \mid 1} = \Sigma_\text{q,i}$.
\item \textbf{For} $t = 2, \hdots, N$ \textbf{do}
\begin{enumerate}
\item Time update
\begin{subequations}
\begin{align}
\hat{q}^\text{nb}_{t \mid t-1} &= \hat{q}^\text{nb}_{t-1 \mid t-1} \odot \expq \left( \tfrac{T}{2} y_{\omega,t-1} \right), \\
P_{t \mid t-1} &= F_{t-1} P_{t-1 \mid t-1} F_{t-1}^\Transp + G_{t-1} Q G_{t-1}^\Transp, \label{eq:quatEKF_TPupdate} %
\end{align}%
\label{eq:ori_qEkf_timeUpdate}%
\end{subequations}%
with $Q = \Sigma_\omega$ and 
\begin{align*}
F_{t-1} = \left( \expq (\tfrac{T}{2} y_{\omega,t-1}) \right)^\rightMult, \, G_{t-1} = -\tfrac{T}{2} \left( \hat{q}^{\text{nb}}_{t-1 \mid t-1} \right)^\leftMult \tfrac{\diff \expq (e_{\omega,t-1})}{\diff e_{\omega,t-1}}.
\end{align*}
\item Measurement update
\begin{subequations}
\begin{align}
\tilde{q}^\text{nb}_{t \mid t} &= \hat{q}^\text{nb}_{t \mid t-1} + K_t \varepsilon_t,  \\
\tilde{P}_{t \mid t} &= P_{t \mid t-1} - K_t S_t K_t^\Transp, 
\end{align}
\label{eq:ori_qEkf_measUpdate}
\end{subequations}%
with $\varepsilon_t$, $K_t$ and $S_t$ defined in~\eqref{eq:oriEst-defEpsKS} and
\begin{align*}
y_t &= \begin{pmatrix} y_{\text{a},t} \\ y_{\text{m},t} \end{pmatrix}, & 
\hat{y}_{t \mid t-1} &= \begin{pmatrix} -\hat{R}^\text{bn}_{t \mid t-1} g^\text{n} \\ \hat{R}^\text{bn}_{t \mid t-1} m^\text{n} \end{pmatrix}, \\
H_t &= \begin{pmatrix} - \left. \tfrac{\partial R^\text{bn}_{t \mid t-1}}{\partial q^\text{nb}_{t \mid t-1}} \right|_{q^\text{nb}_{t \mid t-1}=\hat{q}^\text{nb}_{t \mid t-1}} g^\text{n} \\ \left. \tfrac{\partial R^\text{bn}_{t \mid t-1}}{\partial q^\text{nb}_{t \mid t-1}} \right|_{q^\text{nb}_{t \mid t-1}=\hat{q}^\text{nb}_{t \mid t-1}} m^\text{n} \end{pmatrix}, & 
R &= \begin{pmatrix} \Sigma_\text{a} & 0 \\ 0 & \Sigma_\text{m} \end{pmatrix}.
\end{align*}
\item Renormalize the quaternion and its covariance as
\begin{align}
\hat{q}_{t \mid t}^\text{nb} = \tfrac{\tilde{q}^\text{nb}_{t \mid t}}{\| \tilde{q}^\text{nb}_{t \mid t}\|_2} , \qquad
P_{t \mid t} = J_t \tilde{P}_{t \mid t} J_t^\Transp,
\end{align}
with $J_t = \tfrac{1}{\| \tilde{q}^\text{nb}_{t \mid t} \|_2^3} \tilde{q}^\text{nb}_{t \mid t} \left( \tilde{q}^\text{nb}_{t \mid t} \right)^\Transp$.
\end{enumerate}
\item[] \textbf{end for}
\end{enumerate}
\normalsize
\end{algorithm}

\subsection{Estimating orientation using orientation deviations as states}
\label{sec:oriEst-oriError-ekf}
An alternative \gls{ekf} implementation parametrizes the orientation in terms of an orientation deviation around a linearization point. The linearization point is parametrized in terms of quaternions or rotation matrices and denoted $\tilde{q}^\text{nb}_t$ or equivalently $\tilde{R}^\text{nb}_t$. The orientation deviation $\oriError_t^\text{n}$ is the state vector in the \gls{ekf}. This \gls{ekf} implementation is sometimes referred to as a multiplicative \gls{ekf}~\citep{crassidisMC:2007,markley:2003}. One of its advantages is that its implementation is computationally attractive since it only uses a 3-dimensional state compared to the 4-dimensional state in \Algorithmref{alg:oriEst-ekfQuat}.

Similarly to~\eqref{eq:oriEst-filtOptCost} in \Sectionref{sec:oriEst-filteringOpt}, the dynamics of the state $\oriError^\text{n}$ is given by  
\begin{align}
\oriError_{t+1}^\text{n} &= f_t \left( \oriError_t^\text{n},y_{\omega,t},e_{\omega,t} \right)  \nonumber \\
&= 2 \log \left( \expq(\tfrac{ \oriError^\text{n}_t}{2}) \odot \tilde{q}_{t}^\text{nb} \odot  \expq \left( \tfrac{T}{2} (y_{\omega,t} - e_{\omega,t}) \right) \odot \tilde{q}_{t+1}^\text{bn} \right).  
\label{eq:oriEst-oriErrorEKF-dyn}
\end{align}
In the time update of the \gls{ekf}, we first use~\eqref{eq:models-ssOri-dyn} to directly update the linearization point as
\begin{align}
\tilde{q}^\text{nb}_{t+1 \mid t} = \tilde{q}^\text{nb}_{t \mid t} \odot \expq \left( \tfrac{T}{2} y_{\omega,t} \right).
\label{eq:oriEst-oriErrorEKF-dynUpdateLin}
\end{align}
Around this linearization point, the derivatives of the dynamic model~\eqref{eq:oriEst-oriErrorEKF-dyn} are given by
\begin{subequations}
\begin{align}
F_t &= \givenThat{\tfrac{\partial f_t(\oriError_t^\text{n},y_{\omega,t},e_{\omega,t})}{\partial \oriError_t^\text{n}}}{\oriError^\text{n}_t=0}{e_{\omega,t}=0} = \mathcal{I}_3 , 
\nonumber \\
G_t &= \givenThat{\tfrac{\partial f_t(\oriError^\text{n}_t,y_{\omega,t},e_{\omega,t})}{\partial e_{\omega,t}}}{\oriError^\text{n}_t=0}{e_{\omega,t}=0}  = T \tilde{R}_{t+1 \mid t}^\text{nb}. 
\end{align}
\label{eq:oriEst-oriErrorEKF-dynDer}
\end{subequations}
Note the similarity with~\eqref{eq:oriEst-filtOptFG}.

In the measurement update of the \gls{ekf}, the state $\oriError^\text{n}_t$ is updated using the accelerometer and magnetometer measurements. The accelerometer measurement equation can be written in terms of the state $\oriError^\text{n}_t$ as
\begin{align}
y_{\text{a},t} &= - R^\text{bn}_t g^\text{n} + e_{\text{a},t} \nonumber \\
&= - \tilde{R}^\text{bn}_t \left( \expR([\oriError_t^\text{n} \times]) \right)^\Transp g^\text{n} + e_{\text{a},t} \nonumber \\
&\approx - \tilde{R}^\text{bn}_t \left( \mathcal{I}_3 - [ \oriError^\text{n}_t \times] \right) g^\text{n} + e_{\text{a},t} \nonumber \\
&= - \tilde{R}^\text{bn}_t g^\text{n} - \tilde{R}^\text{bn}_t [g^\text{n} \times] \oriError^\text{n}_t + e_{\text{a},t}.
\end{align}
Equivalently, the magnetometer measurement equation can be written in terms of the state $\oriError_t^\text{n}$ as
\begin{align}
y_{\text{m},t} &= R^\text{bn}_t m^\text{n} + e_{\text{m},t} \nonumber \\
&= \tilde{R}^\text{bn}_t \left( \expR([\oriError_t^\text{n} \times]) \right)^\Transp m^\text{n} + e_{\text{m},t} \nonumber \\
&\approx \tilde{R}^\text{bn}_t \left( \mathcal{I}_3 - [\oriError_t^\text{n} \times] \right) m^\text{n} + e_{\text{m},t} \nonumber \\
&= \tilde{R}^\text{bn}_t m^\text{n} + \tilde{R}_t^\text{bn} [m^\text{n} \times] \oriError_t^\text{n} + e_{\text{m},t}.
\end{align}
From these equations, the derivatives $H_t$ as defined in~\eqref{eq:oriEst-defYhatH} can straightforwardly be computed. 

After the measurement update, the orientation deviation $\hat{\oriError}^\text{n}_t$ is non-zero. Hence, as an additional step in the \gls{ekf}, we update the linearization point and reset the state. In our algorithm, we consider the relinearization as the ``measurement update'' for the linearization point, \ie the relinearization updates the linearization point $\tilde{q}^\text{nb}_{t \mid t-1}$ to $\tilde{q}^\text{nb}_{t \mid t}$ as
\begin{align}
\tilde{q}^\text{nb}_{t \mid t} = \expq \left( \tfrac{\hat{\oriError}^\text{n}_t}{2} \right) \odot \tilde{q}^\text{nb}_{t \mid t-1}.
\end{align}
For the same reasons as in \Sectionref{sec:oriEst-filteringOpt}, the relinearization of the covariance can be omitted. The resulting \gls{ekf} is summarized in \Algorithmref{alg:oriEst-ekfOriError}. Note the similarities between this algorithm and \Algorithmref{alg:oriEst-filteringOpt}.

\begin{algorithm}[ht]
\caption{\textsf{Orientation estimation using an EKF with orientation deviation states}}
\label{alg:oriEst-ekfOriError}
\small
\textsc{Inputs:} Inertial data $\left\{ y_{\text{a},t}, y_{\omega,t} \right\}_{t=1}^N$, magnetometer data $\left\{ y_{\text{m},t}\right\}_{t=1}^N$ and covariance matrices $\Sigma_\omega$, $\Sigma_\text{a}$ and $\Sigma_\text{m}$. \\
\textsc{Outputs:} An estimate of the orientation $\tilde{q}^\text{nb}_{t \mid t}$ and the covariance $P_{t \mid t}$ for $t = 1, \hdots N$.
\algrule[.4pt]
\begin{enumerate}
\item Compute $\initEst{q}^\text{nb}_{1}$ and $\Sigma_\text{i}$ as described in \Sectionref{sec:models-prior} and set $\tilde{q}_{1 \mid 1}^\text{nb} = \initEst{q}_{1}^\text{nb}$ and $P_{1 \mid 1} = \Sigma_{\oriError,\text{i}}$.
\item \textbf{For} $t = 2, \hdots, N$ \textbf{do}
\begin{enumerate}
\item Time update
\begin{subequations}
\begin{align}
\tilde{q}^{\text{nb}}_{t \mid t-1} &= \tilde{q}^{\text{nb}}_{t-1 \mid t-1} \odot \expq \left( \tfrac{T}{2} y_{\omega,t-1} \right), \label{eq:oriErrorEKF_Txupdate} \\
P_{t \mid t-1} &= P_{t-1 \mid t-1} + G_{t-1} Q G_{t-1}^\Transp, \label{eq:oriErrorEKF_TPupdate}
\end{align}
\label{eq:oriErrorEKF_Tupdate}
\end{subequations}
with $G_{t-1} = T \tilde{R}_{t \mid t-1}^{\text{nb}}$ and $Q = \Sigma_\omega$.
\item Measurement update
\begin{subequations}
\begin{align}
\hat{\oriError}_t^\text{n} &= K_t \varepsilon_t, \label{eq:oriErrorEKF_Mxupdate_alg} \\
\tilde{P}_{t \mid t} &= P_{t \mid t-1} - K_t S_t K_t^\Transp, \label{eq:oriErrorEKF_MPupdate_alg}
\end{align}
\label{eq:oriErrorEKF_Mupdate}
\end{subequations}
with $\varepsilon_t$, $K_t$ and $S_t$ defined in~\eqref{eq:oriEst-defEpsKS} and
\begin{align*}
y_t &= \begin{pmatrix} y_{\text{a},t} \\ y_{\text{m},t} \end{pmatrix}, 
& \hat{y}_{t \mid t-1} &= \begin{pmatrix} -\tilde{R}^\text{bn}_{t \mid t-1} g^\text{n} \\ \tilde{R}^\text{bn}_{t \mid t-1} m^\text{n} \end{pmatrix}, 
 \\
H_t &= \begin{pmatrix} - \tilde{R}^\text{bn}_{t \mid t-1} [g^\text{n} \times] \\ \tilde{R}^\text{bn}_{t \mid t-1} [m^\text{n} \times]  \end{pmatrix}, & 
R &= \begin{pmatrix} \Sigma_\text{a} & 0 \\ 0 & \Sigma_\text{m} \end{pmatrix}.
\end{align*}
\item Relinearize
\begin{align}
\tilde{q}^\text{nb}_{t \mid t} = \expq \left( \tfrac{\hat{\oriError}_t^\text{n}}{2} \right) \odot \tilde{q}^\text{nb}_{t \mid t-1}. 
\end{align}
\end{enumerate}
\item[] \textbf{end for}
\end{enumerate}
\normalsize
\end{algorithm}

\section{Complementary filtering}
\label{sec:oriEst-compl}
An alternative to using \glspl{ekf} for orientation estimation is to use complementary filters \cite{higgins:1975,brown:1972}. This type of algorithms again estimates the orientation at time $t$ given the measurements $y_{1:t}$. However, it does not use the probabilistic models presented in \Chapterref{cha:models}. Instead, complementary filters explicitly use the fact that both the gyroscope and the combination of the accelerometer and the magnetometer provide information about the orientation of the sensor. The orientation estimates obtained from the accelerometer and the magnetometer measurements are noisy but accurate over long periods of time. On the other hand, the orientation estimates using the gyroscope measurements are accurate on a short time scale but they drift over time. These properties can be interpreted and exploited in the frequency domain. The orientation estimates using the gyroscope have desirable properties at high frequencies. We would therefore like to filter these using a high-pass filter. Conversely, the orientation estimates obtained from the accelerometer and magnetometer measurements have desirable properties at low frequencies. We would therefore like to filter these orientation estimates using a low-pass filter. 

This can be illustrated by considering the one-dimensional case of estimating an angle $\theta$ from gyroscope measurements $y_\omega$ and magnetometer measurements $y_\text{m}$. We denote the angle estimated by the complementary filter by $\hat{\theta}$, the angle obtained from the magnetometer measurements by $\theta_\text{m}$ and the angle obtained from the gyroscope measurements by $\theta_\omega$. Note that the latter is obtained by integrating the gyroscope measurements. The Laplace transforms of $\theta$, $\theta_\text{m}$, $\theta_\omega$ and $y_{\omega}$ are denoted by $\Theta(s)$, $\Theta_\text{m}(s)$, $\Theta_\omega(s)$ and $Y_{\omega}(s)$, respectively. The complementary filter computes $\Theta(s)$ as
\begin{align}
\label{eq:oriEst-complFilterLaplace}
\Theta(s) &= G(s) \Theta_\text{m}(s) + \left( 1 - G(s) \right) \Theta_{\omega}(s) \nonumber \\
&= G(s) \Theta_\text{m}(s) + \left( 1 - G(s) \right) \tfrac{1}{s} Y_\omega(s),
\end{align}
where $G(s)$ is a low-pass filter and $1 - G(s)$ is hence a high-pass filter. Note that the sum of the two filters, $G(s)$ and $1 - G(s)$, is equal to one, which is the reason behind the name \emph{complementary} filter. Choosing $G(s)$ as a first-order low-pass filter $G(s) = \tfrac{1}{a s + 1}$ and using Euler backward discretization, the filter~\eqref{eq:oriEst-complFilterLaplace} can be written in discrete time as
\begin{align}
\label{eq:oriEst-complFilterDiscrete}
\hat{\theta}_{t} = ( 1 - \gamma ) \theta_{\text{m},t} + \gamma \left( \hat{\theta}_{t-1} + T y_{\omega,t} \right),
\end{align}
where $\gamma = \tfrac{a}{T + a}$. The relation~\eqref{eq:oriEst-complFilterDiscrete} allows us to recursively estimate the angle $\hat{\theta}$. The only parameter that needs to be chosen to run the complementary filter~\eqref{eq:oriEst-complFilterDiscrete} is the parameter $a$ of the low-pass filter~$G(s)$. Choosing a large $a$ (and hence a $\gamma$ close to one) results in a lower cut-off frequency and a more significant contribution of the gyroscope measurements on the orientation estimates. Conversely, a small $a$ (and hence a $\gamma$ close to zero) results in a higher cut-off frequency and a more significant contribution of the magnetometer measurements on the orientation estimates. 

There is a strong relationship between complementary and Kalman filtering for linear models, see \eg \cite{higgins:1975,brown:1972}. To highlight the similarities and differences between complementary and extended Kalman filtering for orientation estimation, let us define the estimate from the complementary filter as $\hat{\theta}_{t \mid t}$ and write the recursion~\eqref{eq:oriEst-complFilterDiscrete} equivalently as 
\begin{subequations}
\label{eq:oriEst-complFilterDiscreteTwoUpdates}
\begin{align}
\hat{\theta}_{t \mid t-1} &= \hat{\theta}_{t-1} + T y_{\omega,t}, \label{eq:oriEst-complFilterDiscreteTwoUpdatesTimeUpdate} \\
\hat{\theta}_{t \mid t} &= ( 1 - \gamma ) \theta_{\text{m},t} + \gamma \hat{\theta}_{t \mid t-1}. \label{eq:oriEst-complFilterDiscreteTwoUpdatesMeasUpdate}
\end{align}
\end{subequations}

So far, we have discussed the one-dimensional case of estimating an angle $\theta$. Let us now focus on estimating orientation in three dimensions and parametrize this as a unit quaternion $q^\text{nb}$. Two well-known complementary filter implementations to estimate the orientation $q^\text{nb}$ are presented in \cite{mahonyHP:2008} and \cite{madgwickHV:2011}. Open-source implementations of both algorithms are available online~\citep{xio:2017}. The filter presented in \cite{madgwickHV:2011} is specifically designed to be computationally efficient and has for instance been implemented in the \gls{ros}~\citep{ros:2017}. In this section, we will derive a complementary filter that is inspired by this implementation but which is also meant to illustrate the similarities and differences between complementary filters and extended Kalman filters for orientation estimation. Because of that, the complementary filter that we will present is not as computationally efficient as the one in~\cite{madgwickHV:2011}. 

Similar to the one-dimensional case~\eqref{eq:oriEst-complFilterDiscreteTwoUpdatesTimeUpdate}, the orientation estimate of the complementary filter can be updated using the gyroscope measurements as 
\begin{align}
\label{eq:oriEst-complFilterGyrUpdate}
\hat{q}^\text{nb}_{t \mid t-1} = \hat{q}^\text{nb}_{t-1 \mid t-1} \odot \expq \left( \tfrac{T}{2} y_{\omega,t} \right),
\end{align}
where $\hat{q}^\text{nb}$ is the orientation estimate from the complementary filter and the double subscripts are defined analogously to~\eqref{eq:oriEst-complFilterDiscreteTwoUpdates}. Note that the update~\eqref{eq:oriEst-complFilterGyrUpdate} is equivalent to the time update in the \gls{ekf} in \Algorithmref{alg:oriEst-ekfQuat}. 

The orientation from the accelerometer and magnetometer measurements, denoted $R_{\text{am},t}^\text{nb}$, can be obtained by solving
\begin{align}
\label{eq:oriEst-complFilterAccMagOri}
q_{\text{am},t}^\text{nb} =& \argmin_{q_{\text{am},t}^\text{nb}} &&\| \bar{y}_{\text{a},t} + q_{\text{am},t}^\text{bn} \odot \bar{g}^\text{n} \odot q_{\text{am},t}^\text{nb} \|^2_{\Sigma_{\text{a}}^{-1}} + \nonumber \\
& && \qquad \qquad \| \bar{y}_{\text{m},t} - q_{\text{am},t}^\text{bn} \odot \bar{m}^\text{n} \odot q_{\text{am},t}^\text{nb} \|^2_{\Sigma_{\text{m}}^{-1}} \nonumber \\
& \st &&\| q_{\text{am},t}^\text{nb} \|_2 = 1,
\end{align}
see also \Sectionref{sec:models-prior}. Similar to \cite{madgwickHV:2011}, we do not completely solve the optimization problem~\eqref{eq:oriEst-complFilterAccMagOri} in each recursion of the complementary filter. Instead, we use an initial estimate $R_{\text{am},t}^{\text{nb},(0)} = \hat{q}^\text{nb}_{t \mid t-1}$ and only perform one iteration of an optimization algorithm. In \cite{madgwickHV:2011}, one iteration of a gradient descent algorithm is performed. We instead perform one Gauss-Newton update as in~\eqref{eq:oriEst-GNiteration}. Hence, using 
\begin{subequations}
\label{eq:oriEst-complFilterAccMagUpdateJe}
\begin{align}
\varepsilon_t &= \begin{pmatrix} \left( y_{\text{a},t} + \hat{R}^\text{bn}_{t \mid t-1} g^\text{n} \right) \Sigma_\text{a}^{-1/2} \\ \left( y_{\text{m},t} - \hat{R}^\text{bn}_{t \mid t-1} m^\text{n} \right) \Sigma_\text{m}^{-1/2} \end{pmatrix}, \\
\mathcal{J}_t &= \tfrac{\partial \varepsilon_t}{\partial q_{\text{am},t}^\text{nb}} = \begin{pmatrix} \left. \tfrac{\partial R_{\text{am},t}^\text{bn}}{\partial q_{\text{am},t}^\text{nb}} \right|_{q_{\text{am},t}^\text{nb}=\hat{q}^\text{nb}_{t \mid t-1}} g^\text{n} \, \Sigma_\text{a}^{-1/2} \\ \left. - \tfrac{\partial R_{\text{am},t}^\text{bn}}{\partial q_{\text{am},t}^\text{nb}} \right|_{q_{\text{am},t}^\text{nb}=\hat{q}^\text{nb}_{t \mid t-1}} m^\text{n} \, \Sigma_\text{m}^{-1/2} \end{pmatrix}, \label{eq:oriEst-complFilterAccMagUpdateJ}
\end{align}
\end{subequations}
the estimate $q_{\text{am},t}^\text{nb}$ is given by
\begin{align}
q_{\text{am},t}^\text{nb} = \hat{q}_{t \mid t-1}^\text{nb} - \beta \left( \mathcal{J}_t^\Transp \mathcal{J}_t \right)^{-1} \mathcal{J}_t^\Transp \varepsilon_t.
\end{align}
Analogously to~\eqref{eq:oriEst-complFilterDiscreteTwoUpdatesMeasUpdate}, the second update of the complementary filter is therefore given by
\begin{align}
\label{eq:oriEst-complFilterAccMagUpdate}
\hat{q}_{t \mid t}^\text{nb} &= ( 1 - \gamma ) \left( \hat{q}_{t \mid t-1}^\text{nb} - \beta \left( \mathcal{J}_t^\Transp \mathcal{J}_t \right)^{-1} \mathcal{J}_t^\Transp \varepsilon_t \right) + \gamma \hat{q}_{t \mid t-1}^\text{nb} \nonumber \\
&=  \hat{q}_{t \mid t-1}^\text{nb} - ( 1 - \gamma ) \beta \left( \mathcal{J}_t^\Transp \mathcal{J}_t \right)^{-1} \mathcal{J}_t^\Transp \varepsilon_t.
\end{align}
There are many similarities between~\eqref{eq:oriEst-complFilterAccMagUpdate} and the measurement update in the \gls{ekf} in \Algorithmref{alg:oriEst-ekfQuat}. First of all, if we neglect the presence of $\Sigma_\text{a}$ and $\Sigma_\text{m}$ for a moment, $\mathcal{J}_t$ and $\varepsilon_t$ in~\eqref{eq:oriEst-complFilterAccMagUpdateJe} are equal to $- H_t$ and $\varepsilon_t$ in the measurement update in \Algorithmref{alg:oriEst-ekfQuat}. Inclusion of $\Sigma_\text{a}$ and $\Sigma_\text{m}$ in~\eqref{eq:oriEst-complFilterAccMagUpdateJe} ensures that the uncertainty of the accelerometer and magnetometer measurements is properly taken into account when obtaining $q_{\text{am},t}^\text{nb}$. The contributions to the orientation estimate $\hat{q}_{t \mid t}^\text{nb}$ from integration of the gyroscope measurements and from $q_{\text{am},t}^\text{nb}$ are determined by the factor $(1 - \gamma) \beta$. In the \gls{ekf} on the other hand, the uncertainty of the gyroscope, accelerometer and magnetometer measurements is taken into account through the covariance matrices $\Sigma_\omega$, $\Sigma_\text{a}$ and $\Sigma_\text{m}$.

Second of all, comparing~\eqref{eq:oriEst-complFilterAccMagUpdate} to the \gls{ekf} measurement update \eqref{eq:oriEst-ekfMeasUpdate}, the update~\eqref{eq:oriEst-complFilterAccMagUpdate} can be interpreted as a scaled version of an \gls{ekf} update with $P_{t \mid t-1} = \mathcal{I}_4$ and $R = 0_{3 \times 3}$. In that case, the matrix $K_t$ in the \gls{ekf} would be given by $K_t = H_t^\Transp \left( H_t H_t^\Transp \right)^{-1}$. The matrix $H_t H_t^\Transp$ is rank deficient. Because of this, to compute $K_t$, we would have to use the pseudo-inverse, see \eg \cite{golubvL:2013}. Denoting the pseudo-inverse of a matrix as $\cdot^\dagger$, $K_t = H_t^\Transp \left( H_t H_t^\Transp \right)^{\dagger} = \left( H_t^\Transp H_t \right)^{-1} H_t^\Transp$. An important difference between the \gls{ekf} and complementary update is that the scaling factor $(1 - \gamma) \beta$ in~\eqref{eq:oriEst-complFilterAccMagUpdate} is constant, while the matrix $P_{t \mid t-1}$ in the \gls{ekf} is time-varying. In the remainder, we will denote this constant $(1 - \gamma) \beta$ by $\alpha$ for notational brevity. 

Our complementary filter implementation is summarized in \Algorithmref{alg:oriEst-compl}. Note again the similarity to \Algorithmref{alg:oriEst-ekfQuat}. To stress this similarity even more, we call the update~\eqref{eq:oriEst-complFilterGyrUpdate} the time update of the complementary filter and~\eqref{eq:oriEst-complFilterAccMagUpdate} the measurement update. 

\begin{algorithm}[ht]
\caption{\textsf{Orientation estimation using a complementary filter}}
\label{alg:oriEst-compl}
\small
\textsc{Inputs:} Inertial data $\left\{ y_{\text{a},t}, y_{\omega,t} \right\}_{t=1}^N$, magnetometer data $\left\{ y_{\text{m},t}\right\}_{t=1}^N$ and covariance matrices $\Sigma_\text{a}$ and $\Sigma_\text{m}$. \\
\textsc{Outputs:} An estimate of the orientation $\hat{q}^\text{nb}_{t \mid t}$ for $t = 1, \hdots N$.
\algrule[.4pt]
\begin{enumerate}
\item Compute $\initEst{q}^\text{nb}_{1}$ as described in \Sectionref{sec:models-prior} and set $\hat{q}_{1 \mid 1}^\text{nb} = \initEst{q}_{1}^\text{nb}$.
\item \textbf{For} $t = 2, \hdots, N$ \textbf{do}
\begin{enumerate}
\item Time update
\begin{align}
\hat{q}^\text{nb}_{t \mid t-1} &= \hat{q}^\text{nb}_{t-1 \mid t-1} \odot \expq \left( \tfrac{T}{2} y_{\omega,t-1} \right), 
\label{eq:ori_compl_timeUpdate}%
\end{align}%
\item Measurement update
\begin{align}
\tilde{q}^\text{nb}_{t \mid t} &= \hat{q}_{t \mid t-1}^\text{nb} - \alpha \left( \mathcal{J}_t^\Transp \mathcal{J}_t \right)^{-1} \mathcal{J}_t^\Transp \varepsilon_t, 
\label{eq:ori_compl_measUpdate}
\end{align}
with $\alpha = ( 1 - \gamma ) \beta$ and
\begin{align*}
\varepsilon_t &= \begin{pmatrix} \left( y_{\text{a},t} + \hat{R}^\text{bn}_{t \mid t-1} g^\text{n} \right) \Sigma_\text{a}^{-1/2} \\ \left( y_{\text{m},t} - \hat{R}^\text{bn}_{t \mid t-1} m^\text{n} \right) \Sigma_\text{m}^{-1/2} \end{pmatrix}, \\
\mathcal{J}_t &= \tfrac{\partial \varepsilon_t}{\partial q_{\text{am},t}^\text{nb}} = \begin{pmatrix} \left. \tfrac{\partial R_{\text{am},t}^\text{bn}}{\partial q_{\text{am},t}^\text{nb}} \right|_{q_{\text{am},t}^\text{nb}=\hat{q}^\text{nb}_{t \mid t-1}} g^\text{n} \, \Sigma_\text{a}^{-1/2} \\ \left. - \tfrac{\partial R_{\text{am},t}^\text{bn}}{\partial q_{\text{am},t}^\text{nb}} \right|_{q_{\text{am},t}^\text{nb}=\hat{q}^\text{nb}_{t \mid t-1}} m^\text{n} \, \Sigma_\text{m}^{-1/2} \end{pmatrix}.
\end{align*}
\item Renormalize the quaternion as
\begin{align}
\hat{q}_{t \mid t}^\text{nb} = \tfrac{\tilde{q}^\text{nb}_{t \mid t}}{\| \tilde{q}^\text{nb}_{t \mid t}\|_2}.
\end{align}
\end{enumerate}
\item[] \textbf{end for}
\end{enumerate}
\normalsize
\end{algorithm}

\section{Evaluation based on experimental and simulated data}
\label{sec:oriEst-orientationEstimation}
In this section, we apply the algorithms described in \Sectionref{sec:oriEst-smoothingOpt}--\Sectionref{sec:oriEst-compl} to both simulated and experimental data. Some general characteristics of the orientation estimation algorithms will be illustrated and the quality of the different algorithms will be analyzed. The simulated data allows for controlled analysis of the workings of the algorithms. Furthermore, it allows us to compare the different algorithms using Monte Carlo simulations. The experimental data shows the applicability to real-world scenarios. We will start by introducing the data sets. 

\begin{figure}
    \centering
  \includegraphics[scale = 1]{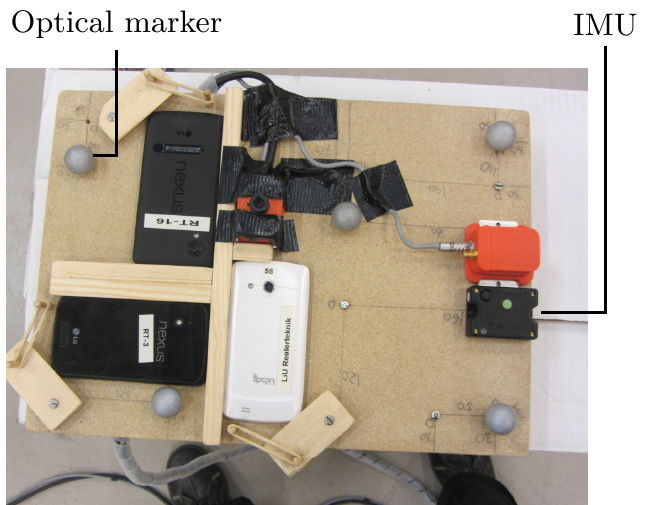}
  \caption{Experimental setup where an \gls{imu} is used to collected inertial and magnetometer measurements. Optical markers are tracked using multiple cameras, leading to accurate reference position and orientation estimates. Note that the experimental setup also contains additional \glspl{imu} and smartphones. The data from these sensors is not considered in this work.}
  \label{fig:oriEst-expSetup}
\end{figure}

\begin{figure}
	\centering
	\includegraphics[scale = 1]{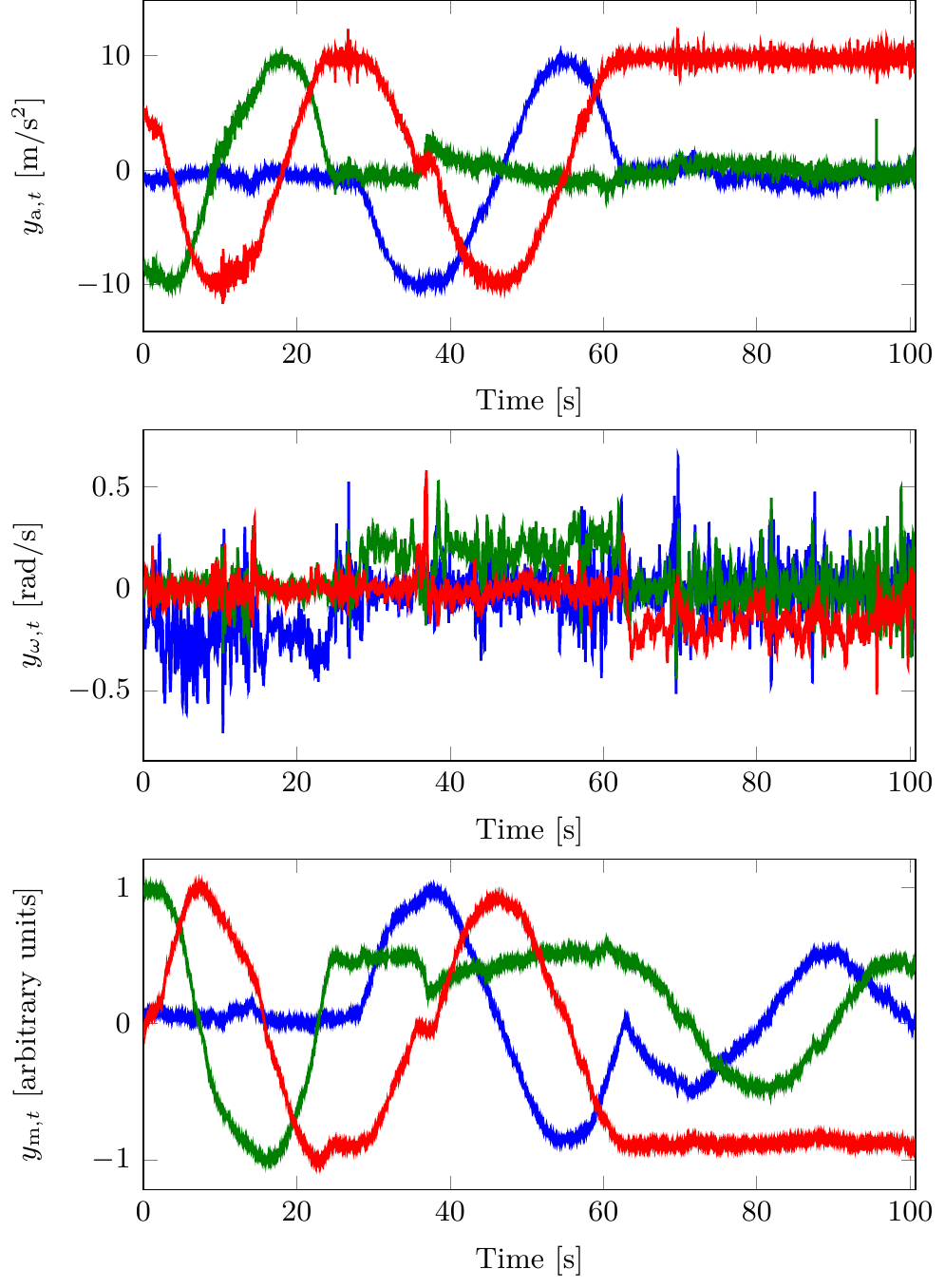}
    	\caption{Measurements from an accelerometer ($y_{\text{a},t}$, top), a gyroscope ($y_{\omega,t}$, middle) and a magnetometer ($y_{\text{m},t}$, bottom) for $100$ seconds of data collected with the \gls{imu} shown in \Figureref{fig:oriEst-expSetup}.}
	\label{fig:oriEst-expDataViconLab}
\end{figure}

Experimental data is collected using the setup shown in \Figureref{fig:oriEst-expSetup}, where data is collected using multiple mobile \glspl{imu} and smartphones. The algorithms presented in this section can be applied to measurements from any of these devices. However, we focus our analysis on the data from the Trivisio Colibri Wireless \gls{imu}~\citep{trivisio-tutorial}. In \Figureref{fig:oriEst-expDataViconLab}, the inertial and magnetometer measurements from this \gls{imu} are displayed for around $100$ seconds during which the sensor is rotated around all three axes. The experiments are performed in a lab equipped with multiple cameras~\citep{vicon-tutorial}, able to track the optical markers shown in \Figureref{fig:oriEst-expSetup}. This provides highly accurate ground truth reference position and orientation information, against which we can compare our estimates. For comparison, the optical and \gls{imu} data need to be time-synchronized and aligned. We synchronize the data by correlating the norms of the gyroscope measurements and of the angular velocity estimated by the optical system. Alignment is done using the orientation estimates in combination with Theorem 4.2 from \cite{hol:2011}.

\begin{figure}
	\centering
	\includegraphics[scale = 1]{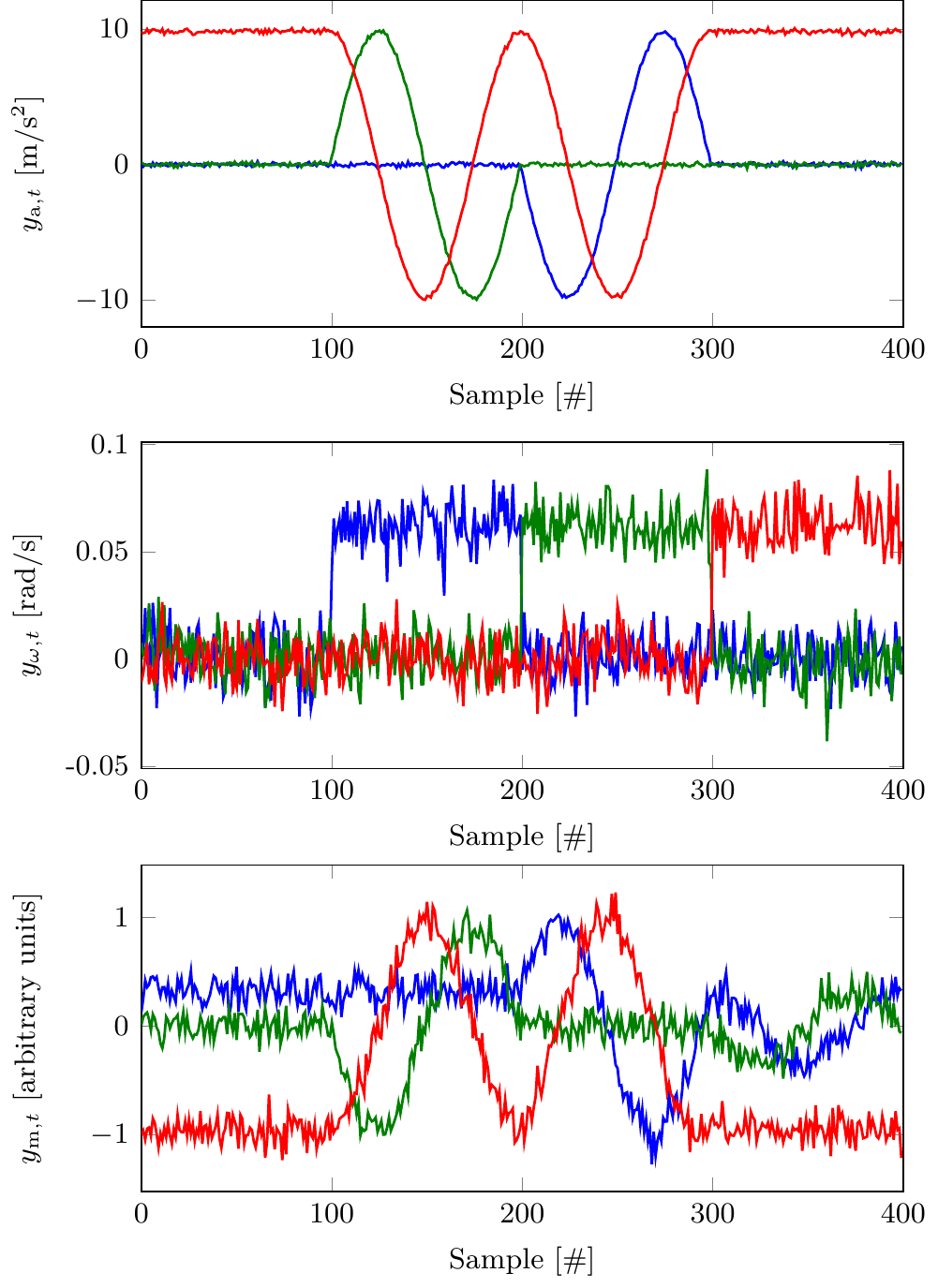}
    	\caption{Simulated measurements from an accelerometer ($y_{\text{a},t}$, top), a gyroscope ($y_{\omega,t}$, middle) and a magnetometer ($y_{\text{m},t}$, bottom).}
	\label{fig:oriEst-simData}	
\end{figure}

In \Figureref{fig:oriEst-simData}, simulated inertial and magnetometer measurements are displayed. The data represents a sensor that is kept stationary for 100 samples, after which it is rotated around all three axes. The sensor is assumed to be rotated around the origin of the accelerometer triad. Hence, during the entire data set, the accelerometer is assumed to only measure the gravity vector. The magnitude of the simulated gravity vector is $9.82~\metrepersquaresecond$. The magnitude of the simulated local magnetic field is equal to one. Its direction is approximately equal to that in Link\"oping, Sweden, where a dip angle of $71^\circ$ leads to a magnetic field  $m^\text{n} = \begin{pmatrix} 0.33 & 0 & - 0.95 \end{pmatrix}^\Transp$. The simulated noise levels are
\begin{align*}
e_{\text{a},t} &\sim \mathcal{N}(0, \sigma_\text{a}^2 \, \mathcal{I}_3), \qquad & \sigma_\text{a} &= 1 \cdot 10^{-1}, \\
e_{\omega,t} &\sim \mathcal{N}(0, \sigma_\omega^2 \, \mathcal{I}_3), \qquad & \sigma_\omega &= 1 \cdot 10^{-2}, \\
e_{\text{m},t} &\sim \mathcal{N}(0, \sigma_\text{m}^2 \, \mathcal{I}_3), \qquad & \sigma_\text{m} &= 1 \cdot 10^{-1}. 
\end{align*}
Note that we deliberately chose the noise levels to be fairly high, to clearly illustrate the workings of the different algorithms. 

Although our algorithms parametrize the orientation as quaternions, it is typically more intuitive to visualize the orientation estimates in Euler angles. Hence, we visualize our results in terms of roll, pitch and heading (yaw) angles. Both for the experimental data and for the simulated data, we are able to compare our estimates $\hat{q}^{\text{nb}}_t$ to reference orientations denoted $q^{\text{nb}}_{\text{ref},t}$. To represent the orientation error, we compute a difference quaternion $\Delta q_t$ as
\begin{align}
\Delta q_t = \hat{q}^{\text{nb}}_t \odot \left( q_{\text{ref},t}^\text{nb} \right)^\conj,
\end{align}
which can be converted to Euler angles for visualization. Note that using this definition, the orientation errors in Euler angles can be interpreted as the errors in roll, pitch and heading.

\subsection{General characteristics}
In this section, we will discuss some general characteristics of the orientation estimation problem and illustrate them in three different examples. Our goal is not to compare the different estimation algorithms, but to illustrate some characteristics common to all of them. 

In \Exampleref{ex:oriEst-simOriErrors} we focus on the accuracy of the orientation estimates that can be obtained if the state space model~\eqref{eq:models-ssOri} is completely true. We illustrate that it is typically easier to obtain accurate roll and pitch estimates than it is to obtain accurate heading estimates. 

\begin{figure}[t]
	\centering
	\includegraphics[scale = 1]{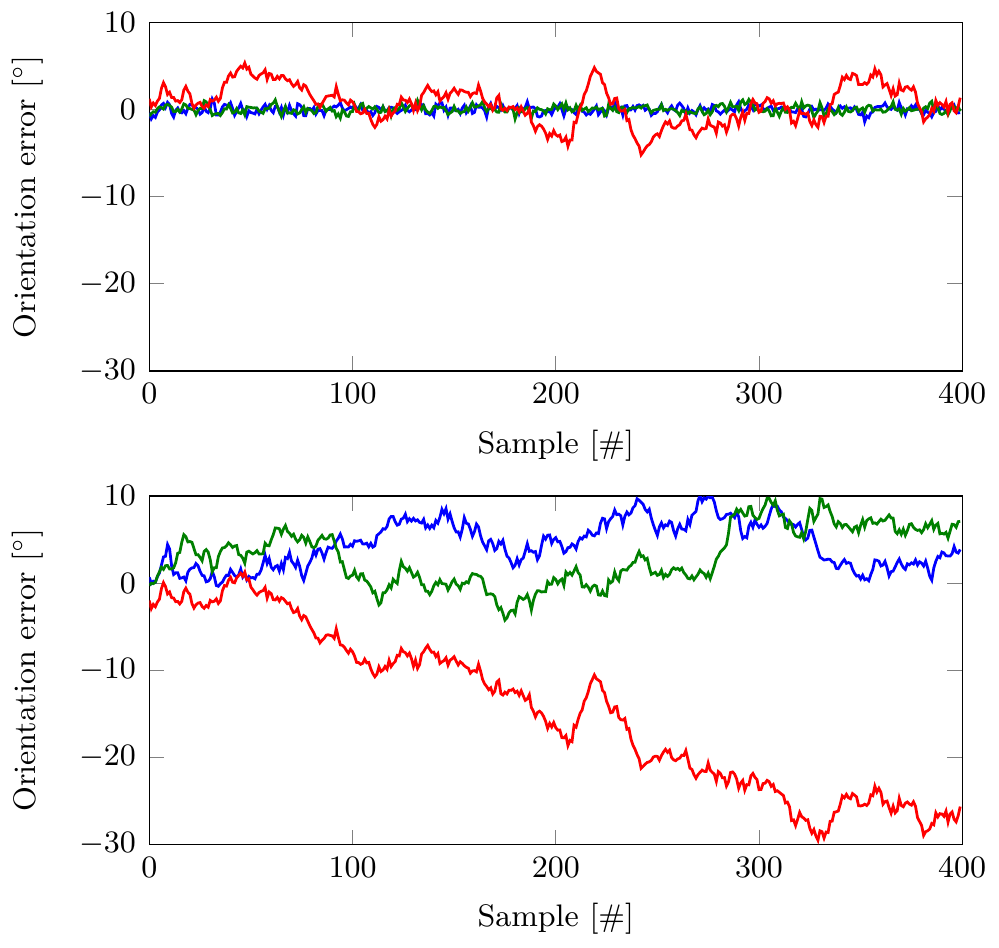}
    	\caption{Orientation errors in roll (blue), pitch (green) and heading (red) using simulated measurements for (top) \Algorithmref{alg:oriEst-smoothingOpt} using inertial and magnetometer measurements and (bottom) dead-reckoning of the gyroscope measurements.}
	\label{fig:oriEst-simOriErrors}
\end{figure}

\begin{myexample}{Orientation estimation using inertial and magnetometer data}%
\label{ex:oriEst-simOriErrors}%
The orientation errors from the smoothing optimization approach in \Algorithmref{alg:oriEst-smoothingOpt} using simulated inertial and magnetometer measurements as illustrated in \Figureref{fig:oriEst-simData} are depicted in the top plot of \Figureref{fig:oriEst-simOriErrors}. For comparison we also show the orientation errors from dead-reckoning the gyroscope measurements in the bottom plot (see also \Sectionref{sec:intro-imusForPose}). These errors can be seen to drift over time. 

Although the accelerometer and the magnetometer measurement noises are of equal magnitude, the heading angle is estimated with less accuracy compared to the roll and pitch angles. The reason for this is twofold. First, the signal to noise ratio for the magnetometer is worse than that of the accelerometer, since the magnetometer signal has a magnitude of $1$ while the accelerometer signal has a magnitude of $9.82~\metrepersquaresecond$. Second, only the horizontal component of the local magnetic field vector provides heading information. This component is fairly small due to the large dip angle ($71^\circ$) in Link\"oping, Sweden. 
\end{myexample}

The accelerometer provides inclination information, while the magnetometer provides heading information, see \Sectionref{sec:models-measModels}. In case only inertial measurements and no magnetometer measurements are available, the heading can only be estimated using the gyroscope measurements and will therefore drift over time. This is illustrated in \Exampleref{ex:oriEst-noMagData}.

\begin{myexample}{Orientation estimation using only inertial measurements}%
\label{ex:oriEst-noMagData}%
The orientation errors from the smoothing optimization approach in \Algorithmref{alg:oriEst-smoothingOpt} using simulated inertial measurements as presented in \Figureref{fig:oriEst-simData} can be found in \Figureref{fig:oriEst-noMagData}. The roll and pitch angles can be seen to be accurate, while the heading angle drifts significantly. To obtain the results in \Figureref{fig:oriEst-noMagData}, we used data with the same noise realization as in \Exampleref{ex:oriEst-simOriErrors}. Hence, we refer to \Figureref{fig:oriEst-simOriErrors} for comparison to the orientation errors from dead-reckoning the gyroscope measurements. The drift in the heading angle can be seen to be similar to the drift from dead-reckoning of the gyroscope measurements. 

\begin{figure}[t]
	\centering
	\includegraphics[scale = 1]{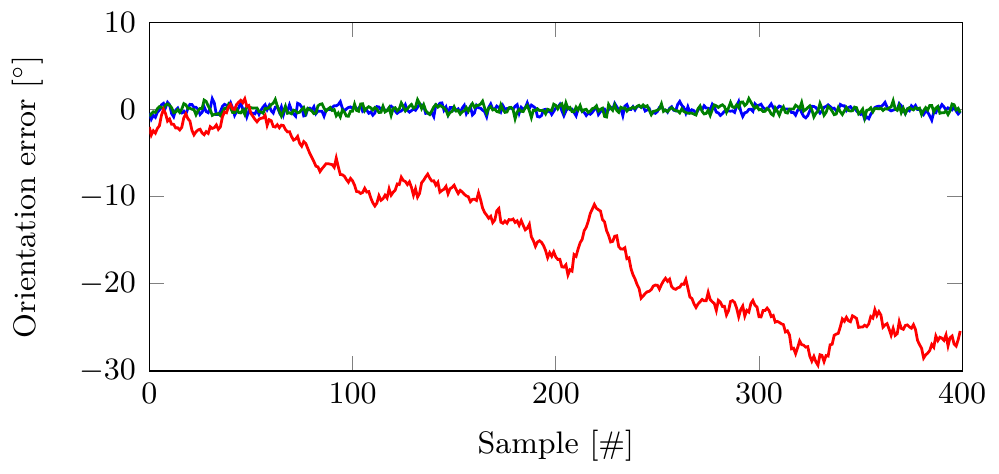}
    	\caption{Orientation errors in roll (blue), pitch (green) and heading (red) using simulated measurements for \Algorithmref{alg:oriEst-smoothingOpt} using inertial measurements only.}
	\label{fig:oriEst-noMagData}
\end{figure}
\end{myexample}

The two examples above assume that our state space model~\eqref{eq:models-ssOri} is an accurate description of the measurements. In practice, however, this is not always the case, for instance due to the presence of magnetic material in the vicinity of the sensor. In \Exampleref{ex:oriEst-magDist} we illustrate that if the state space model does not accurately describe the data, it is not possible to obtain accurate orientation estimates. 

\begin{myexample}{Orientation estimation in the presence of magnetic material}%
\label{ex:oriEst-magDist}%
We simulate 400 samples of stationary data. Between samples 150 and 250, we simulate the presence of a magnetic material, causing a change in the magnetic field of $\begin{pmatrix} 0.1 & 0.3 & 0.5 \end{pmatrix}^\Transp$. In \Figureref{fig:oriEst-magDist}, we show the adapted magnetometer data and the orientation estimates using the smoothing optimization approach from \Sectionref{sec:oriEst-smoothingOpt}. As can be seen, the heading estimates show significant errors when the magnetic material is present. Depending on the amount of disturbance and the uncertainty in the inertial and magnetometer measurements, magnetic material can also result in errors in the roll and pitch estimates. 

\begin{figure}[t]
	\centering
	\includegraphics[scale = 1]{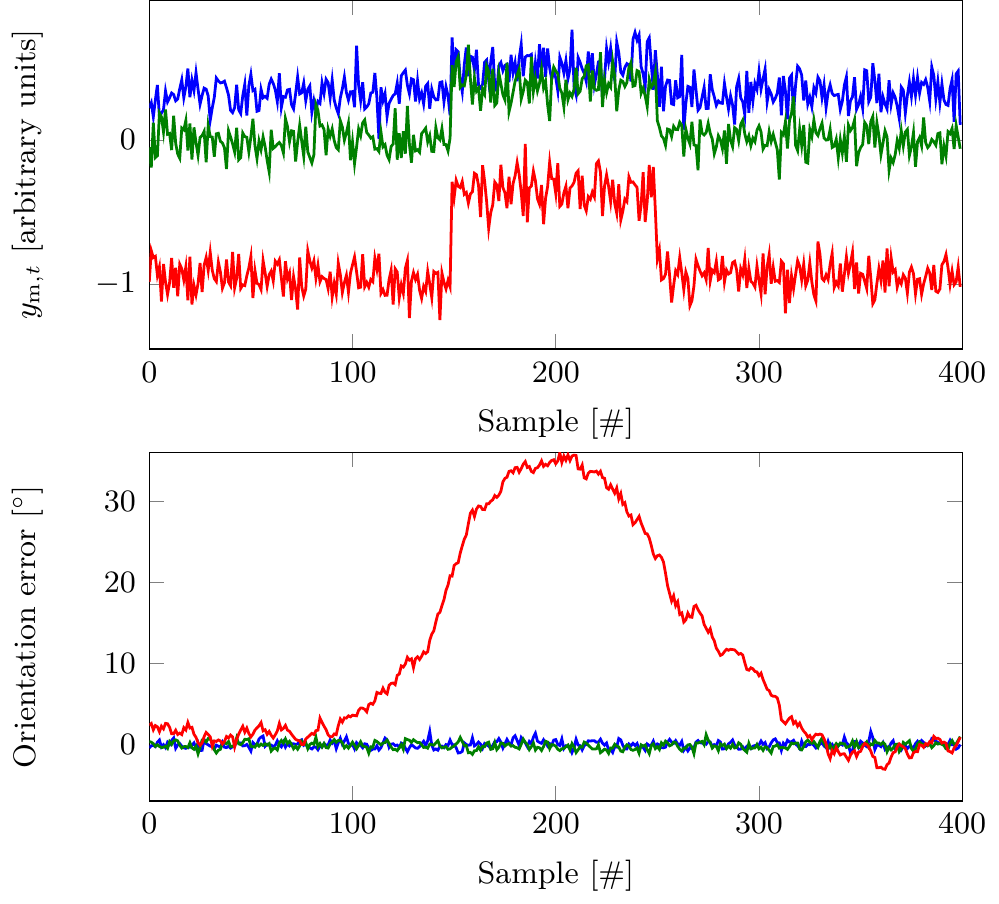}
    	\caption{Top: Simulated magnetometer measurements $y_{\text{m},t}$ for $400$ samples of stationary data. Between samples 150 and 250 we simulate the presence of a magnetic material in the vicinity of the sensor. Bottom: Orientation estimates in roll (blue), pitch (green) and heading (red) using the simulated inertial and magnetometer measurements.}
	\label{fig:oriEst-magDist}
\end{figure}
\end{myexample}

\subsection{Representing uncertainty}
\label{sec:oriEst-repreUncert}
So far, we have discussed the \emph{quality} of the orientation estimates in three different examples. However, we did not discuss the \emph{uncertainty} of the estimates. Algorithms~\ref{alg:oriEst-smoothingOpt}--\ref{alg:oriEst-ekfOriError} provide estimates of the uncertainties. We will now discuss how these uncertainties can be displayed and interpreted and highlight some difficulties with this. 

Both the optimization and the \gls{ekf} approaches discussed in \Sectionref{sec:oriEst-smoothingOpt}--\Sectionref{sec:oriEst-ekf} compute the uncertainty of the estimates in terms of a covariance matrix. Let us use the more general notation $\cov ( \hat{\eta}_t^\text{n})$ for the covariance of the orientation deviation states $\hat{\eta}_t^\text{n}$, $t = 1, \hdots, N$ computed in Algorithms~\ref{alg:oriEst-smoothingOpt},~\ref{alg:oriEst-filteringOpt} and~\ref{alg:oriEst-ekfOriError}, and $\cov ( \hat{q}_t^\text{nb})$ for the covariance of the quaternion states $\hat{q}_t^\text{nb}$, $t = 1, \hdots, N$ computed in \Algorithmref{alg:oriEst-ekfQuat}. If the states would be in normal, Euclidean space, the square root of the diagonal of these matrices would represent the standard deviation $\sigma$ of the estimates in the different directions. These could then be visualized by for instance plotting $3 \sigma$ confidence bounds around the estimates. 

\begin{figure}
	\centering
	\includegraphics[scale = 1]{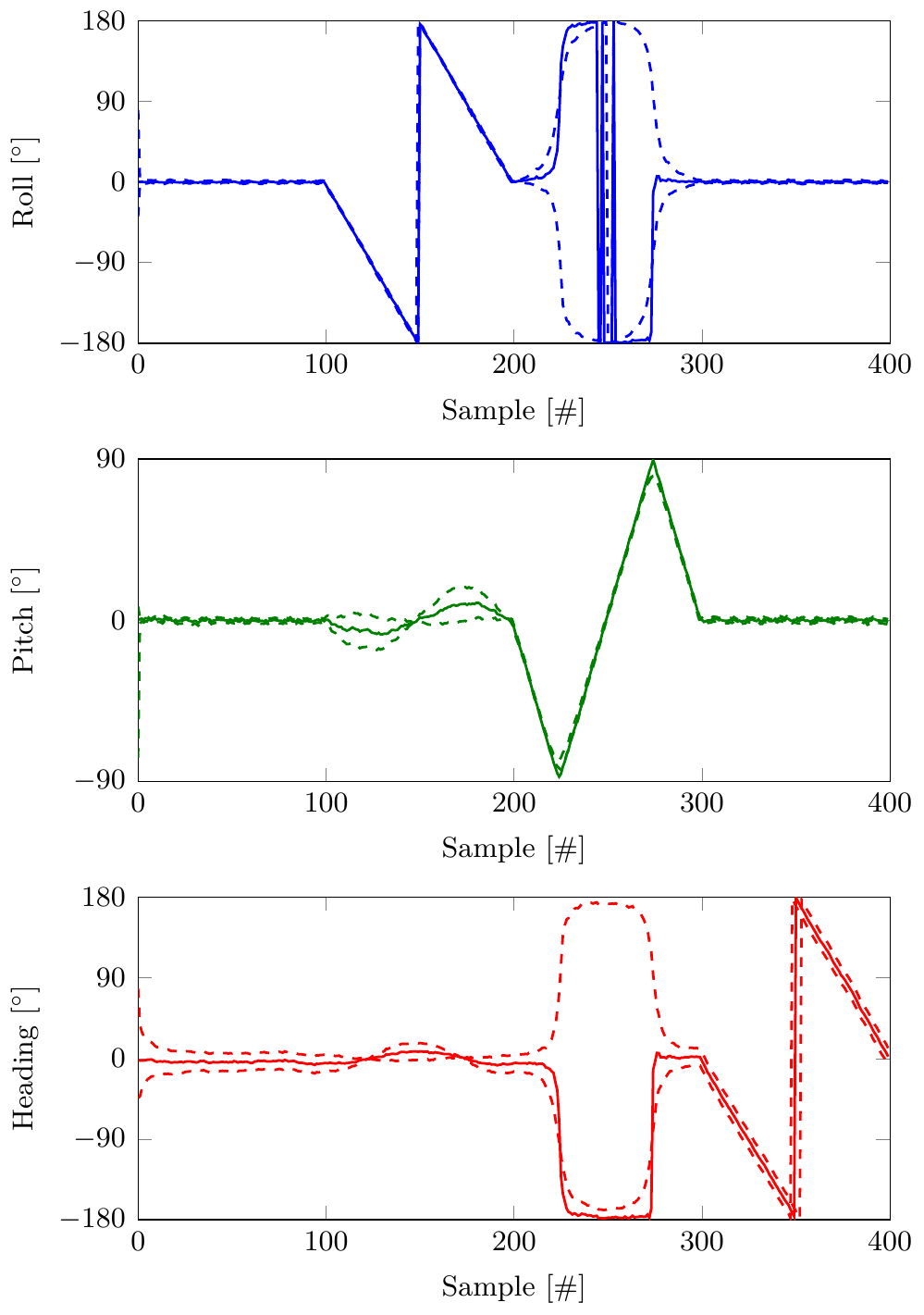}
    	\caption{Orientation estimates (solid) and $3 \sigma$ bounds (dashed) in roll (blue), pitch (green) and heading (red) using inertial and magnetometer measurements.}
	\label{fig:oriEst-eulerAngleBounds}
\end{figure}

One could imagine that an equivalent way of visualizing the orientation deviation uncertainties, would be to compute the $3 \sigma$ bounds in terms of orientation deviations in each of the three directions as
\begin{subequations}
\begin{align}
\left( \Delta \oriError_{i,t}^\text{n} \right)_{+3 \sigma} &= + 3 \sqrt{\left( \cov (\hat{\oriError}_t^\text{n} \right)_{ii}}, \quad & i &= 1, \hdots, 3, \\
\left( \Delta \oriError_{i,t}^\text{n} \right)_{-3 \sigma} &= - 3 \sqrt{\left( \cov (\hat{\oriError}_t^\text{n} \right)_{ii}}, \quad & i &= 1, \hdots, 3,
\end{align}
\end{subequations}
after which the bounds can be parametrized in terms of quaternions as
\begin{subequations}
\begin{align}
\left( q_t^\text{nb} \right)_{+3 \sigma} = \expq \left( \Delta \oriError_{t}^\text{n} \right)_{+3 \sigma} \odot \hat{q}_t^\text{nb}, \\
\left( q_t^\text{nb} \right)_{-3 \sigma} = \expq \left( \Delta \oriError_{t}^\text{n} \right)_{-3 \sigma} \odot \hat{q}_t^\text{nb}.
\end{align}
\end{subequations}
The resulting estimates and bounds are visualized in terms of Euler angles in \Figureref{fig:oriEst-eulerAngleBounds} for simulated data similar to the data presented in \Figureref{fig:oriEst-simData}. The orientation and its covariance are estimated using \Algorithmref{alg:oriEst-ekfOriError}. As can be seen, the bounds are difficult to interpret due to the wrapping of the Euler angles. 

As argued in~\cite{forsterCDS:2016}, it is more intuitive to directly represent the uncertainty in terms of orientation deviations. The covariance $\cov ( \hat{\oriError}_t^\text{n})$ can be interpreted as the uncertainty in the roll, pitch and heading angles as illustrated in \Exampleref{ex:oriEst-noMagDataCov}.

\begin{myexample}{Orientation estimation using only inertial measurements (continued)}
\label{ex:oriEst-noMagDataCov}%
Since the accelerometer provides only inclination information, in the case of \Exampleref{ex:oriEst-noMagData} where magnetometer measurements are unavailable, we expect only the roll and pitch angles to be estimated with small uncertainty. In fact, we expect the uncertainty of the heading at $t = 1$ to be equal to the uncertainty of the initial $\Sigma_\text{i}$ from \Sectionref{sec:models-prior} and to steadily grow over time, depending on the amount of gyroscope noise. In \Figureref{fig:oriEst-uncertaintySmoothingNoMag}, we plot the standard deviation $\sigma$ of the orientation estimates computed using the smoothing algorithm from \Sectionref{sec:oriEst-smoothingOpt} as the square root of the diagonal elements of $\cov ( \hat{\eta}_t^\text{n})$. As can be seen, the standard deviation of the yaw angle at $t = 1$ is indeed $20^\circ$ as modeled in \Sectionref{sec:models-prior}. The increase in the uncertainty in the yaw angle exactly matches the increase of the uncertainty due to dead-reckoning. 

\begin{figure}
	\centering
	\includegraphics[scale = 1]{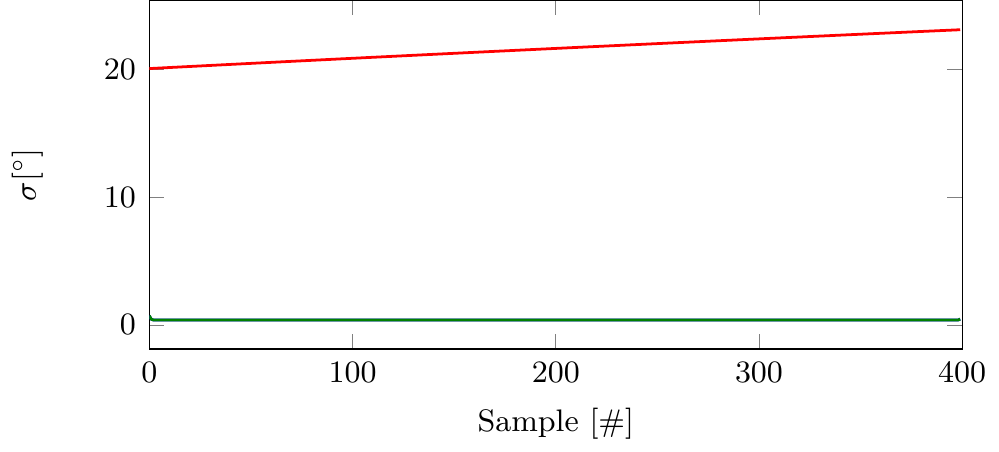}
    	\caption{Standard deviation $\sigma$ in degrees of the orientation estimates in roll (blue), pitch (green) and heading (red) using only inertial measurements.}
	\label{fig:oriEst-uncertaintySmoothingNoMag}
\end{figure}
\end{myexample}

From \Exampleref{ex:oriEst-noMagDataCov} it can be concluded that $\cov ( \hat{\oriError}_t^\text{n})$ seems to be an intuitive measure of the uncertainty of the orientation estimates. The covariance $\cov (\hat{q}_t^\text{nb})$ computed by \Algorithmref{alg:oriEst-ekfQuat} relates to $\cov ( \hat{\oriError}_t^\text{n})$ as
\begin{subequations}
\label{eq:oriEst-convertCovqn}
\begin{align}
\cov (\hat{q}_t^\text{nb}) &= \cov \left(\expq (\tfrac{\hat{\oriError}_t^\text{n}}{2} ) \odot \tilde{q}_t^\text{nb} \right) \nonumber \\
&= \tfrac{1}{4} \left( \tilde{q}_t^\text{nb} \right)^\rightMult \tfrac{\diff \expq(\hat{\oriError}_t^\text{n})}{\diff \hat{\oriError}_t^\text{n}} \cov (\hat{\oriError}_t^\text{n}) \left( \tfrac{\diff \expq(\hat{\oriError}_t^\text{n})}{\diff \hat{\oriError}_t^\text{n}} \right)^\Transp \left( \tilde{q}_t^\text{bn} \right)^\rightMult, \\
\cov (\hat{\oriError}_t^\text{n}) &= \cov \left( 2 \logq ( \hat{q}_t^\text{nb} \odot \tilde{q}_t^\text{bn}  ) \right) \nonumber \\
&= 4 \tfrac{\diff \logq (q)}{\diff q} \left( \tilde{q}_t^\text{bn} \right)^\rightMult \cov ( \hat{q}_t^\text{nb} ) \left( \tilde{q}_t^\text{nb} \right)^\rightMult \left( \tfrac{\diff \logq (q)}{\diff q} \right)^\Transp.
\end{align}
\end{subequations}
The relation~\eqref{eq:oriEst-convertCovqn} allows us to compare the orientation estimates and covariances from the different algorithms in more detail in \Exampleref{ex:oriEst-noMagDataCovComp}. 

\begin{myexample}{Orientation estimation using only inertial measurements (continued)}
\label{ex:oriEst-noMagDataCovComp}
As discussed in \Exampleref{ex:oriEst-noMagDataCov}, using only inertial measurements and no magnetometer measurements, we can only expect to be able to accurately estimate the inclination. The uncertainty of the heading estimates grows over time. We will now analyze the behavior of the different algorithms in more detail for this specific example. In \Tableref{tab:oriEst-rmsNoMagData}, we show the \gls{rmse} values over $100$ Monte Carlo simulations for Algorithms~\ref{alg:oriEst-smoothingOpt}--\ref{alg:oriEst-ekfOriError}. In \Figureref{fig:oriEst-noMagDataEstAll}, we also represent the orientation estimates from the four algorithms for one of these realizations. As can be seen from both \Tableref{tab:oriEst-rmsNoMagData} and \Figureref{fig:oriEst-noMagDataEstAll}, as expected, the smoothing algorithm outperforms the other algorithms. However, more surprisingly, the \gls{ekf} with quaternion states has much larger errors in the heading angle. In \Figureref{fig:oriEst-noMagDataCovAll}, we also show the standard deviations of the estimates from all algorithms. As can be seen, the \gls{ekf} with quaternion states over-estimates its confidence in the estimates of the heading direction. This can most likely be attributed to linearization issues. 

\begin{table}
\caption{Mean \gls{rmse} values over $100$ Monte Carlo simulations estimating orientation using only inertial measurements.}
\label{tab:oriEst-rmsNoMagData}
\begin{center}
\small
\begin{tabular}{lccc}
\toprule
\Gls{rmse} & Roll [$^\circ$]& Pitch [$^\circ$] & Heading [$^\circ$] \\
\midrule
Smoothing optimization (\Algref{alg:oriEst-smoothingOpt}) & 0.39 & 0.39 & 7.46 \\
Filtering optimization (\Algref{alg:oriEst-filteringOpt}) & 0.46 & 0.46 & 7.46 \\
EKF quaternions (\Algref{alg:oriEst-ekfQuat}) & 0.46 & 0.46 & 17.49 \\
EKF orientation deviation (\Algref{alg:oriEst-ekfOriError}) & 0.46 & 0.46 & 7.46 \\
\bottomrule
\end{tabular}
\normalsize
\end{center}
\end{table}

\begin{figure}[t]
	\centering
  	\subfigure[Smoothing optimization (\Algref{alg:oriEst-smoothingOpt}).]{
		\includegraphics[scale = 1]{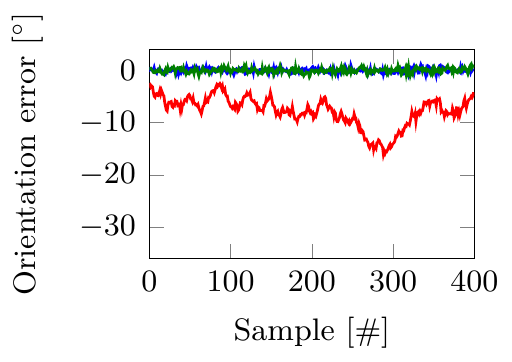}
		}
	\subfigure[Filtering optimization (\Algref{alg:oriEst-filteringOpt}).]{
		\includegraphics[scale = 1]{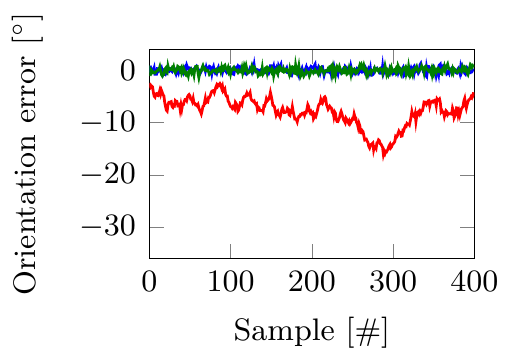}
		} \\
	\subfigure[EKF quaternions (\Algref{alg:oriEst-ekfQuat})]{
		\includegraphics[scale = 1]{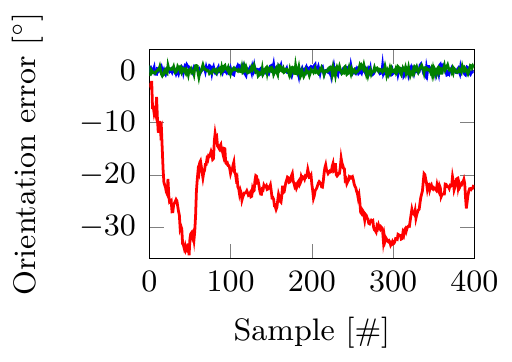}
		}
	\subfigure[EKF orientation deviation (\Algref{alg:oriEst-ekfOriError}).]{
		\includegraphics[scale = 1]{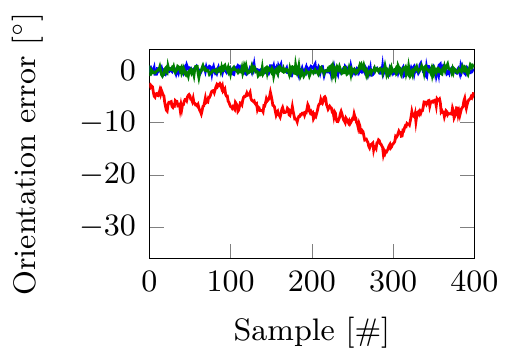}
		}
  \caption{Orientation estimates of Algorithms~\ref{alg:oriEst-smoothingOpt}--\ref{alg:oriEst-ekfOriError} in roll (blue), pitch (green) and heading (red) using only inertial measurements.}
  \label{fig:oriEst-noMagDataEstAll}
\end{figure}

\begin{figure}[t]
	\centering
  	\subfigure[Smoothing optimization (\Algref{alg:oriEst-smoothingOpt}).]{
		\includegraphics[scale = 1]{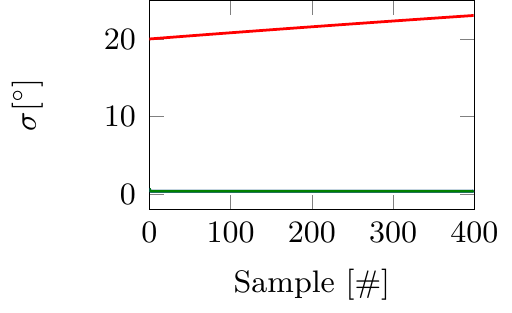}
		}
	\subfigure[Filtering optimization (\Algref{alg:oriEst-filteringOpt}).]{
		\includegraphics[scale = 1]{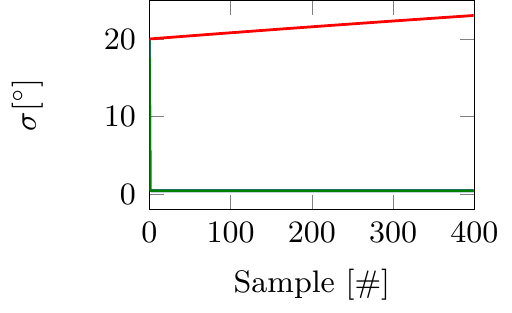}
		} \\
	\subfigure[EKF quaternions (\Algref{alg:oriEst-ekfQuat}).]{
		\includegraphics[scale = 1]{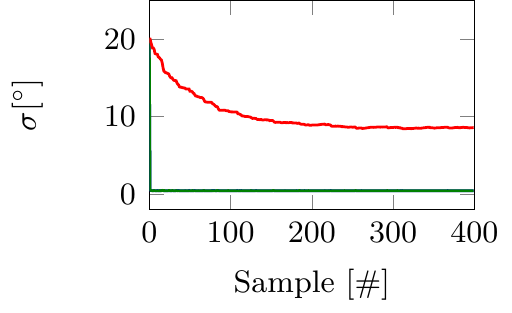}
		}
	\subfigure[EKF orientation deviation (\Algref{alg:oriEst-ekfOriError}).]{
		\includegraphics[scale = 1]{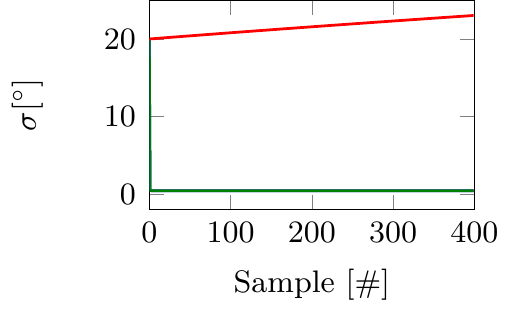}
		}
  \caption{Standard deviation $\sigma$ in degrees of the orientation estimates of Algorithms~\ref{alg:oriEst-smoothingOpt}--\ref{alg:oriEst-ekfOriError} in roll (blue), pitch (green) and heading (red) using only inertial measurements.}
  \label{fig:oriEst-noMagDataCovAll}
\end{figure}
\end{myexample}

From \Exampleref{ex:oriEst-noMagDataCovComp}, it can be concluded that not properly estimating the covariance can have negative effects on the quality of the estimates. It can also be concluded from this section that covariances can best be represented in terms of $\cov(\hat{\oriError}_t^\text{n})$ but that they are difficult to visualize in Euler angles. Because of that, in the remainder of this section, we will typically plot the uncertainty in a separate plot like in \Figureref{fig:oriEst-noMagDataCovAll}.

\subsection{Comparing the different algorithms}
\begin{figure}
	\centering
	\includegraphics[scale = 1]{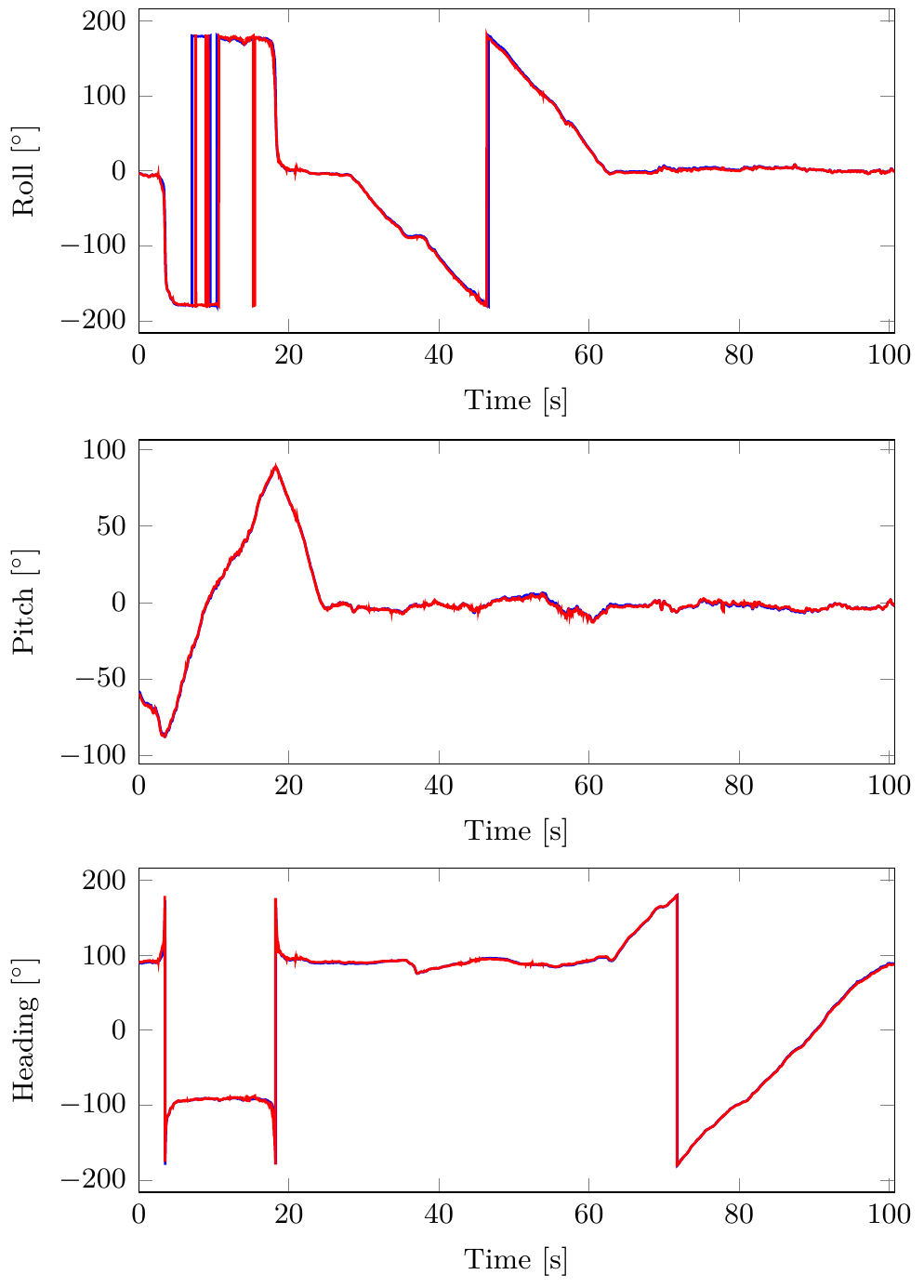}
    	\caption{Red: Orientation from the optical reference system. Blue: Orientation estimates obtained using \Algorithmref{alg:oriEst-smoothingOpt} for the experimental data from \Figureref{fig:oriEst-expDataViconLab}.}
	\label{fig:oriEst-oriExpResults}
\end{figure}

We use the experimental data presented in \Figureref{fig:oriEst-expDataViconLab} to assess the quality of the estimates from the different algorithms. We also have a short sequence of stationary data available that allows us to determine the gyroscope bias and the sensor noise covariances. The gyroscope bias is subtracted from the data to allow for better orientation estimates. The standard deviation of the gyroscope noise is experimentally determined to be $\sigma_\omega = 4.9 \cdot 10^{-3}$. As discussed in \Sectionref{sec:models-measModels}, the covariance matrices in the measurement models reflect not only the sensor noise, but also the model uncertainty. Both the assumptions that the sensor's acceleration is zero and that the magnetic field is homogeneous are only approximately true. Because of this, we choose $\sigma_\text{a} = 2.6 \cdot 10^{-1}$ and $\sigma_\text{m} = 2.5 \cdot 10^{-1}$. These values are a factor 10 respectively 100 larger than the experimentally determined noise values. Note that this choice is highly dependent on the data set. The \gls{rmse} values as compared to the optical reference system for the different methods described in this chapter are summarized in \Tableref{tab:oriEst-rmsExp}. A good value for $\alpha$ in the complementary filter is experimentally determined to be $0.001$ for this specific data set. As an illustration of the estimates, the orientation estimates as obtained using the smoothing algorithm and the orientations from the optical reference system are shown in \Figureref{fig:oriEst-oriExpResults}. 

It is of course difficult to draw quantitative conclusions based on only one data set. The \gls{rmse} values in \Tableref{tab:oriEst-rmsExp} should therefore mainly be seen as an indication of the performance. Comparing the different algorithms amongst each other is hard. In fact, the algorithms perform differently with respect to each other for different choices of the covariance matrices. Because of this, we will study the accuracy of the different methods using simulated data instead. 

\begin{table}
\caption{RMSE of the orientation estimates obtained using Algorithms~\ref{alg:oriEst-smoothingOpt}--\ref{alg:oriEst-compl} and the experimental data presented in \Figureref{fig:oriEst-expDataViconLab}.}
\label{tab:oriEst-rmsExp}
\begin{center}
\small
\begin{tabular}{lccc}
\toprule
\Gls{rmse} & Roll [$^\circ$]& Pitch [$^\circ$] & Heading [$^\circ$] \\
\midrule
Smoothing optimization (\Algref{alg:oriEst-smoothingOpt}) & 1.03 & 0.48 & 0.81 \\
Filtering optimization (\Algref{alg:oriEst-filteringOpt}) & 1.14 & 0.56 & 1.28 \\
EKF quaternions (\Algref{alg:oriEst-ekfQuat}) & 1.13 & 0.52 & 1.04 \\
EKF orientation deviation (\Algref{alg:oriEst-ekfOriError}) & 1.14 & 0.56 & 1.28 \\
Complementary filter (\Algref{alg:oriEst-compl}) & 0.95 & 0.62 & 1.55 \\
\bottomrule
\end{tabular}
\normalsize
\end{center}
\end{table}

We run 100 Monte Carlo simulations where the simulated data illustrated in \Figureref{fig:oriEst-simData} is generated with different noise realizations.
\Tableref{tab:oriEst-rmsSim} shows the mean \gls{rmse} for the five estimation algorithms. Algorithms~\ref{alg:oriEst-smoothingOpt}--\ref{alg:oriEst-ekfOriError} use the noise covariance matrices that are also used to generate the data. The complementary filter also uses these to determine the orientation from the accelerometer and magnetometer data. To combine this with the orientation from the gyroscope measurements, we need to determine a good value for $\alpha$. We empirically choose a value that leads to good performance for this data set. The orientation from the accelerometer and magnetometer measurements can be expected to be more accurate in the roll and the pitch than in the heading, see \Exampleref{ex:oriEst-simOriErrors}. Since $\alpha$ is scalar, however, it is not possible to weigh these contributions differently. Because of this, we present results both for $\alpha = 0.07$, which leads to small \gls{rmse} values in the heading but relatively larger \gls{rmse} in the roll and the pitch, and for $\alpha = 0.7$, which leads to small \gls{rmse} values in the roll and pitch but large \gls{rmse} in the heading. 

\begin{figure}
	\centering
  	\subfigure[Smoothing optimization (\Algref{alg:oriEst-smoothingOpt}).]{
		\includegraphics[scale = 1]{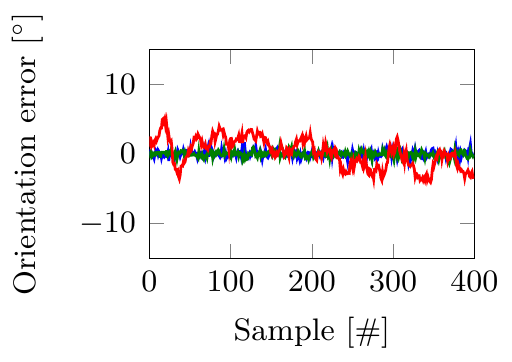}
		}
	\subfigure[Filtering optimization (\Algref{alg:oriEst-filteringOpt}).]{
		\includegraphics[scale = 1]{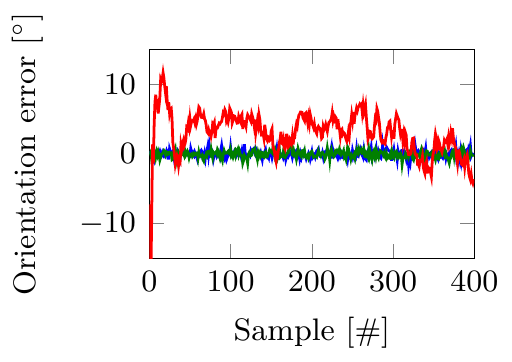}
		} \\
	\subfigure[EKF quaternions (\Algref{alg:oriEst-ekfQuat}).]{
		\includegraphics[scale = 1]{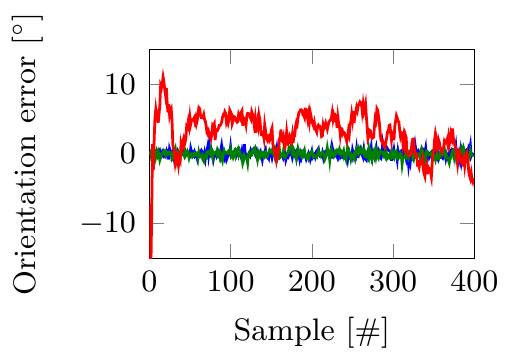}
		}
	\subfigure[EKF orientation deviation (\Algref{alg:oriEst-ekfOriError}).]{
		\includegraphics[scale = 1]{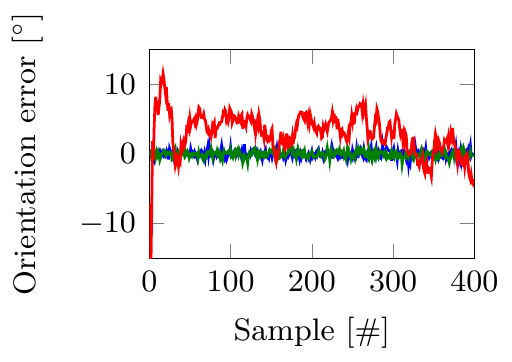}
		}
	\subfigure[Complementary filter (\Algref{alg:oriEst-compl}), $\alpha = 0.07$.]{
		\includegraphics[scale = 1]{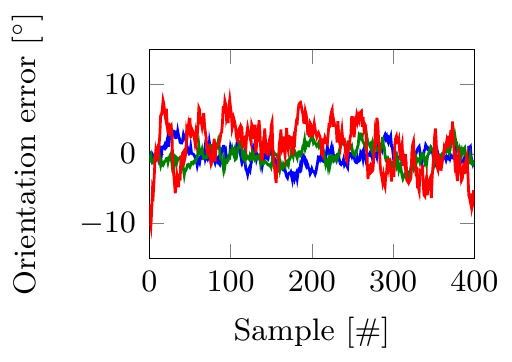}
		}
	\subfigure[Complementary filter (\Algref{alg:oriEst-compl}), $\alpha = 0.7$.]{
		\includegraphics[scale = 1]{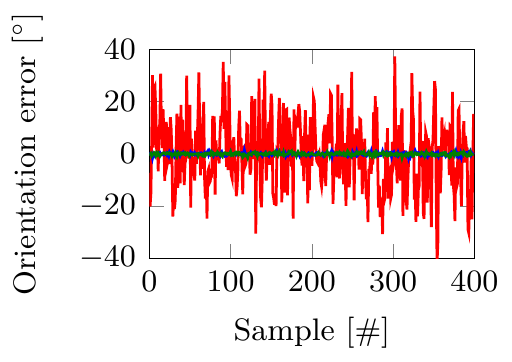}
		}		
  \caption{Orientation errors of Algorithms~\ref{alg:oriEst-smoothingOpt}--\ref{alg:oriEst-compl} in roll (blue), pitch (green) and heading (red) using simulated inertial and magnetometer measurements.}
  \label{fig:oriEst-oriSimResultsEstAll}
\end{figure}

\begin{figure}[t]
	\centering
  	\subfigure[Smoothing optimization (\Algref{alg:oriEst-smoothingOpt}).]{
		\includegraphics[scale = 1]{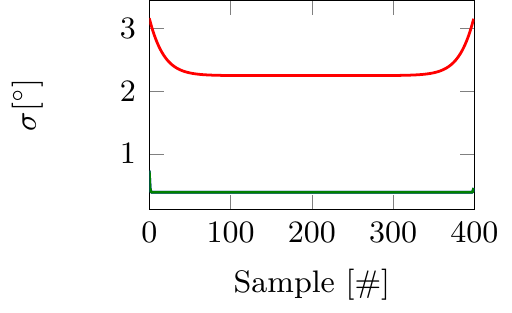}
		}
	\subfigure[Filtering optimization (\Algref{alg:oriEst-filteringOpt}).]{
		\includegraphics[scale = 1]{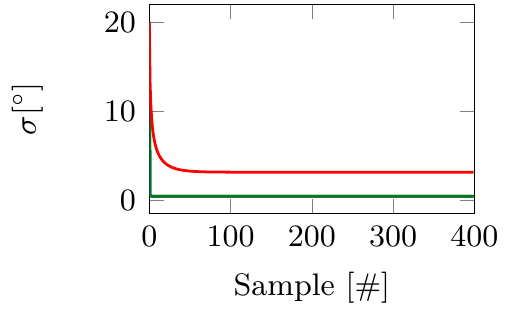}
		} \\
	\subfigure[EKF quaternions (\Algref{alg:oriEst-ekfQuat}).]{
		\includegraphics[scale = 1]{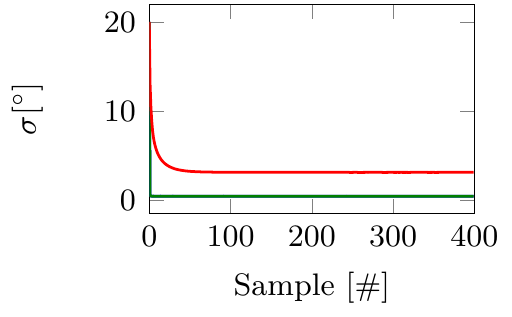}
		}
	\subfigure[EKF orientation deviation (\Algref{alg:oriEst-ekfOriError}).]{
		\includegraphics[scale = 1]{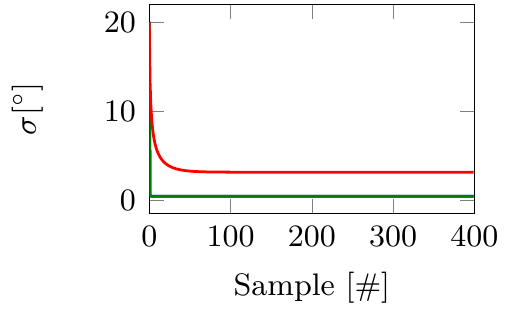}
		}
  \caption{Standard deviation $\sigma$ in degrees of the orientation estimates of Algorithms~\ref{alg:oriEst-smoothingOpt}--\ref{alg:oriEst-ekfOriError} in roll (blue), pitch (green) and heading (red) using simulated inertial and magnetometer measurements. Note the different scale on the vertical axis of (a) as compared to (b)--(d).}
  \label{fig:oriEst-oriSimResultsCovAll}
\end{figure}

From the results summarized in \Tableref{tab:oriEst-rmsSim}, it can be seen that the smoothing approach outperforms the filtering approaches. The estimates for one of the noise realizations are shown in \Figureref{fig:oriEst-oriSimResultsEstAll}. For Algorithms~\ref{alg:oriEst-smoothingOpt}--\ref{alg:oriEst-ekfOriError}, we also depict the  covariance estimates in \Figureref{fig:oriEst-oriSimResultsCovAll}. The filtering approaches from Algorithms~\ref{alg:oriEst-filteringOpt}--\ref{alg:oriEst-ekfOriError} estimate the standard deviation of the orientation errors at $t = 1$ to be equal to $20^\circ$. After this, they can be seen to converge to around $3.16^\circ$ degrees for the heading angle and $0.46^\circ$ for roll and pitch angles. The smoothing algorithm estimates an uncertainty in the heading angle of around $3.17^\circ$ for the first and last sample, while converging to a standard deviation of $2.25^\circ$ for the middle of the data set. For the roll and pitch angles, the initial and final uncertainties are estimated to be around $0.73^\circ$, converging to $0.39^\circ$ for the middle of the data set. Note that these values correspond fairly well with the \gls{rmse} values in \Tableref{tab:oriEst-rmsSim}. 

\begin{table}[t]
\caption{Mean \gls{rmse} of the orientation estimates from $100$ Monte Carlo simulations using Algorithms~\ref{alg:oriEst-smoothingOpt}--\ref{alg:oriEst-compl}.}
\label{tab:oriEst-rmsSim}
\begin{center}
\small
\begin{tabular}{lccc}
\toprule
\Gls{rmse} & Roll [$^\circ$]& Pitch [$^\circ$] & Heading [$^\circ$] \\
\midrule
Smoothing optimization (\Algref{alg:oriEst-smoothingOpt}) & 0.39 & 0.39 & 2.30 \\
Filtering optimization (\Algref{alg:oriEst-filteringOpt}) &  0.45 & 0.45 & 3.54 \\
EKF quaternions (\Algref{alg:oriEst-ekfQuat}) & 0.45 & 0.45 & 3.57 \\
EKF orientation deviation (\Algref{alg:oriEst-ekfOriError}) & 0.45 & 0.45 & 3.55 \\
Complementary filter (\Algref{alg:oriEst-compl}), $\alpha = 0.07$ & 1.44 & 1.43 & 4.39 \\
Complementary filter (\Algref{alg:oriEst-compl}), $\alpha = 0.7$ & 0.47 & 0.47 & 12.98 \\
\bottomrule
\end{tabular}
\normalsize
\end{center}
\end{table}

\begin{table}
\caption{Mean \gls{rmse} of the orientation estimates from $100$ Monte Carlo simulations. The estimate of the initial orientation is assumed to be normal distributed around the true initial orientation with a standard deviation of $20^\circ$.}
\label{tab:oriEst-rmsSimWrongInit}
\begin{center}
\small
\begin{tabular}{lccc}
\toprule
\Gls{rmse} & Roll [$^\circ$]& Pitch [$^\circ$] & Heading [$^\circ$] \\
\midrule
Smoothing optimization (\Algref{alg:oriEst-smoothingOpt}) & 0.39 & 0.39 & 2.29 \\
Filtering optimization (\Algref{alg:oriEst-filteringOpt}) &  1.06 & 0.95 & 3.55 \\
EKF quaternions (\Algref{alg:oriEst-ekfQuat}) & 1.08 & 0.97 & 4.41 \\
EKF orientation deviation (\Algref{alg:oriEst-ekfOriError}) & 1.08 & 0.96 & 3.57 \\
Complementary filter (\Algref{alg:oriEst-compl}), $\alpha = 0.07$ & 3.02 & 2.77 & 4.48 \\
Complementary filter (\Algref{alg:oriEst-compl}), $\alpha = 0.7$ & 1.13 & 1.01 & 12.99 \\
\bottomrule
\end{tabular}
\normalsize
\end{center}
\end{table}

For the Monte Carlo simulations described above, the three filtering algorithms perform similarly. However, differences can be seen when an update of the filter needs to correct the orientation estimates significantly. Examples for when this happens are when the initial orientation is not accurately known or when magnetometer measurements are not available for a longer period of time. In these cases, the uncertainty of the state is large and large corrections to the state estimates are needed when measurements become available. To analyze this case in more detail, we assume that the estimate of the initial orientation $\initEst{q}_1^\text{nb}$ is normal distributed around the true initial orientation with a standard deviation of $20^\circ$. Hence, we do not use the first accelerometer and magnetometer data for initialization. Note that a standard deviation of $20^\circ$ is equal to the uncertainty on the initial state assumed by the algorithms. The results for $100$ Monte Carlo simulations are summarized in \Tableref{tab:oriEst-rmsSimWrongInit}. As can be seen, specifically the \gls{ekf} with quaternion states and the complementary filter with $\alpha = 0.07$ perform worse than \Algorithmref{alg:oriEst-filteringOpt} and \Algorithmref{alg:oriEst-ekfOriError} for this data.

Which algorithm to use is highly application-specific. However, in general it can be concluded that all five algorithms actually produce fairly good orientation estimates, assuming that the models from \Sectionref{sec:models-resultingProbModel} are indeed valid. The smoothing algorithm performs better than the filtering approaches but it is also the most computationally expensive. The \gls{ekf} with quaternion states and the complementary filter suffer from linearization issues when large orientation corrections need to be made or when magnetometer data is unavailable.

\section{Extending to pose estimation}
\label{sec:oriEst-poseEstimation}
In \Sectionref{sec:oriEst-orientationEstimation}, we have evaluated the performance of Algorithms~\ref{alg:oriEst-smoothingOpt}--\ref{alg:oriEst-compl} for orientation estimation. The estimation methods presented in \Sectionref{sec:oriEst-smoothingOpt}--\Sectionref{sec:oriEst-ekf} can also be used to estimate the sensor's pose using the state space model~\eqref{eq:models-ssPose}. Complementary filtering is predominantly used for orientation estimation and will therefore not be considered in this section. 

The pose estimation problem can be written as a smoothing optimization problem as 
\begin{align}
\label{eq:oriEst-posSmoothing}
\hat{x}_{1:N} = \argmin_{x_{1:N}} &\underbrace{\vphantom{\sum_{t = 2}^N} \| e_{\text{p,i}} \|_{\Sigma_{\text{p,i}}^{-1}}^2 + \| e_{\text{v,i}} \|_{\Sigma_{\text{v,i}}^{-1}}^2 + \| e_{\oriError,\text{i}} \|_{\Sigma_{\oriError,\text{i}}^{-1}}^2}_{\text{Prior}} + \underbrace{\sum_{t = 2}^N \| e_{\text{p},t} \|_{\Sigma_\text{p}^{-1}}^2}_{\text{Measurement model}} + \nonumber \\
& \quad \underbrace{\sum_{t = 2}^N \| e_{\omega,t} \|_{\Sigma_\omega^{-1}}^2 + \| e_{\text{a,p},t} \|_{\Sigma_\text{a,p}^{-1}}^2 + \| e_{\text{a,v},t} \|_{\Sigma_\text{a,v}^{-1}}^2}_{\text{Dynamics}},
\end{align}
with $x_t = \begin{pmatrix} p_t^\Transp & v_t^\Transp & (\oriError_t^\text{n})^\Transp \end{pmatrix}^\Transp$ and
\begin{subequations}
\label{eq:oriEst-poseSmoothingTerms}
\begin{align}
e_\text{p,i} &= p_1^\text{n} - y_{\text{p},1}, \quad & e_{\text{p,i}} &\sim \mathcal{N}(0,\Sigma_\text{p,i}), \\
e_\text{v,i} &= v_1, \quad & e_{\text{v,i}} &\sim \mathcal{N}(0,\Sigma_\text{v,i}), \\
e_{\oriError,\text{i}} &=  2 \logq \left( q_1^\text{nb} \odot \initEst{q}_1^\text{bn} \right), \quad & e_{\text{i}} &\sim \mathcal{N}(0,\Sigma_{\oriError,\text{i}}), \\
e_{\text{p,a},t} &= \tfrac{2}{T^2} \left( p_{t+1}^\text{n} - p_t^\text{n} - T v_t^\text{n} \right) - & &\nonumber \\
 & \qquad R^\text{nb}_t y_{\text{a},t} -  g^\text{n}, \quad & e_{\text{p,a},t} &\sim \mathcal{N}(0,\Sigma_\text{a}), \label{eq:oriEst-poseSmoothingTerms-posDyn} \\
e_{\text{v,a},t} &= \tfrac{1}{T} \left( v_{t+1}^\text{n} - v_t^\text{n} \right) - R^\text{nb}_t y_{\text{a},t} -  g^\text{n}, \quad & e_{\text{v,a},t} &\sim \mathcal{N}(0,\Sigma_\text{a}), \label{eq:oriEst-poseSmoothingTerms-velDyn} \\
e_{\omega,t} &= \tfrac{2}{T} \logq \left( q_t^\text{bn} \odot q^\text{nb}_{t+1}\right) - y_{\omega,t}, \quad & e_{\omega,t} &\sim \mathcal{N}(0,\Sigma_\omega),  \\
e_{\text{p},t} &= y_{\text{p},t} - p^\text{n}_t, \quad & e_{\text{p},t} &\sim \mathcal{N}(0,\Sigma_\text{p}). 
\end{align}
\end{subequations}
In this section, we will discuss some details about the workings of the pose estimation algorithm using this model. We will not go through a complete derivation of the four algorithms. However, the adaptations that are needed to use Algorithms~\ref{alg:oriEst-smoothingOpt}--\ref{alg:oriEst-ekfOriError} for pose estimation can be found in \Appendixref{app:poseEst}. 

An important observation is that $e_{\text{a,p},t}$ and $e_{\text{a,v},t}$ in \eqref{eq:oriEst-poseSmoothingTerms-posDyn} and \eqref{eq:oriEst-poseSmoothingTerms-velDyn} depend on the orientation $R_t^\text{nb}$. Because of this, the position, velocity and orientation states are coupled. The position measurements therefore do not only provide information about the position and velocity, but also about the orientation of the sensor. This is the reason why it is no longer essential to include magnetometer data and to assume that the acceleration is approximately zero. However, the accuracy of the orientation estimates depends on the movements of the sensor. This will be illustrated below. For this, we simulate 400 samples of inertial and position measurements for a non-rotating sensor with noise levels
\begin{align*}
e_{\text{a},t} &\sim \mathcal{N}(0, \sigma_\text{a}^2 \, \mathcal{I}_3), \qquad & \sigma_\text{a} &= 1 \cdot 10^{-1}, \\
e_{\omega,t} &\sim \mathcal{N}(0, \sigma_\omega^2 \, \mathcal{I}_3), \qquad & \sigma_\omega &= 1 \cdot 10^{-2}, \\
e_{\text{p},t} &\sim \mathcal{N}(0, \sigma_\text{p}^2 \, \mathcal{I}_3), \qquad & \sigma_\text{p} &= 1 \cdot 10^{-2}.
\end{align*}
We consider four different sensor motions. The results in this section are based on the solution to the smoothing optimization problem~\eqref{eq:oriEst-posSmoothing}. First, we simulate data assuming that the sensor is stationary. For this case, the position measurements provide information about the inclination of the sensor, but not about its heading. This is illustrated in \Exampleref{ex:oriEst-poseOriError-stat}.

\begin{figure}[t]
	\centering
  	\subfigure[Stationary.]{
		\includegraphics[scale = 1]{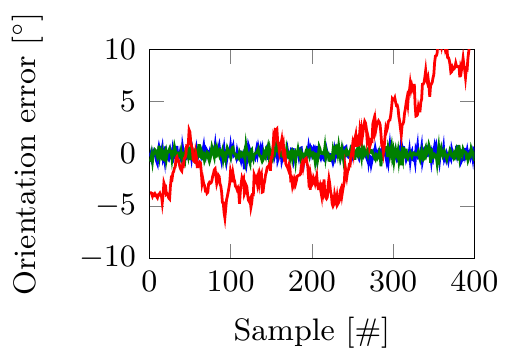}
		\label{fig:oriEst-poseOriError-stat}
		}
	\subfigure[Constant acceleration.]{
		\includegraphics[scale = 1]{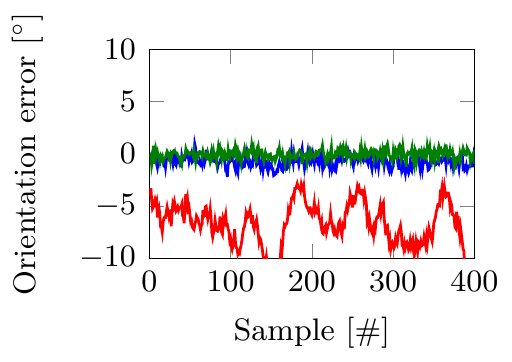}
		\label{fig:oriEst-poseOriError-const}
		}
		\\
	\subfigure[Acceleration $\left( y_{\text{a},t} \right)_y \sim \mathcal{N}(0,0.5)$.]{
		\includegraphics[scale = 1]{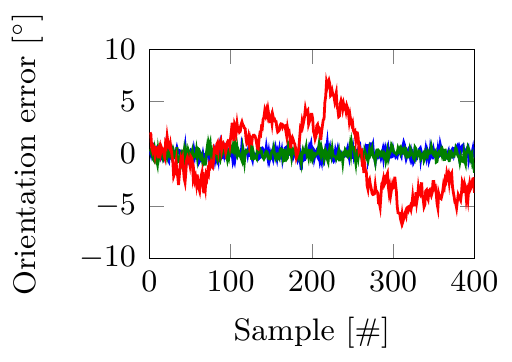}
		\label{fig:oriEst-poseOriError-rand05}
		}
	\subfigure[Acceleration $\left( y_{\text{a},t} \right)_y \sim \mathcal{N}(0,5)$.]{
		\includegraphics[scale = 1]{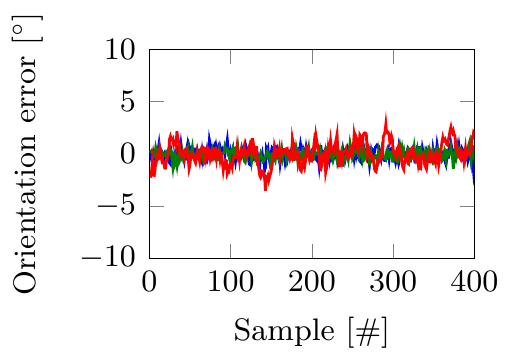}
		\label{fig:oriEst-poseOriError-rand5}
		\tikzexternalenable 
		}
  \caption{Orientation errors of the roll (blue), pitch (green) and heading (red) using simulated inertial and magnetometer measurements.}
  \label{fig:oriEst-poseOriError}
\end{figure}

\begin{myexample}{Pose estimation for a stationary sensor}%
\label{ex:oriEst-poseOriError-stat}%
We estimate the pose of a stationary sensor using simulated data and a smoothing algorithm that solves~\eqref{eq:oriEst-posSmoothing} as described in \Sectionref{sec:oriEst-smoothingOpt}. The orientation error for a specific noise realization is depicted in \Figureref{fig:oriEst-poseOriError-stat}. The inclination errors can be seen to be small, while the heading estimates drift. 
\end{myexample}

Next, in \Exampleref{ex:oriEst-poseOriError-const}, we consider the case where the sensor has a constant linear acceleration. For this case, a drift in the orientation estimates can be seen in the direction that is orthogonal to the direction of the accelerometer measurements.

\begin{myexample}{Pose estimation for a sensor with constant linear acceleration}
\label{ex:oriEst-poseOriError-const}
We estimate the pose of a sensor with an acceleration of $1~\metrepersquaresecond$ in the $y$-direction using simulated data and obtain smoothed estimates by solving~\eqref{eq:oriEst-posSmoothing}. The orientation error for a specific noise realization is depicted in \Figureref{fig:oriEst-poseOriError-const}. Again, a drift can be seen in the orientation estimates. This drift is no longer only in the heading direction, but there is also a small drift on the roll angle.  
\end{myexample}

Finally, in \Exampleref{ex:oriEst-poseOriError-rand} we consider the case of a time-varying linear acceleration. Based on simulated data, we show that accurate heading estimates can be obtained for this case. Furthermore, we show that the larger the acceleration, the more accurate the heading estimates will be. 

\begin{myexample}{Pose estimation for a sensor with time-varying linear acceleration}
\label{ex:oriEst-poseOriError-rand}
We estimate the pose of a sensor with an acceleration in the $y$-direction of $\left( y_{\text{a},t} \right)_y \sim \mathcal{N}(0,0.5)~\metrepersquaresecond$ using simulated data and compute smoothed estimates by solving~\eqref{eq:oriEst-posSmoothing}. The orientation error for a specific noise realization is depicted in \Figureref{fig:oriEst-poseOriError-rand05}. Furthermore, we simulate data with $\left( y_{\text{a},t} \right)_y \sim \mathcal{N}(0,5)~\metrepersquaresecond$. The orientation errors based on this data can be found in \Figureref{fig:oriEst-poseOriError-rand5}. As can be seen, for these cases, it is possible obtain reliable heading estimates using the state space model~\eqref{eq:models-ssPose}. The larger the acceleration, the more accurate the heading estimates.
\end{myexample}

In general, it can be concluded that it is possible to estimate position and orientation using the state space model~\eqref{eq:models-ssPose}. Except in the cases of constant or zero acceleration, it is possible to obtain drift-free orientation estimates. The heading accuracy depends on the amount of acceleration. This is summarized in \Tableref{tab:oriEst-poseOriError} where the mean \gls{rmse} of the state estimates over 100 Monte Carlo simulations is shown. Four cases were considered, inspired by Examples~\ref{ex:oriEst-poseOriError-stat}--\ref{ex:oriEst-poseOriError-rand}.

\begin{table}
\caption{Mean \gls{rmse} of the position and orientation estimates from $100$ Monte Carlo simulations. Considered are a stationary sensor, a sensor with constant acceleration and two cases of time-varying accelerations with different magnitudes.}
\label{tab:oriEst-poseOriError}
\begin{center}
\small
\begin{tabular}{lcccc}
\toprule
\Gls{rmse} & Roll [$^\circ$] & Pitch [$^\circ$] & Heading [$^\circ$] & Position [\centi\meter]\\
\midrule
Stationary & 0.41 & 0.41 & 11.84 & 0.97 \\
Constant acceleration & 1.23 & 0.46 & 11.37 & 0.97 \\ 
Acceleration & \multirow{2}{*}{0.41} & \multirow{2}{*}{0.41} & \multirow{2}{*}{2.68} & \multirow{2}{*}{0.97} \\ 
$\left( y_{\text{a},t} \right)_y \sim \mathcal{N}(0,0.5)$ & & & & \\
Acceleration & \multirow{2}{*}{0.46} & \multirow{2}{*}{0.39} & \multirow{2}{*}{0.87} & \multirow{2}{*}{0.97}  \\ 
$\left( y_{\text{a},t} \right)_y \sim \mathcal{N}(0,5)$ & & & & \\
\bottomrule
\end{tabular}
\normalsize
\end{center}
\end{table}

\chapter{Calibration}
\label{cha:calibration} 
In \Chapterref{cha:orientationEstimation}, we assumed that the sensors were properly calibrated. In practice, however, there are often calibration parameters to be taken into account. Examples of calibration parameters are the inertial sensor biases discussed in \Chapterref{cha:sensors}. Furthermore, calibration is specifically of concern when combining the inertial data with other sensors. In these cases, it is important that the inertial sensor axes and the axes of the additional sensors are aligned. Examples include using inertial sensors in combination with magnetometers~\citep{kokS:2016,salehiMB:2012,bonnetBGLB:2009} and with cameras~\citep{holSG:2010b,loboD:2007,mirzaeiR:2008}.

In this chapter we will introduce several useful calibration methods. In \Sectionref{sec:calibration-map} we explain how calibration parameters can be included as unknowns in the smoothing and filtering algorithms from \Sectionref{sec:oriEst-smoothingOpt}--\Sectionref{sec:oriEst-ekf}. This results in \gls{map} estimates of the parameters. In \Sectionref{sec:calibration-ml} we instead focus on obtaining \gls{ml} estimates of the parameters. In \Sectionref{sec:calibration-gyrBias}, the workings of the calibration algorithms are illustrated by considering the gyroscope bias to be unknown in the orientation estimation problem. Finally, in \Sectionref{sec:calibration-identifiability}, we discuss the topic of \emph{identifiability}. Parameters are said to be identifiable if they can be estimated from the available data.  

\section{Maximum a posteriori calibration}
\label{sec:calibration-map}
As discussed in \Sectionref{sec:models-probModeling}, unknown parameters $\theta$ can be estimated in the smoothing problem~\eqref{eq:models-smoothing} as
\begin{align}
\label{eq:calibration-smoothing}
\left \{ \hat{x}_{1:N},\hat{\theta} \right \} = \argmax_{x_{1:N}, \theta} p(x_{1:N}, \theta \mid y_{1:N}),
\end{align}
with 
\begin{align}
p(x_{1:N}, \theta \mid y_{1:N} ) &\propto p(\theta) p(x_1) \prod_{t=1}^N p(x_t \mid x_{t-1}, \theta) p(y_t \mid x_t, \theta).
\end{align}
Recall that a discussion on the choice of the prior of the parameters $p(\theta)$ and the states $p(x_1)$ can be found in \Sectionref{sec:models-prior}. 

Within the filtering context we typically model the parameters as slowly time-varying states. These can be estimated by solving
\begin{align}
\label{eq:calibration-filtering}
\left \{ \hat{x}_t,\hat{\theta}_t \right \} = \argmax_{x_t, \theta_t} p(x_t, \theta_t \mid y_{1:t}),
\end{align}
where
\begin{subequations}
\label{eq:calibration-filteringDist}
\begin{align}
p(x_t, \theta_t \mid y_{1:t}) &\propto p(y_t \mid x_t, \theta_t) p(x_t, \theta_t \mid y_{1:t-1} ),
\intertext{and}
p(x_t, \theta_t \mid y_{1:t-1}) &= \nonumber \\ 
& \hspace{-2cm} \iint p(x_t, \theta_t \mid x_{t-1}, \theta_{t-1}) p(x_{t-1}, \theta_{t-1} \mid y_{1:t-1}) \dint x_{t-1} \dint \theta_{t-1}.
\end{align}
\end{subequations}
Note that compared to \Sectionref{sec:models-probModeling}, in~\eqref{eq:calibration-filteringDist} we do not consider the parameters to be part of $x_t$ but instead represent them explicitly. A prior $p(\theta_1)$ on the parameters at $t = 1$ has to be included as well as a dynamic model of the parameters. 

Both the formulations~\eqref{eq:calibration-smoothing} and~\eqref{eq:calibration-filtering} compute \gls{map} estimates of the parameters. Algorithms~\ref{alg:oriEst-smoothingOpt}--\ref{alg:oriEst-ekfOriError} presented in \Chapterref{cha:orientationEstimation} can straightforwardly be extended to also estimate these unknown parameters $\theta$ or $\theta_{1:N}$. This is illustrated in \Exampleref{ex:calibration-gyrBias-map} for the case of orientation estimation in the presence of an unknown gyroscope bias. Note that \Algorithmref{alg:oriEst-compl} can also be extended to estimate an unknown gyroscope bias. We will, however, not consider this. Instead, we refer the reader to~\cite{mahonyHP:2008}.

\begin{myexample}{MAP estimates of the gyroscope bias}%
\label{ex:calibration-gyrBias-map}%
It is possible to estimate an unknown gyroscope bias in the state space model~\eqref{eq:models-ssOri}. For this, the dynamic model~\eqref{eq:models-ssOri-dyn} in the smoothing problem described in \Sectionref{sec:oriEst-smoothingOpt} is assumed to include a constant gyroscope bias $\delta_\omega$ as
\begin{subequations}
\begin{align}
q^\text{nb}_{t+1} &= q^\text{nb}_t \odot \expq \left( \tfrac{T}{2} \left( y_{\omega,t} - \delta_\omega - e_{\omega,t} \right) \right). \\
\intertext{In the filtering algorithms in \Sectionref{sec:oriEst-filteringOpt} and~\Sectionref{sec:oriEst-ekf}, the dynamic model is instead assumed to include a slowly time-varying gyroscope bias $\delta_{\omega,t}$ as}
q^\text{nb}_{t+1} &= q^\text{nb}_t \odot \expq \left( \tfrac{T}{2} \left( y_{\omega,t} - \delta_{\omega,t} - e_{\omega,t} \right) \right), \\
\intertext{where the dynamics of the gyroscope bias can be described as a random walk (see also \Sectionref{sec:models-stateDynamics})}
\delta_{\omega,t+1} &= \delta_{\omega,t} + e_{\delta_\omega,t}, \qquad e_{\delta_\omega,t} \sim \mathcal{N}(0,\Sigma_{\delta_{\omega,t}} ). \label{eq:calibration-randWalkGyrBias}
\end{align}
\end{subequations}
The smoothing algorithm presented in \Sectionref{sec:oriEst-smoothingOpt} can be extended to also estimate $\delta_\omega$. Furthermore, the filtering algorithms presented in \Sectionref{sec:oriEst-filteringOpt} and~\Sectionref{sec:oriEst-ekf} can be extended to estimate $\delta_{\omega,t}$ for $t = 1, \hdots, N$. Only minor changes to the algorithms presented in these sections are needed. These mainly concern including derivatives with respect to the additional unknowns $\delta_{\omega}$ or $\delta_{\omega,t}$. Explicit expressions for these can be found in \Appendixref{app:estGyroBias}. 
\end{myexample}

\section{Maximum likelihood calibration}%
\label{sec:calibration-ml}%
Alternatively, it is possible to obtain \gls{ml} estimates of the parameters $\theta$ as
\begin{align}
\label{eq:calibration-ml}
\hat{\theta}^{\textsc{ml}} = \argmax_{\theta\in\Theta} \mathcal{L}(\theta ; y_{1:N}).
\end{align}
Here, $\Theta \subseteq\mathbb{R}^{n_{\theta}}$ and $\mathcal{L}(\theta ; y_{1:N})$ is referred to as the likelihood function. It is defined as $\mathcal{L}(\theta ; y_{1:N}) \triangleq p_\theta(Y_{1:N} = y_{1:N})$, where $Y_{1:N}$ are random variables and $y_{1:N}$ are a particular realization of $Y_{1:N}$. Using conditional probabilities and the fact that the logarithm is a monotonic function we have the following equivalent formulation of~\eqref{eq:calibration-ml},
\begin{align}
  \label{eq:calibration-logml}
  \hat{\theta}^{\textsc{ml}} = \argmin_{\theta\in\Theta}{-\sum_{t=1}^{N}\log p_{\theta}(Y_t = y_{t} \mid
    Y_{1:t-1} = y_{1:t-1})},
\end{align}
where we use the convention that $y_{1:0} \triangleq \emptyset$. 
The \gls{ml} estimator~\eqref{eq:calibration-logml} enjoys well-understood theoretical properties including strong consistency, asymptotic normality, and asymptotic efficiency~\citep{ljung:1999}.

Due to the nonlinear nature of the orientation parametrization, our estimation problems are nonlinear, implying that there is no closed form solution available for the one
step ahead predictor $p_{\theta}(Y_t = y_{t}\mid Y_{1:t-1} = y_{1:t-1})$ in~\eqref{eq:calibration-logml}. However, similar to the filtering approaches from \Chapterref{cha:orientationEstimation}, it is possible to approximate the one step ahead predictor according to
\begin{align}
  \label{eq:calibration-oneStepPred-approx}
  p_{\theta}(Y_t = y_t \mid Y_{1:t-1} = y_{1:t-1}) \approx \mathcal{N}\left(y_t \, ; \, \hat{y}_{t\mid t-1}(\theta), S_t (\theta) \right),
\end{align}
where $\hat{y}_{t\mid t-1}(\theta)$ and $S_t(\theta)$ are defined in~\eqref{eq:oriEst-defYhatH} and~\eqref{eq:oriEst-defEpsKS}, respectively. Inserting \eqref{eq:calibration-oneStepPred-approx} into~\eqref{eq:calibration-logml} and neglecting all constants not depending on $\theta$ results in the following optimization problem,
\begin{align}
  \label{eq:calibration-MLapprox}
  \hat{\theta} = \argmin_{\theta\in\Theta} \frac{1}{2}\sum_{t=1}^{N}\|y_t - \hat{y}_{t\mid t-1}(\theta)\|_{S_t^{-1}(\theta)}^2 + \log\det S_t (\theta).
\end{align}

Unlike the optimization problems discussed so far, it is not straightforward to obtain an analytical expression of the gradient of~\eqref{eq:calibration-MLapprox}. This is because it is defined recursively through the filtering update equations. In \cite{astrom:1980} and \cite{segalW:1989}, different approaches to derive analytical expressions for objective functions of the same type as~\eqref{eq:calibration-MLapprox} are provided. They, however, consider the case of a \emph{linear} model. Some methods for obtaining \gls{ml} estimates of parameters in nonlinear models are explained in the tutorial by \cite{schonLDWNSD:2015}.

Instead of deriving analytical expressions for the gradient of~\eqref{eq:calibration-MLapprox}, it is also possible to compute a numerical approximation of the gradient. Numerical gradients can be used in a number of different optimization algorithms, such as the \gls{bfgs} method, see \eg \cite{nocedalW:2006}. Similar to the Gauss-Newton method, in the \gls{bfgs} method, the parameters are iteratively updated until convergence. However, instead of using the Hessian approximation~\eqref{eq:oriEst-approxHessian}, \gls{bfgs} iteratively estimates the Hessian using information from previous iterations. Hence, solving~\eqref{eq:calibration-MLapprox} using \gls{bfgs} with numerical gradients, requires running at least $n_\theta+1$ filtering algorithms for each iteration. These are required to evaluate the objective function and to compute the numerical gradients. More evaluations can be necessary to compute a step length, see also \Sectionref{sec:oriEst-smoothingOpt}.

\begin{myexample}{ML estimates of the gyroscope bias}%
To obtain \gls{ml} estimates of the gyroscope bias, we run the \gls{ekf} with orientation deviation states from \Algorithmref{alg:oriEst-ekfOriError} to obtain $\hat{y}_{t\mid t-1}(\delta_\omega)$ and $S_t(\delta_\omega)$ for a given value of $\delta_\omega$. This allows us to evaluate the objective function in~\eqref{eq:calibration-MLapprox}. To compute $\hat{\delta}_\omega$, the optimization problem~\eqref{eq:calibration-MLapprox} is solved iteratively using \gls{bfgs}.   
\end{myexample}

\section{Orientation estimation with an unknown gyroscope bias}
\label{sec:calibration-gyrBias}
We estimate the gyroscope bias in simulated data as described in \Sectionref{sec:oriEst-orientationEstimation} and illustrated in \Figureref{fig:oriEst-simData}. Compared to the data presented in \Sectionref{sec:oriEst-orientationEstimation}, however, a constant gyroscope bias to be estimated is added. Using Monte Carlo simulations of this data, we illustrate a few specific features of the different ways to estimate the bias. 

\begin{figure}[t]
	\centering
    	\includegraphics[scale = 1]{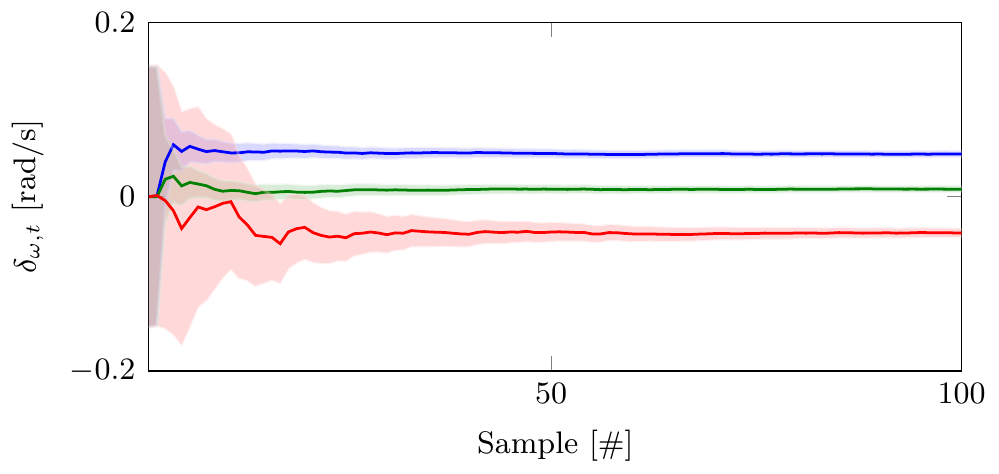}
    	\caption{Filtering estimates of the gyroscope bias and their $3 \sigma$ confidence intervals with the $x$-, $y$- and $z$-components in blue, green and red, respectively. Note that the estimates for only the first $100$ samples are shown, to focus on the period in which the estimates converge.}
	\label{fig:calibration-filteringConv}
\end{figure}

First, we focus on obtaining \gls{map} estimates of the bias using the smoothing and filtering approaches as described in \Sectionref{sec:calibration-map}. We simulate the measurement noise as described in \Sectionref{sec:oriEst-orientationEstimation} and simulate the gyroscope bias as 
\begin{align}
\label{eq:calibration-simSettingsGyrBias}
\delta_\omega \sim \mathcal{N}(0, \sigma_{\delta_\omega}^2 \mathcal{I}_3), \qquad \sigma_{\delta_\omega} = 5 \cdot 10^{-2}. 
\end{align}
Note that $\sigma_{\delta_\omega}$ is a factor 10 larger than the value discussed in \Sectionref{sec:models-prior} to clearly illustrate the effect of the presence of a gyroscope bias. The priors $p(\theta)$ and $p(\theta_1)$ in the smoothing and filtering algorithms are set equal to the distribution in~\eqref{eq:calibration-simSettingsGyrBias}. The covariance of the random walk model~\eqref{eq:calibration-randWalkGyrBias} is set as $\Sigma_{\delta_{\omega,t}} = \sigma_{\delta_{\omega,t}}^2 \mathcal{I}_3$ with $\sigma_{\delta_{\omega,t}} = 1 \cdot 10^{-10}$. This small value ensures that after convergence, the bias estimate is quite constant. The resulting mean \glspl{rmse} of the orientation over 100 Monte Carlo simulations are summarized in \Tableref{tab:calibration-rmsSim}. Since the filtering algorithms typically have similar characteristics as discussed in \Sectionref{sec:oriEst-orientationEstimation}, we only consider the \gls{ekf} with orientation deviation states here. Comparing these results to the ones presented in \Tableref{tab:oriEst-rmsSim}, the \glspl{rmse} of the smoothing optimization algorithm are almost the same as when there was no gyroscope bias present. However, the filtering results are worse. This is because the bias needs some time to be properly estimated. This is illustrated in \Figureref{fig:calibration-filteringConv} where the gyroscope bias estimates and their uncertainties are shown for the filtering algorithm.

\begin{table}
\caption{Mean \gls{rmse} of the orientation estimates from $100$ Monte Carlo simulations in the presence of a gyroscope bias that is being estimated.}
\label{tab:calibration-rmsSim}
\begin{center}
\small
\begin{tabular}{lccc}
\toprule
\Gls{rmse} & Roll [$^\circ$]& Pitch [$^\circ$] & Heading [$^\circ$] \\
\midrule
Smoothing optimization & 0.39 & 0.39 & 2.29 \\
EKF orientation deviation & 0.46 & 0.46 & 4.20 \\
\bottomrule
\end{tabular}
\normalsize
\end{center}
\end{table}

A major difference between the \gls{map} and the \gls{ml} approaches, is that the \gls{map} takes into account a prior on the gyroscope bias. We analyze the effect of this prior using $500$ Monte Carlo simulations, simulating the gyroscope bias to be $\delta_\omega = \begin{pmatrix} 0.05 & 0.01 & -0.04 \end{pmatrix}^\Transp~\radianpersecond$. We study the estimated gyroscope biases using \gls{ml} estimation, and using \gls{map} estimation by including the gyroscope bias as an unknown in the smoothing algorithm. The smoothing algorithm assumes two different priors on the gyroscope bias $\delta_\omega \sim \mathcal{N}(0, \sigma^2_{\delta_\omega} \, \mathcal{I}_3)$. In the first case, the prior on the gyroscope bias can well describe the data ($\sigma_{\delta_\omega} = 0.05$). In the other case, the prior is too tight ($\sigma_{\delta_\omega} = 1 \cdot 10^{-3}$). The mean and standard deviations for the gyroscope bias estimates are summarized in \Tableref{tab:calibration-mcResults}. As can be seen, when the prior is chosen appropriately, the \gls{ml} and \gls{map} estimates are comparable. If the prior is too tight, the \gls{map} estimates can be seen to be biased towards zero. 
\begin{table}[h]
\caption{Mean and standard deviation of the gyroscope estimates over 500 Monte Carlo simulations with $\left( 0.05 \, \, \, 0.01 \, \, -0.04 \right)^\Transp~\radianpersecond$. Considered are the cases of \gls{ml} estimation and \gls{map} estimation by including the gyroscope bias as an unknown in a smoothing algorithm with a prior on the gyroscope bias of $\delta_\omega \sim \mathcal{N}(0, \sigma^2_{\delta_\omega} \mathcal{I}_3)$.}
\label{tab:calibration-mcResults}
\begin{center}
\small
\begin{tabularx}{1 \columnwidth}{l  *{3}{M} *{5}{M}}
\toprule
\Gls{rmse} & \multicolumn{3}{c}{Mean $\hat{\delta}_\omega$ ($\cdot 10^2$)} & \multicolumn{5}{c}{Standard deviation $\hat{\delta}_\omega$ ($\cdot 10^4$)} \\
\midrule
& $x$ & $y$ & $z$ & & $x$ & $y$ & $z$ &\\
\cmidrule(lr){2-4} 
\cmidrule(lr){6-8}
ML & 5.0 & 1.0 & -4.0 & \hspace{1pt} & 5.1 & 5.3 & 6.4 & \hspace{1pt}\\
MAP $\sigma_{\delta_\omega} = 0.05$ & 5.0 & 1.0 & -4.0 & & 5.1 & 5.3 & 6.4 & \\
MAP $\sigma_{\delta_\omega} = 1 \cdot 10^{-3}$ & 3.9 & 0.8 & -2.8 & & 4.1 & 4.0 & 4.7 & \\
\bottomrule
\end{tabularx}
\normalsize
\end{center}
\end{table}

\section{Identifiability}
\label{sec:calibration-identifiability}
Parameters are said to be \emph{identifiable} if it is possible to determine a unique parameter value from the data and if this value is equal to the true value~\citep{ljung:1999}. The concept of identifiability is closely related to the concept of \emph{observability} which is concerned with the question of if the time-varying states can be determined from the available data~\citep{kailath:1980}. The states discussed in \Chapterref{cha:orientationEstimation} are typically observable. Identifiability, however, becomes of concern when estimating calibration parameters. Specifically, in many applications, certain parameters are not identifiable when the sensor is completely stationary and sufficient excitation in terms of change in position and orientation is needed to make the parameters identifiable. This is illustrated in \Exampleref{ex:calibration-ident} for the case of identifiability of the gyroscope bias.

\begin{myexample}{Identifiability of the gyroscope bias}%
\label{ex:calibration-ident}%
We consider the example of orientation estimation using only inertial measurements in the presence of a gyroscope bias. We simulate data as described in \Sectionref{sec:oriEst-orientationEstimation}. The filtering estimates of the gyroscope bias and their uncertainties from an \gls{ekf} with orientation deviation states are shown in \Figureref{fig:calibration-identifiability}. Using only inertial measurements, the gyroscope bias of the sensor's $z$-axis is not identifiable when the sensor is placed horizontally. However, when the sensor is rotated, the accelerometer provides information that aids the estimation and the bias can be seen to converge. Note the difference with \Figureref{fig:calibration-filteringConv}, where only the first 100 samples were displayed and the bias estimates in the $z$-axis converged significantly faster due to the inclusion of magnetometer measurements. 

\begin{figure}
	\centering
    	\includegraphics[scale = 1]{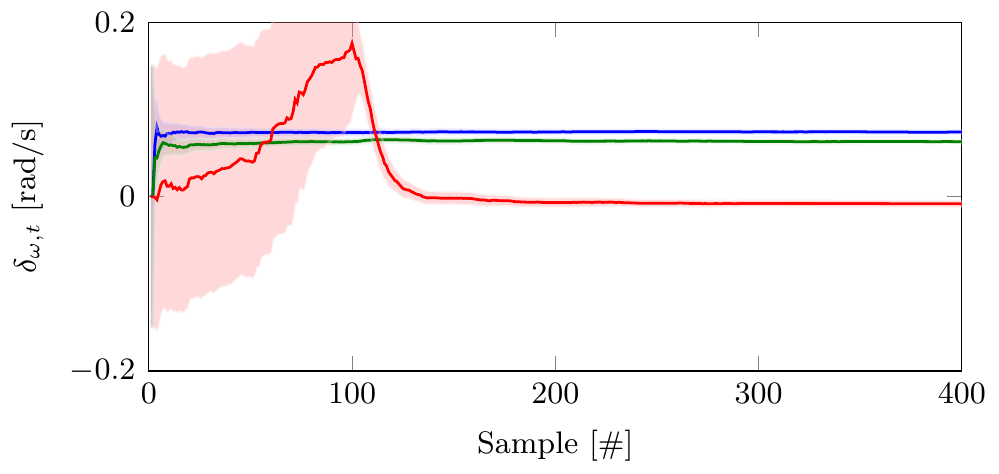}
    	\caption{Filtering estimates of the gyroscope bias and their $3 \sigma$ confidence intervals with the $x$-, $y$- and $z$-components in blue, green and red, respectively. The estimates are obtained using an \gls{ekf} and inertial (but no magnetometer) measurements. As can be seen, the estimates of the gyroscope bias in the $x$- and $y$-axes converge quickly while the estimates of the gyroscope bias in the $z$-axis only start to converge after 100 samples, when the sensor starts to rotate around the $x$-axis.}
	\label{fig:calibration-identifiability}
\end{figure}
\end{myexample}

\chapter{Sensor Fusion Involving Inertial Sensors}
\label{cha:applications} 
In this tutorial, we have presented a number of algorithms and modeling possibilities for position and orientation estimation using inertial sensors. The algorithms derived in \Chapterref{cha:orientationEstimation} are based on the simple state space models~\eqref{eq:models-ssPose} and~\eqref{eq:models-ssOri}. These models assume the availability of inertial measurements supplemented with either magnetometer or position measurements. In this chapter, we will illustrate how the same or similar algorithms can be used when additional and/or different supplementary sensors are available and discuss a number of possible extensions. The models used in this chapter will be more complex, but the information provided by the inertial sensors will remain one of the basic building blocks. We are not aiming for a full coverage, since that would require too much space. Instead we provide a number of case studies to illustrate what can be done when it comes to including the developments from Chapters~\ref{cha:introduction}--\ref{cha:calibration} into a more complex setting. The case studies are the modeling of a non-Gaussian noise distribution to make use of radio-based time of arrival measurements (\Sectionref{sec:appl-uwb}), the fusion of information from inertial sensors with cameras (\Sectionref{sec:appl-vision}), and motion capture based on information from multiple inertial sensors in combination with information about the human motion (\Sectionref{sec:appl-motionCapture} and \Sectionref{sec:appl-actRecogn}).

\section{Non-Gaussian noise distributions: time of arrival measurements}
\label{sec:appl-uwb}
In the algorithms presented in \Chapterref{cha:orientationEstimation}, we assumed that the measurement noise of the sensors is properly described using a Gaussian distribution. This was shown to be a fair assumption for the inertial measurement noise based on experimental data in \Exampleref{ex:sensors-inertialMeasurements}. In this section we will illustrate the possibility of modeling non-Gaussian measurement noise using a practical example. We will also describe possible adaptations of the algorithms presented in \Chapterref{cha:orientationEstimation} to incorporate non-Gaussian noise models. 

\begin{figure}
	\centering
		\includegraphics[scale = 1]{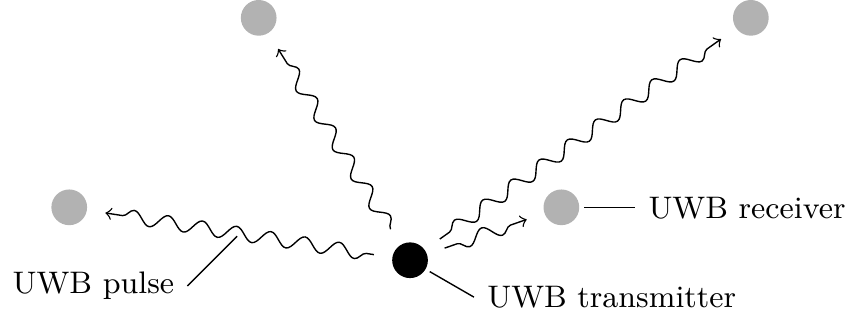}
  \caption{A \gls{uwb} setup consisting of a number of
    		stationary receivers obtaining \acrlong{toa} measurements of
    		signal pulses originating from a mobile transmitter.}
  \label{fig:appl-uwb-uwbSetup}
\end{figure}

In the state space model~\eqref{eq:models-ssPose}, we assumed the presence of position measurements with additive Gaussian noise. Many sensors, however, do not directly measure the position, see \eg \cite{gustafssonG:2005}. An example of a type of measurements that can be obtained are \emph{\gls{toa}} measurements from a \gls{uwb} system. These \gls{toa} measurements provide information about when a radio pulse sent out by a mobile transmitter arrives at a stationary receiver. This is graphically illustrated in \Figureref{fig:appl-uwb-uwbSetup}. These types of measurements can be modeled as
\begin{align}
  \label{eq:appl-uwb-measModel}
  y_{\text{u},mk} = \tau_{k}
    + \tfrac{1}{c} \| r_m - t_{k} \|_2 
    + \Delta\tau_m
    +   e_{\text{u},mk}.
\end{align}
Here, $c$ denotes the speed of light, $t_{k}$
is the position of the transmitter at the time of transmitting the $k^\text{th}$ pulse,
$r_m$ is the position of the $m^\text{th}$ receiver and $\Delta\tau_m$ is its clock-offset. The time when pulse $k$ is transmitted is denoted by $\tau_{k}$. Finally, $e_{\text{u},mk}$ is measurement noise. 

Using time of arrival measurements from a number of receivers it is possible to compute the position of the transmitter at the time when the pulse was sent. In many applications, the mobile transmitters have a much less accurate clock than the stationary receivers and $\tau_{k}$ is therefore assumed to be an unknown variable to be estimated. For a setup with $M$ stationary transmitters, the position of the transmitter when transmitting the pulse $k$ and the time when the pulse was transmitted can be estimated using multilateration~\citep{chanH:1994, geziciTGKMPS:2005,sayedTK:2005, sahinogluGG:2008}. 

\begin{figure}
	\centering
	\includegraphics[scale = 1]{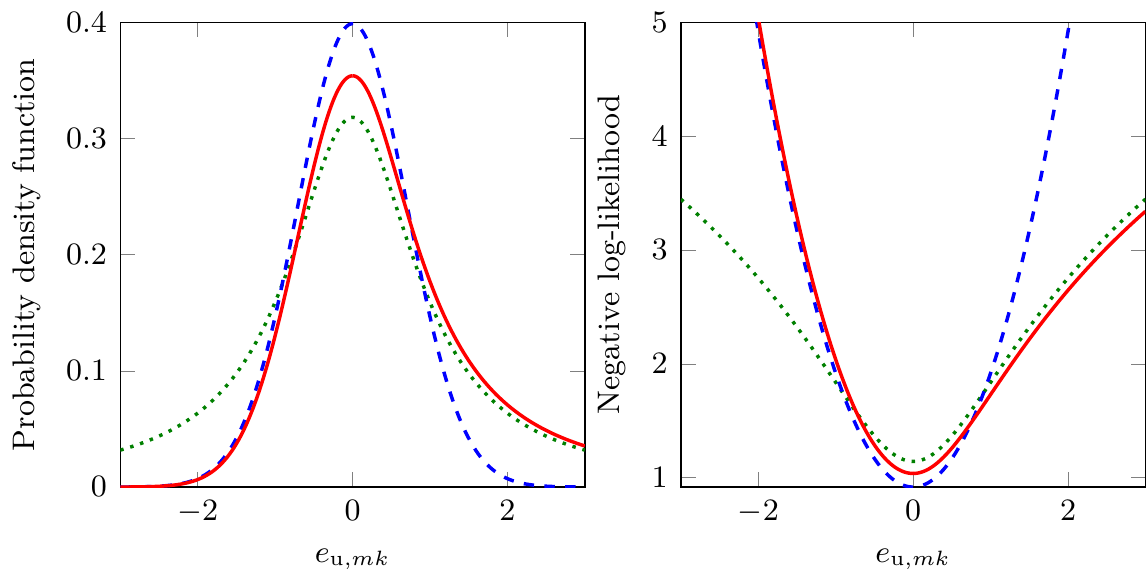}
		\caption{Probability density function (left) and negative log-likelihood (right) of a $\mathcal{N}(0,1)$ distribution (blue, dashed), a $\text{Cauchy}(0,1)$ distribution (green, dotted) and the asymmetric distribution~\eqref{eq:appl-uwb-noiseMeasModel} assuming $\sigma = \gamma = 1$ (red).}
	\label{fig:appl-uwb-loglikelihood}
\end{figure}

Due to \acrlong{nlos} conditions and/or multipath we expect a small number of measurements to be delayed. In~\cite{kokHS:2015}, we present an approach that models the measurement noise $e_{\text{u},mk}$ using a tailored asymmetric heavy-tailed distribution. Here, a heavy-tailed Cauchy distribution allows for measurement \textit{delays}, while a Gaussian distribution excludes the physically unreasonable possibility of pulses traveling faster than the speed of light as
\begin{subnumcases}{\label{eq:appl-uwb-noiseMeasModel} e_{\text{u},mk} \sim}
\left( 2 - \alpha \right) \mathcal{N}(0,\sigma^2)  & for $e_{\text{u},mk} < 0$, \\
\alpha \, \text{Cauchy}(0,\gamma) & for $e_{\text{u},mk} \geq 0$. 
\end{subnumcases}
The presence of the constants $\alpha$ and $2 - \alpha$ is motivated by the fact that the proposed asymmetric probability density function needs to integrate to one. The parameter $\alpha$ is defined as a function of $\sigma$ and $\gamma$. The proposed asymmetric probability density function and its corresponding negative log-likelihood, given by
\begin{subnumcases}{\label{eq:appl-uwb-negLogLik} - \log p \left( e_{\text{u},mk} \right)=}
\mathcal{L}_G & for $e_{\text{u},mk} < 0$,\\
\mathcal{L}_C & for $e_{\text{u},mk} \geq 0$, 
\end{subnumcases}%
\begin{align*}
\mathcal{L}_G &\triangleq \tfrac{e_{\text{u},mk}^2}{2 \sigma^2} + \tfrac{1}{2} \log \sigma^2 + \tfrac{1}{2} \log 2 \pi - \log \left( 2 - \alpha \right), \\
\mathcal{L}_C &\triangleq \log \left( 1 + \tfrac{e_{\text{u},mk}^2}{\gamma^2} \right) + \tfrac{1}{2} \log \gamma^2 + \log \pi - \log \alpha,
\end{align*}
are both depicted in \Figureref{fig:appl-uwb-loglikelihood} in red. For comparison, the Gaussian and Cauchy probability density functions are also depicted, in blue and green, respectively. Related studies have also modeled this presence of delayed measurements using skew-$t$ distributions \citep{nurminenAPG:2015,mullerPRS:2016}, Gaussian mixture models \citep{mullerWP:2014} or by introducing a number of unknown delays~\cite{hol:2011}.

The filtering and smoothing algorithms posed as optimization problems, see \Sectionref{sec:oriEst-smoothingOpt} and \Sectionref{sec:oriEst-filteringOpt}, can straightforwardly be adapted to include the asymmetric probability density function by using~\eqref{eq:appl-uwb-negLogLik} in the objective function instead of the weighted least squares term representing the measurement model, see \eg \eqref{eq:oriEst-posSmoothing}. Extended Kalman filters intrinsically assume that the process and measurement noise are Gaussian. For the extended Kalman filters implementations presented in \Sectionref{sec:oriEst-quat-ekf} and \Sectionref{sec:oriEst-oriError-ekf}, the extension to non-Gaussian noise is therefore less straightforward. Adaptations of the Kalman filter to non-Gaussian noise are, however, possible, see \eg \cite{rothAOG:2017} and the references therein for a discussion on Kalman filtering in the presence of student's $t$ distributed noise.

\section{Using more complex sensor information: inertial and vision}
\label{sec:appl-vision}
In the model~\eqref{eq:models-ssPose}, we assumed that a sensor directly measuring the position was available. In \Sectionref{sec:appl-uwb}, we instead considered the case of \gls{toa} measurements from a \gls{uwb} system. In this section, we will focus on pose estimation using a camera and an \gls{imu} that are rigidly attached to each other. Inertial sensors and camera images have successfully been combined for pose estimation, see \eg \cite{brysonJS:2009,luptonS:2012,jungT:2001,sjanicSG:2017,leuteneggerLBSF:2015,forsterCDS:2016,bleserS:2009,foxlinN:2003,nyqvistG:2013}. Application areas are for instance robotics, \acrlong{vr} and \acrlong{ar}.

To use camera images of the environment of the sensor to provide information about where the sensor is located in this environment, a number of \emph{features} is typically extracted from each image. These features are subsequently associated with a point in the environment. Advanced feature extraction and association algorithms are available, see \eg \cite{harrisS:1988,bayTG:2006,lowe:2004,fischlerB:1981}. 

\begin{figure}
  	\centering
    	\subfigure[Illustration of the pinhole camera model where an object $l$ results in an image $i$ in the image plane. The image plane lies a distance $f$ behind the pinhole.]{
	\includegraphics[scale = 1]{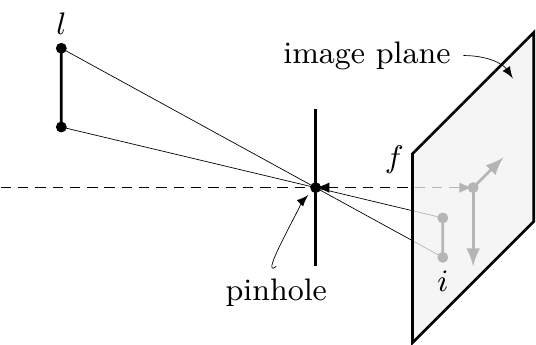}
	\label{fig:appl-vision-pinhole}}
    	\subfigure[2D visualization of the relation between the position of the point $l$ with respect to the pinhole and the position of its image $i$.]{
	\includegraphics[scale = 1]{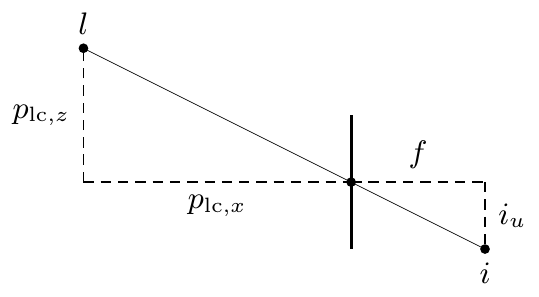}
	\label{fig:appl-vision-pointProj}}	
    	\caption[]{Illustrations of the relation between a point or object $l$ in the environment and its image $i$.}
    	\label{fig:appl-vision}
\end{figure}

To be able to use the features and their associated points in the real world for pose estimation, a model is required that relates points in the image to the environment. A simple model that can be used for this is a pin-hole model. Such a model is illustrated in \Figureref{fig:appl-vision-pinhole}, where an object $l$ results in an image $i$ in the image plane. The size of $i$ depends on the size of the object, the distance of the object to the pinhole and on the \emph{focal length}~$f$. The focal length is an intrinsic property of a camera and is typically assumed to be known. 

Let us now assume that the object $l$ is simply a point and denote its position with respect to the pinhole center by $p_{\text{lc}} = \begin{pmatrix} p_{\text{lc},x} & p_{\text{lc},y} & p_{\text{lc},z} \end{pmatrix}^\Transp$. We denote the position of the object $l$ in the image frame by $i = \begin{pmatrix} i_u & i_v \end{pmatrix}$. A two-dimensional illustration of this can be found in \Figureref{fig:appl-vision-pointProj}. The position $i$ can be expressed in terms of $p_{\text{lc}}$ and the focal length $f$ as
\begin{align}
\label{eq:appl-imPoints}
\begin{pmatrix} i_{u} \\ i_{v} \end{pmatrix} = f
\begin{pmatrix} \tfrac{p_{\text{lc},z}^\text{c}}{p_{\text{lc},x}^\text{c}} & \tfrac{p_{\text{lc},z}^\text{c}}{p_{\text{lc},y}^\text{c}} \end{pmatrix}^\Transp.
\end{align}
Here, the superscript $c$ indicates that the position $p_{\text{lc}}$ is expressed in the \emph{camera frame} $c$. The origin of the camera frame is located at the pinhole and its axes are aligned with the image plane. It is possible to express the position $p_{\text{lc}}$ in terms of the position of the camera and the position of the object $l$ as
\begin{align}
\label{eq:appl-imNav}
p_{\text{lc},t}^\text{c} &= R^\text{cn}_t \left( p^\text{n}_\text{l} - p^\text{n}_{\text{c},t} \right).
\end{align}
Here, $p^\text{n}_{\text{c},t}$ denotes the position of the camera at time $t$ expressed in the navigation frame. The rotation from the navigation frame $n$ to the camera frame $c$ at time $t$ is denoted by $R^\text{cn}_t$. Finally, the position of the point $l$ in the navigation frame is denoted by $p^\text{n}_\text{l}$. Note that the objects in the environment are assumed to be stationary, \ie $p^\text{n}_\text{l}$ does not change over time. Assuming that the position of the object $p^\text{n}_\text{l}$ is known, using~\eqref{eq:appl-imPoints} and~\eqref{eq:appl-imNav}, the image $i$ provides information about the position $p^\text{n}_{\text{c},t}$ and the orientation $R^\text{cn}_t$ of the camera.

To be able to combine the information from the camera images with inertial measurements, it is possible to express the position $p_{\text{c},t}$ and the orientation $R^\text{cn}_t$ of the camera in terms of the position $p^\text{n}_{t}$ and the orientation $R^\text{bn}_{t}$ of the body frame as
\begin{subequations}
\label{eq:appl-vision-cambody}
\begin{align}
p^\text{n}_{\text{c},t} &= p^\text{n}_{t} + d_\text{cb}^\text{n}, \\
R^\text{cn}_{t} &= R^\text{cb} R^\text{bn}_{t}.
\end{align}
\end{subequations}
Here, the distance between the body frame and the camera frame is denoted by $d_\text{cb}$ and the rotation between these two frames is denoted by $R^\text{cb}$. Both are assumed to be constant and can be determined using dedicated calibration algorithms~\cite{holSG:2010b,loboD:2007,kellyS:2011,mirzaeiR:2008,rehderS:2017}. The relation~\eqref{eq:appl-imNav} can therefore be expressed in terms of $R^\text{bn}_{t}$ and $p^\text{n}_{t}$ as
\begin{align}
\label{eq:appl-vision-imupos}
p_{\text{lc},t}^\text{c} &= R^\text{cb} R^\text{bn}_t \left( p^\text{n}_\text{l} - p^\text{n}_{t} \right) - d_\text{cb}^\text{c}.
\end{align}

Using~\eqref{eq:appl-imPoints}, the measurement model~\eqref{eq:models-ssPose-meas} can now be replaced with measurements $y_{\text{c},t}$ from the camera as
\begin{align}
y_{\text{c},t} &= \begin{pmatrix} i_{u,t} \\ i_{v,t} \end{pmatrix} = f
\begin{pmatrix} \tfrac{p_{\text{lc},z,t}^\text{c}}{p_{\text{lc},x,t}^\text{c}} & \tfrac{p_{\text{lc},z,t}^\text{c}}{p_{\text{lc},y,t}^\text{c}} \end{pmatrix}^\Transp + e_{\text{c},t}, 
\end{align}
where $e_{\text{c},t}$ is measurement noise and $p_{\text{lc},t}^\text{c}$ is given by~\eqref{eq:appl-vision-imupos}. Including this measurement model in~\eqref{eq:models-ssPose} allows us to combine inertial sensors with camera images to estimate the position and orientation of the sensor. 

Note that at each time instance there can be multiple measurements from the camera image. One version of this adapted state space model estimates the position and orientation of the sensor assuming that the position of the objects $p^\text{n}_\text{l}$ in the environment is known and constant. Alternatively, the positions $p^\text{n}_\text{l}$ for objects $l = 1, \hdots, L$ are considered to be unknown and part of the state vector. The latter is called \emph{\gls{slam}}, where a map of the environment is built from measurements of a mobile sensor whose position is unknown and to be estimated. There exist many algorithms for doing \gls{slam}, both using \gls{ekf} and optimization-based approaches, see \eg the tutorial papers~\cite{durrantWhyteB:2006,baileyD:2006} or \cite{cadenaCCLSNRL:2016} for a more recent overview.

\section{Including non-sensory information: inertial motion capture}
\label{sec:appl-motionCapture}
In the models~\eqref{eq:models-ssPose} and~\eqref{eq:models-ssOri}, we have only made use of information obtained through sensor data. In some situations, however, additional non-sensory information is available. An example of this is inertial motion capture. 

\begin{figure}
  \centering
  	\includegraphics[width = 0.3\columnwidth]{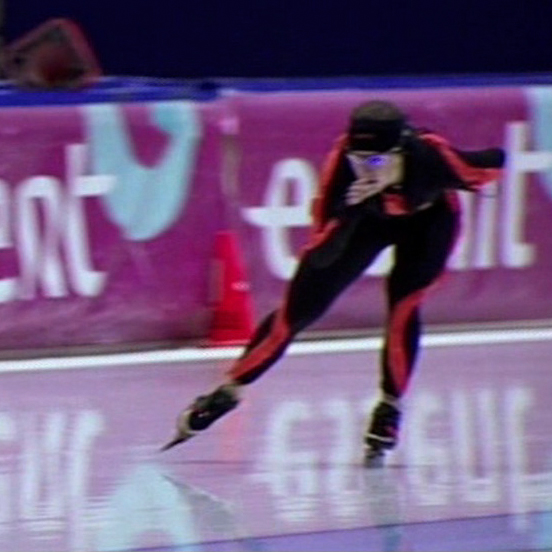}
	\includegraphics[width = 0.3\columnwidth]{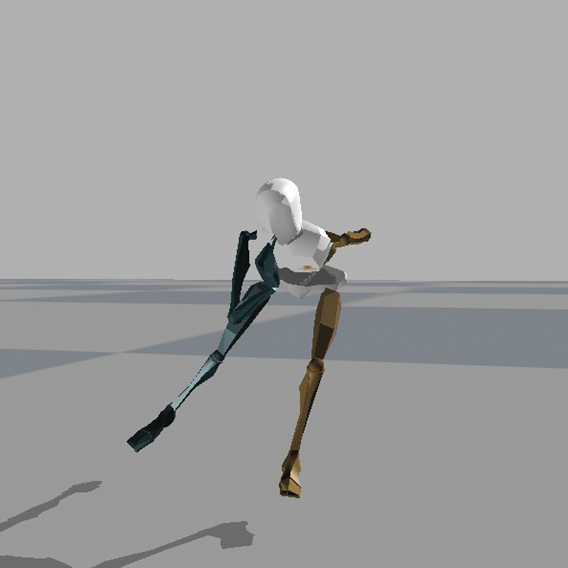}
	\label{fig:kappa-intro-applications-wust} 
  \caption{Left: Olympic and world champion speed skating Ireen W\"ust wearing sensors on her body. Right: graphical representation of the estimated orientation and position of her body segments.}
  \label{fig:appl-wust}
\end{figure}

In inertial sensor human body motion capture, \glspl{imu} placed on different body segments are used to estimate the body's pose. This information can be useful for rehabilitation or for improving sports performance. An example can be seen in \Figureref{fig:appl-wust} where Olympic and world champion speed skating Ireen W\"ust wears sensors on her body that give information about her posture while ice skating. One can imagine that she can use this information to analyze which angles her knees and hips should have to skate as fast as possible and if her posture changes when she gets tired. It is also possible to use the information about how a person moves for motion capture in movies and games. This was illustrated in \Figureref{fig:intro-motionCaptureApplications-ted}, where the actor Seth MacFarlane wears sensors on his body that measure his movements to animate the bear Ted. Inertial motion capture is successfully used in a large number of applications, see \eg \cite{roetenbergLS:2013,kangJPK:2011,yunB:2006}. 

In~\cite{kokHS:2014}, we present an algorithm for inertial motion capture which solves a smoothing optimization problem similar to~\eqref{eq:oriEst-posSmoothing}. A significant part of the model used in the algorithm in~\cite{kokHS:2014} is the state space model for pose estimation presented in~\eqref{eq:models-ssPose} and used in \Sectionref{sec:oriEst-poseEstimation}. This model is used for every sensor that is placed on the human body. The following adaptations are made to the model: 
\begin{description}[style=unboxed]
\item[Placement of the sensors on the body segments.]
The position and orientation of each sensor can be expressed in terms of its position and orientation on the corresponding body segment. Ideally, the sensor would be rigidly attached to the segment. However, it is physically impossible to place the sensor directly on the bone. Since it needs to be placed on the soft tissue instead, the sensor will inevitably move slightly with respect to the bone. We therefore model the position and orientation of each sensor on its corresponding body segment as 
approximately constant.
\item[Joints between the body segments.] A number of equality constraints enforce the body segments to be connected at the joint locations at all times. Note that equality constraints can straightforwardly be included in optimization problems \citep{boydV:2004,nocedalW:2006}.
\item[Exclusion of magnetometer measurements.] The magnetic field in indoor environments is not constant, see \eg \cite{ligorioS:2016}. This is specifically of concern in motion capture applications, since the magnetic field measured at the different sensor locations is typically different~\citep{luingeVB:2007,cooper:2009,favreJAA:2008,seelRS:2014}. Because of this, we do not include magnetometer measurements in the model. The inclusion of the equality constraints enforcing the body segments to be connected allows us to do so, since incorporating these constraints, the sensor's \textit{relative} position and orientation become observable as long as the subject is not standing completely still~\citep{hol:2011}.
\end{description}
As before, we denote the time-varying states by $x_t$ and the constant parameters by $\theta$. However, to highlight the different parts of the model, we split the states $x_t$ in states $x^{\text{S}_i}_t$ pertaining to the sensor $\text{S}_i$ and states $x^{\text{B}_i}_t$ pertaining to the body segment $\text{B}_i$, for $i = 1, \hdots , N_S$. Here, $N_S$ is the number of sensors attached to the body and it is assumed to be equal to the number of body segments that are modeled. Using this notation, the optimization problem that is solved in~\cite{kokHS:2014} can be summarized as
\begin{align}\label{eq:appl-optProblem-motionCapture}
\min_{x_{1:N},\theta} & \quad 
\underbrace{- \sum_{t = 2}^{N_T} \sum_{i = 1}^{N_S} \log p(x^{\text{S}_i}_t \mid x^{\text{S}_i}_{t-1})}_{\text{Dynamics of the state }x^{\text{S}_i}_t}
\underbrace{- \sum_{t = 1}^{N_T} \sum_{i = 1}^{N_S} \log p(x^{\text{B}_i}_t \mid x^{\text{S}_i}_{t})}_{\begin{subarray}{c}\text{Placement of sensor $\text{S}_i$}\\
    \text{on body segment $\text{B}_i$}\end{subarray}}
     \nonumber \\
& \hspace{20mm}\underbrace{- \sum_{i = 1}^{N_S} \log p(x^{\text{S}_i}_1) - \sum_{i = 1}^{N_S}\log p(\theta^{\text{S}_i})}_{\text{Prior}}
\nonumber \\
\st& \quad c(x_{1:N}) = 0,
\end{align}
where the number of time steps is denoted $N_T$ and $c(x_{1:N})$ denote the equality constraints from assuming that the body segments are connected at the joints. Note the similarities and differences between the structure of~\eqref{eq:appl-optProblem-motionCapture} as compared to~\eqref{eq:models-smoothingProbs} or~\eqref{eq:oriEst-posSmoothing}. The problem~\eqref{eq:appl-optProblem-motionCapture} is formulated as a smoothing algorithm. In \cite{miezalTB:2016}, a similar method is presented in which a sliding window approach is used to allow for online estimation. In \gls{ekf} formulations it is not straightforward to include equality constraints. However, as discussed in \cite{miezalTB:2016}, it is possible to include the equality constraints as a measurement update in the \gls{ekf} instead. 

Optionally, additional knowledge about the human body can be included in~\eqref{eq:appl-optProblem-motionCapture}. For instance, in \cite{kokHS:2014}, we include the knowledge that for some joints, it is known that their rotation is (mainly) limited to one or two axes. An example of this is the knee which is a hinge joint, although it can in practice flex a little around the other axes too. The optimization problem for that case would include a term that minimizes the rotation around certain axes. In \cite{taetzBM:2016}, a solution to the inertial motion capture problem is presented that includes additional biomechanical information. For instance, the shape of the body segments is approximated by a capsule and the sensors are assumed to be placed on the surface of the capsule. In~\cite{marcardRBP:2017}, the focus is instead on estimation of human motion using a much smaller amount of sensors. Because of the smaller number of sensors, the information about the human body and its motions is more crucial in their problem formulation. 

\section{Beyond pose estimation: activity recognition}
\label{sec:appl-actRecogn}
In this tutorial we have focused on position and orientation estimation using inertial sensors. A separate but related field of research is that of activity recognition using inertial sensors, where the aim is to determine the type of motion that is performed by a person wearing or carrying a device containing inertial sensors, see \cite{bullingBS:2014,avciBMMH:2010} for an overview. Since these sensors have become smaller, cheaper and more publicly available over the years, the amount of data that can be used for activity recognition has increased dramatically. For instance, inertial measurements collected from a smartphone can be used to classify daily activities. It is also possible to classify these activities using measurements from fitness and activity trackers, see \eg \cite{fitbit:2017,garmin:2017,polar:2017}. Standard activities which can be classified are for instance walking, running and sleeping, but there is a large number of other activities that one can think of as well. 

A simple example of activity recognition that is relevant for the motion capture application discussed in \Sectionref{sec:appl-motionCapture}, is footstep detection. As mentioned in \Sectionref{sec:appl-motionCapture}, the \emph{relative} position and orientation of the body can be determined using inertial sensors and a biomechanical model about the body segments being connected. However, the \emph{absolute} position can not be determined from this information. Because of that, it is not possible to determine if the subject is standing on the floor or floating in the air by only using the model~\eqref{eq:appl-optProblem-motionCapture}. Including the knowledge that during standing, walking and running, the feet are stationary on the floor during certain periods of time solves this problem \cite{foxlin:2005,colomarNH:2012}. It is possible to detect these stationary periods for example by identifying when both the angular velocity and the sensor's acceleration are approximately zero, see \cite{skogHNR:2010} for a comparison between different detection methods. The knowledge that the sensor is stationary can subsequently be included in~\eqref{eq:appl-optProblem-motionCapture} by adding a term to the cost function modeling the velocity of the sensor as being almost zero during these time periods. These are so-called zero velocity updates (ZUPTs)~\cite{foxlin:2005,colomarNH:2012}. 

Footstep detection is a very simple example where pose estimation can benefit from activity recognition. Another example is presented in~\cite{grzonkaDKB:2010} where maps of indoor environments are build using inertial measurements from sensors placed on the human body, very similar to the approach presented in \Sectionref{sec:appl-motionCapture}. In addition to the model presented in \Sectionref{sec:appl-motionCapture} and footstep detection, motions are detected of the subject opening and closing doors. Combining this information opens up for possibilities of performing \gls{slam}. In conclusion, activity recognition can serve as a source of additional information that can be included in our models for pose estimation. Conversely, in~\cite{hardeggerRCT:2016,reissHBS:2010} it is argued that activity recognition can also benefit from combination with pose estimation. 

\chapter{Concluding Remarks}
\label{cha:conclusions} 
The goal of this tutorial was not to give a complete overview of all algorithms that can be used for position and orientation estimation. Instead, our aim was to give a pedagogical introduction to the topic of position and orientation estimation using inertial sensors, allowing newcomers to this problem to get up to speed as fast as possible by reading one paper. By integrating the inertial sensor measurements (so-called dead-reckoning), it is possible to obtain information about the position and orientation of the sensor. However, errors in the measurements will accumulate and the estimates will drift. Because of this, to obtain accurate position and orientation estimates using inertial measurements, it is necessary to use additional sensors and additional models. In this tutorial, we have considered two separate estimation problems. The first is orientation estimation using inertial and magnetometer measurements, assuming that the acceleration of the sensor is approximately zero. Including magnetometer measurements removes the drift in the heading direction (as illustrated in \Exampleref{ex:oriEst-noMagData}), while assuming that the acceleration is approximately zero removes the drift in the inclination. The second estimation problem that we have considered is pose estimation using inertial and position measurements. Using inertial measurements, the position and orientation estimates are coupled and the position measurements therefore provide information also about the orientation. 

A number of algorithms for position and orientation estimation have been introduced in \Chapterref{cha:orientationEstimation}. These include smoothing and filtering solved as an optimization problem, two different extended Kalman filter implementations and a version of the complementary filter. The filtering approaches use the data up to a certain time $t$ to estimate the position and orientation at this time $t$. Smoothing instead makes use of all data from time $t = 1, \hdots , N$. In general, using all data to obtain the estimates will naturally lead to better estimates. The filtering approaches can be seen to be quite uncertain about their estimates for the first samples and require some time to ``converge'' to accurate estimates. This is even more pronounced in the presence of calibration parameters as discussed in \Chapterref{cha:calibration}. Although smoothing algorithms give better estimates, practical applications might not allow for computing smoothing estimates because of computational limitations or real-time requirements. For the examples discussed in this paper, the optimization-based filtering algorithm and the \gls{ekf} with orientation deviation states perform very similarly. The \gls{ekf} with quaternion states, however, was able to handle wrong initial orientations less well as shown in \Tableref{tab:oriEst-rmsSimWrongInit}. Furthermore, it underestimated the uncertainty in the heading direction in the absence of magnetometer measurements, see also \Exampleref{ex:oriEst-noMagDataCovComp}. The complementary filter is a good alternative to these methods, but its estimation accuracy decreases when orientation estimates from the accelerometer and magnetometer measurements are of significantly better quality in the inclination than in the heading direction. In \Chapterref{cha:applications} we have discussed possible extensions to the simple models considered earlier in this tutorial. One of the benefits of the optimization-based approaches is that these extensions can straightforwardly be incorporated into the framework. 

Apart from the differences between the estimation algorithms discussed in this tutorial, it can also be concluded that the position and orientation estimation problems using inertial sensors are actually quite forgiving. Any of the algorithms introduced in this tutorial can give reasonably good estimates with fairly little effort. However, careful modeling is important since the quality of the estimates of the algorithms highly depends on the validity of the models. This was illustrated in \Exampleref{ex:oriEst-magDist} for the influence of magnetic material in the vicinity of the sensor on the quality of the orientation estimates. In recent years, inertial sensors have undergone major developments. The quality of their measurements has improved while their cost has decreased, leading to an increase in availability. Furthermore, available computational resources are steadily increasing. Because of these reasons, we believe that inertial sensors can be used for even more diverse applications in the future. 

\chapter{Notation and Acronyms}
\label{cha:notation}
\begin{center}
\printglossary[nonumberlist,style=mystyleAbbr]
\end{center}

\begin{center}
\printglossary[type=coordinateFrames,nonumberlist,style=mystyleCoord]
\end{center}

\begin{center}
\printglossary[type=notation,nonumberlist,style=mystyleNot]
\end{center}

\begin{center}
\printglossary[type=operators,nonumberlist,style=mystyleNot]
\end{center}

\chapter*{Acknowledgements}
This research was financially supported by the projects \emph{Probabilistic modeling of dynamical systems} (Contract number:
621-2013-5524), \emph{NewLEADS - New Directions in Learning Dynamical Systems} (Contract number: 621-2016-06079) and \emph{CADICS}, all funded by the Swedish Research Council, the project \emph{ASSEMBLE} (Contract number: RIT15-0012) funded by the Swedish Foundation for Strategic Research (SSF) and the EPSRC grant \emph{Autonomous behaviour and learning in an uncertain world} (Grant number: EP/J012300/1). High accuracy reference measurements are provided through the use of the Vicon real-time tracking system courtesy of the UAS Technologies Lab, Artificial Intelligence and Integrated Computer Systems Division (AIICS) at the Department of Computer and Information Science (IDA), Link\"oping University, Sweden \url{http://www.ida.liu.se/divisions/aiics/aiicssite/index.en.shtml}. The authors would also like to thank Fredrik Olsson, Michael Lorenz, \dr Oliver Woodford, \dr Gustaf Hendeby, \prof Fredrik Gustafsson and \prof Magnus Herberthson for comments and suggestions that greatly improved this tutorial. 

\appendix

\chapter{Orientation Parametrizations}
\label{app:rotation}
In \Sectionref{sec:app-rotation-quatAlg} of this appendix, we will summarize some important results on quaternion algebra that we make frequent use of throughout this tutorial. In \Sectionref{sec:app-rotation-conv}, we summarize some results on how to convert between the different orientation parametrizations. 

\section{Quaternion algebra}
\label{sec:app-rotation-quatAlg}
A quaternion is a $4$-dimensional vector $q$, 
\begin{align}
q = \begin{pmatrix} q_0 & q_v^\Transp \end{pmatrix}^\Transp = \begin{pmatrix} q_0 & q_1 & q_2 & q_3 \end{pmatrix}^\Transp.
\end{align}
A special case is the unit quaternion, for which $\| q\|_2 = 1$. We use unit quaternions as a parametrization of orientations. An example of a quaternion that is not a unit quaternion and that we frequently encounter in this tutorial is the quaternion representation of a vector. For a vector $v$, its quaternion representation is given by 
\begin{align}
\bar{v} = 
\begin{pmatrix} 
0 \\ v
\end{pmatrix}.
\end{align}
The rotation of a vector $v$ by a unit quaternion $q$ can be written as
\begin{align}
q \odot \bar{v} \odot q^\conj.
\end{align}
Here, the quaternion multiplication $\odot$ of two quaternions $p$ and $q$ is defined as
\begin{align}
p \odot q = \begin{pmatrix} p_0 q_0 - p_v \cdot q_v \\ p_0 q_v + q_0 p_v + p_v \times q_v \end{pmatrix}.
\end{align}
This can alternatively be defined in terms of the left and right multiplication matrices
\begin{align} 
p \odot q = p^\leftMult q = q^\rightMult p,
\end{align}
with 
\begin{align}
p^\leftMult &\triangleq \begin{pmatrix} p_0 & -p_v^\Transp \\ p_v & p_0 \mathcal{I}_3 + [p_v \times] \end{pmatrix}, \qquad q^\rightMult \triangleq \begin{pmatrix} q_0 & -q_v^\Transp \\ q_v & q_0 \mathcal{I}_3 - [q_v \times] \end{pmatrix},
\end{align}
where $[q_v \times]$ denotes the cross product matrix
\begin{align}
[q_v \times] = \begin{pmatrix} 0 & -q_3 & q_2 \\ q_3 & 0 & -q_1 \\ -q_2 & q_1 & 0 \end{pmatrix}.
\label{eq:app-matrixCross}
\end{align}
The quaternion conjugate is given by 
\begin{align}
q^\conj = \begin{pmatrix} q_0 \\ - q_v\end{pmatrix}.
\end{align}
Hence, the rotation of the vector $v$ is given by
\begin{align}
q \odot \bar{v} \odot q^\conj &= q^\leftMult \left( q^\conj \right)^\rightMult \bar{v} \nonumber \\
&= \begin{pmatrix} q_0 & -q_v^\Transp \\ q_v & q_0 \mathcal{I}_3 + [q_v \times] \end{pmatrix} \begin{pmatrix} q_0 & -q_v^\Transp \\ q_v & q_0 \mathcal{I}_3 - [q_v \times] \end{pmatrix}  \begin{pmatrix} 0 \\ v \end{pmatrix} \nonumber \\
&=\begin{pmatrix} 1 & 0_{1 \times 3} \\ 0_{3 \times 1} & q_v q_v^\Transp  + q_0^2 \mathcal{I}_3 + 2 q_0 [q_v \times]  + [q_v \times ]^2 \end{pmatrix}  
\begin{pmatrix} 0 \\ v \end{pmatrix}.
\label{eq:app-rotVectorq}
\end{align}

\section{Conversions between different parametrizations}
\label{sec:app-rotation-conv}
A quaternion $q$ can be converted into a rotation matrix $R$ as
\begin{align}
R &= q_v q_v^\Transp  + q_0^2 \mathcal{I}_3 + 2 q_0 [q_v \times]  + [q_v \times ]^2 \nonumber \\
&= \begin{pmatrix} 2 q_0^2 + 2 q_1^2 - 1 & 2 q_1 q_2 - 2 q_0 q_3 & 2 q_1 q_3 + 2 q_0 q_2 \\ 2 q_1 q_2 + 2 q_0 q_3 & 2 q_0^2 + 2 q_2^2 - 1 & 2 q_2 q_3 - 2 q_0 q_1 \\ 2 q_1 q_3 - 2 q_0 q_2 & 2 q_2 q_3 + 2 q_0 q_1 & 2 q_0^2 + 2 q_3^2 -1 \end{pmatrix}.
\label{eq:app-defRq}
\end{align}
Note the similarity with~\eqref{eq:app-rotVectorq}, where the rotation of the vector $v$ can equivalently be expressed as $Rv$. Conversely, a rotation matrix
\begin{align}
R = \begin{pmatrix}
R_{11} & R_{12} & R_{13} \\
R_{21} & R_{22} & R_{23} \\
R_{31} & R_{32} & R_{33} 
\end{pmatrix},
\end{align}
can be converted into a quaternion as
\begin{align}
q_0 = \tfrac{\sqrt{1 + \Tr{R}}}{2}, \qquad q_v = \tfrac{1}{4 q_0} \begin{pmatrix} 
R_{32} - R_{23} \\
        R_{13} - R_{31} \\
        R_{21} - R_{12}
\end{pmatrix}. 
\label{eq:app-quatToRmat}
\end{align}
Note that a practical implementation needs to take care of the fact that the conversion~\eqref{eq:app-quatToRmat} only leads to sensible results if $1 + \Tr{R} > 0$ and $q_0 \neq 0$. To resolve this issue, the conversion is typically performed in different ways depending on the trace of the matrix $R$ and its diagonal values, see \eg\cite{euclideanSpace}. 

A rotation vector $\oriError$ can be expressed in terms of a unit quaternion $q$ via the quaternion exponential as
\begin{align}
q = \expq \oriError = \begin{pmatrix} \cos \| \oriError \|_2 \\ \tfrac{\oriError}{\| \oriError \|_2} \sin \| \oriError \|_2 \end{pmatrix}.
\label{eq:app-quatExp}
\end{align}
Note that any practical implementation needs to take care of the fact that this equation is singular at $\oriError = 0$, in which case $\expq \oriError = \begin{pmatrix} 1 & 0 & 0 & 0 \end{pmatrix}^\Transp$. The inverse operation is executed by the quaternion logarithm, 
\begin{align}
\oriError = \logq q = \tfrac{\arccos q_0}{\sin \arccos q_0} q_v = \tfrac{\arccos q_0}{\| q_v\|_2} q_v.
\label{eq:app-quatLog}
\end{align}
Note that this equation is singular at $q_v = 0$. In this case, $\log q$ should return $0_{3 \times 1}$. 
 
The rotation vector $\oriError$ can also be converted into a rotation matrix as 
\begin{align}
R = \expR \oriError = \exp \left( [\oriError \times] \right), \qquad \oriError = \logR R = \begin{pmatrix} (\log R)_{32} \\ (\log R)_{13} \\ (\log R)_{21} \end{pmatrix},
\end{align}
where $\log R$ is the matrix logarithm and $\logR$ and $\expR$ are the mappings introduced in~\eqref{eq:models-expqR-map} and~\eqref{eq:models-logqR}.

A rotation in terms of Euler angles can be expressed as a rotation matrix $R$ as 
{\footnotesize{
\begin{align}
\label{eq:app-rotMatrix}
R 
&= \begin{pmatrix} 1 & 0 & 0 \\ 0 & \cos \phi & \sin \phi \\ 0 & -\sin \phi & \cos \phi \end{pmatrix}
\begin{pmatrix} \cos \theta & 0 & -\sin \theta \\ 0 & 1 & 0 \\ \sin \theta & 0 & \cos \theta \end{pmatrix}
\begin{pmatrix} \cos \psi & \sin \psi & 0 \\ -\sin \psi & \cos \psi & 0 \\ 0 & 0 & 1 \end{pmatrix} \\
&= \begin{pmatrix} \cos \theta \cos \psi & \cos \theta \sin \psi & -\sin \theta \\ 
\sin \phi \sin \theta \cos \psi - \cos \phi \sin \psi & \sin \phi \sin \theta \sin \psi + \cos \phi \cos \psi & \sin \phi \cos \theta \\ 
\cos \phi \sin \theta \cos \psi + \sin \phi \sin \psi & \cos \phi \sin \theta \sin \psi - \sin \phi \cos \psi & \cos \phi \cos \theta \end{pmatrix}.\nonumber
\end{align}}}%
The rotation matrix $R$ can be converted into Euler angles as
\begin{subequations}
\begin{align}
\psi &= \tan^{-1} \left( \tfrac{R_{12}}{R_{11}} \right) = \tan^{-1} \left( \tfrac{2 q_1 q_2 - 2 q_0 q_3}{2 q_0^2 + 2 q_1^2 - 1} \right), \\
\theta &= -\sin^{-1} \left( R_{13} \right) = - \sin^{-1} \left( 2 q_1 q_3 + 2 q_0 q_2 \right), \\
\phi &= \tan^{-1} \left( \tfrac{R_{23}}{R_{33}} \right) = \tan^{-1} \left( \tfrac{2 q_2 q_3 - 2 q_0 q_1}{2 q_0^2 + 2 q_3^2 - 1} \right).
\end{align}
\end{subequations}

Using the relations presented in this section, it is possible to convert the parametrizations discussed in \Sectionref{sec:models-paramOri} into each other.

\chapter{Pose Estimation}
\label{app:poseEst}
In this appendix, we will introduce the necessary components to extend Algorithms~\ref{alg:oriEst-smoothingOpt}--\ref{alg:oriEst-ekfOriError} to pose estimation algorithms using the state space model~\eqref{eq:models-ssPose}.
\section{Smoothing in an optimization framework}
In \Sectionref{sec:oriEst-poseEstimation}, we presented the smoothing optimization problem for pose estimation. To adapt \Algorithmref{alg:oriEst-smoothingOpt}, the derivatives~\eqref{eq:oriEst-derInit}--\eqref{eq:oriEst-derDynModelPrev} in combination with the following derivatives are needed
\begin{subequations}
\label{eq:oriEst-derPose}
\begin{align}
\tfrac{\diff e_{\text{p,i}} }{\diff p_1^\text{n}}&= \mathcal{I}_3, \, &
\tfrac{\diff e_{\text{v,i}} }{\diff v_1^\text{n}}&= \mathcal{I}_3, \, &
\tfrac{\diff e_{\text{p},t} }{\diff p_t^\text{n}}&= -\mathcal{I}_3,
\label{eq:app-poseEst-smoothingDerPosVelInitPos} \\
\tfrac{\diff e_{\text{p,a},t} }{\diff p_{t+1}^\text{n}}&= \tfrac{2}{T^2}\mathcal{I}_3,  \, &
\tfrac{\diff e_{\text{p,a},t} }{\diff p_t^\text{n}}&= - \tfrac{2}{T^2} \mathcal{I}_3, && \nonumber \\
\tfrac{\diff e_{\text{p,a},t} }{\diff v_t^\text{n}}&= - \tfrac{1}{T} \mathcal{I}_3, \, &
\tfrac{\diff e_{\text{p,a},t} }{\diff \oriError_t^\text{n}}&= [ \tilde{R}^\text{nb}_t y_{\text{a},t} \times], && \label{eq:app-poseEst-smoothingDerPos} \\
\tfrac{\diff e_{\text{v,a},t} }{\diff v_{t+1}^\text{n}}&= \tfrac{1}{T}\mathcal{I}_3, \, &
\tfrac{\diff e_{\text{v,a},t} }{\diff v_t^\text{n}}&= - \tfrac{1}{T} \mathcal{I}_3, \, &
\tfrac{\diff e_{\text{v,a},t} }{\diff \oriError_t^\text{n}}&= [ \tilde{R}^\text{nb}_t y_{\text{a},t} \times]. \label{eq:app-poseEst-smoothingDerVel} 
\end{align}
\end{subequations}

\section{Filtering in an optimization framework}
To obtain position, velocity and orientation estimates in a filtering framework, the optimization problem~\eqref{eq:oriEst-filtOptNLS} is adapted to
\begin{align}
\hat{x}_t &= \argmin_{x_{t}} - \log p(x_{t} \mid y_{1:t}) \nonumber \\
&= \argmin_{x_{t}} \| e_{\text{f},t} \|_{P_{t \mid t-1}^{-1}} + \| e_{\text{p},t} \|_{\Sigma_\text{p}^{-1}},
\end{align}
where $\| e_{\text{p},t} \|_{\Sigma_\text{p}^{-1}}$ is the position measurement model also used in the smoothing optimization problem presented in \Sectionref{sec:oriEst-poseEstimation}. Furthermore, $e_{\text{f},t}$ is extended from~\eqref{eq:oriEst-costDerFiltMarg} to be 
\begin{align}
e_{\text{f},t} = \begin{pmatrix} p_{t}^\text{n} - p_{t-1}^\text{n} - T v_{t-1}^\text{n} - \tfrac{T^2}{2} \left( R_{t-1}^\text{nb} y_{\text{a},t-1} + g^\text{n} \right) \\
v_{t}^\text{n} - v_{t-1}^\text{n} - T \left( R_{t-1}^\text{nb} y_{\text{a},t-1} + g^\text{n} \right) \\
\oriError^\text{n}_{t}- 2 \logq \left(\hat{q}_{t-1}^\text{nb} \odot \expq (\tfrac{T}{2} y_{\omega,t-1} ) \odot  \tilde{q}_{t}^\text{bn} \right) \end{pmatrix}, 
\end{align}
where $\tfrac{\diff e_{\text{f},t}}{\diff x_t} = \mathcal{I}_9$ and $x_t = \begin{pmatrix} p_t^\Transp & v_t^\Transp & (\oriError^\text{n}_{t})^\Transp \end{pmatrix}^\Transp$.

The covariance matrix $P_{t+1 \mid t}$ is given by $P_{t+1 \mid t} = F_{t} P_{t \mid t} F_{t}^\Transp + G_{t} Q G_{t}^\Transp$ with
\begin{subequations}
\label{eq:app-estPose-timeUpdateFiltOpt}
\begin{align}
F_t &= \begin{pmatrix} \mathcal{I}_3 & T \mathcal{I}_3 & - \tfrac{T^2}{2}  [\hat{R}_t^\text{nb} y_{\text{a},t} \times] \\
0 & \mathcal{I}_3 & - T [\hat{R}_t^\text{nb} y_{\text{a},t} \times] \\
0 & 0 & \mathcal{I}_3 \end{pmatrix}
, \qquad
G_t = \begin{pmatrix} \mathcal{I}_6 & 0 \\ 0 & T \tilde{R}_{t+1}^{(0)} \end{pmatrix}, \label{eq:app-estPose-timeUpdateFiltOptFG} \\
Q &= \begin{pmatrix} \Sigma_\text{a} & 0 & 0 \\ 0 & \Sigma_\text{a} & 0 \\ 0 & 0 & \Sigma_\omega \end{pmatrix}.
\end{align}
\end{subequations}
Similar to the update of the linearization point in~\eqref{eq:oriEst-filteringOpt-updateLinPointIt0}, we also update the estimates of the position and velocity before starting the optimization algorithm, such that $e_{\text{f},t}$ is equal to zero for iteration $k = 0$.

\section{Extended Kalman filter with quaternion states}
Following the notation in \Sectionref{sec:oriEst-ekf}, the following matrices are needed for implementing the \gls{ekf} for pose estimation with quaternion states
\begin{subequations}
\label{eq:app-poseEst-ekfMeas}
\begin{align}
F_t &= \begin{pmatrix} \mathcal{I}_3 & T \mathcal{I}_3 & \tfrac{T^2}{2} \left. \tfrac{\partial R^\text{nb}_{t \mid t}}{\partial q^\text{nb}_{t \mid t}} \right|_{q^\text{nb}_{t \mid t}=\hat{q}^\text{nb}_{t \mid t}} y_{\text{a},t} \\
0 & \mathcal{I}_3 & T \left. \tfrac{\partial R^\text{nb}_{t \mid t}}{\partial q^\text{nb}_{t \mid t}} \right|_{q^\text{nb}_{t \mid t}=\hat{q}^\text{nb}_{t \mid t}} y_{\text{a},t} \\
0 & 0 & \left( \expq ( \tfrac{T}{2} y_{\omega,t} ) \right)^\rightMult \end{pmatrix}, \quad
H = \mathcal{I}_3, \quad R = \Sigma_\text{p},
 \\
G_t &= \begin{pmatrix} \mathcal{I}_6 & 0 \\ 0 & -\tfrac{T}{2} \left( \hat{q}^\text{nb}_{t \mid t} \right)^\leftMult \tfrac{\diff \expq (e_{\omega,t})}{\diff e_{\omega,t}} \end{pmatrix}, 
\quad
Q = \begin{pmatrix} \Sigma_\text{a} & 0 & 0 \\ 0 & \Sigma_\text{a} & 0 \\ 0 & 0 & \Sigma_\omega \end{pmatrix}. 
\end{align}
\end{subequations}

\section{Extended Kalman filter with orientation deviation states}
In the time update of the pose estimation algorithm with orientation deviation states, the linearization point is again directly updated as in~\eqref{eq:oriEst-oriErrorEKF-dynUpdateLin}. The position and velocity states are updated according to the dynamic model~\eqref{eq:models-ssPose-dyn}. Furthermore, the matrices $Q$, $H$ and $R$ from~\eqref{eq:app-poseEst-ekfMeas} are needed for implementing the \gls{ekf} for pose estimation in combination with 
\begin{align}
F_t = \begin{pmatrix} \mathcal{I}_3 & T \mathcal{I}_3 & - \tfrac{T^2}{2}  [\tilde{R}_{t \mid t}^\text{nb} y_{\text{a},t} \times] \\
0 & \mathcal{I}_3 & - T [\tilde{R}_{t \mid t}^\text{nb} y_{\text{a},t} \times] \\
0 & 0 & \mathcal{I}_3 \end{pmatrix}
, \qquad
G_t = \begin{pmatrix} \mathcal{I}_6 & 0 \\ 0 & T \tilde{R}_{t+1 \mid t} \end{pmatrix}.
\end{align}

\chapter{Gyroscope Bias Estimation}
\label{app:estGyroBias}
In this appendix, we will introduce the necessary components to extend Algorithms~\ref{alg:oriEst-smoothingOpt}--\ref{alg:oriEst-ekfOriError} to estimate an unknown gyroscope bias. Note that the measurement models are independent of the gyroscope bias. The dynamic models and time update equations are adapted as described below.

\section{Smoothing in an optimization framework}
To include a gyroscope bias in \Algorithmref{alg:oriEst-smoothingOpt}, a prior on the bias is included, leading to an additional term in the objective function as
\begin{align}
e_{\delta_\omega} = \delta_\omega, \qquad e_{\delta_\omega} \sim \mathcal{N}(0, \Sigma_{\delta_\omega}).
\end{align}
The derivatives in~\eqref{eq:oriEst-der} are complemented with
\begin{align}
\tfrac{\diff e_{\omega,t} }{\diff \delta_\omega} = \mathcal{I}_3, \qquad \tfrac{\diff e_{\delta_\omega} }{\diff \delta_\omega} = \mathcal{I}_3.
\end{align}

\section{Filtering in an optimization framework}
To include estimation of a gyroscope bias in \Algorithmref{alg:oriEst-filteringOpt},~\eqref{eq:oriEst-filtOptCost} needs to take the gyroscope bias into account as 
\begin{align}
\tilde{q}^{\text{nb},(0)}_{t+1} = \hat{q}^\text{nb}_t \odot \expq \left( \tfrac{T}{2} ( y_{\omega,t} - \hat{\delta}_{\omega,t} ) \right).
\end{align}
Modeling the dynamics of the gyroscope bias as a random walk, $e_{\text{f},t}$ is extended from~\eqref{eq:oriEst-costDerFiltMarg} to 
\begin{align}
e_{\text{f},t} = \begin{pmatrix} 
\oriError^\text{n}_{t}- 2 \logq \left(\hat{q}_{t-1}^\text{nb} \odot \expq (\tfrac{T}{2} y_{\omega,t-1} ) \odot  \tilde{q}_{t}^\text{bn} \right) \\
\delta_{\omega,t} - \hat{\delta}_{\omega,t-1}
\end{pmatrix}, 
\end{align}
where $\tfrac{\diff e_{\text{f},t}}{\diff x_t} = \mathcal{I}_6$ and $x_t = \begin{pmatrix} (\oriError^\text{n}_{t})^\Transp & \delta_{\omega,t}^\Transp \end{pmatrix}^\Transp$.

The covariance matrix $P_{t+1 \mid t}$ is given by $P_{t+1 \mid t} = F_{t} P_{t \mid t} F_{t}^\Transp + G_{t} Q G_{t}^\Transp$ with
\begin{subequations}
\label{eq:app-estGyroBias-timeUpdateFiltOpt}
\begin{align}
F_t &= \begin{pmatrix}
\mathcal{I}_3 & -T \tilde{R}_{t+1}^{\text{nb},(0)} \\
0 & \mathcal{I}_3
\end{pmatrix}, \label{eq:app-estGyroBias-timeUpdateFiltOpt-F} \\
G_t &= \begin{pmatrix} T \tilde{R}_{t+1}^{\text{nb},(0)} & 0 \\ 0 & \mathcal{I}_3 \end{pmatrix}, \qquad
Q = \begin{pmatrix} \Sigma_\omega & 0 \\ 0 & \Sigma_{\delta_{\omega,t}} \end{pmatrix}. \label{eq:app-estGyroBias-timeUpdateFiltOpt-GQ}
\end{align} 
\end{subequations}
Note that a prior on the gyroscope bias at $t = 1$ also needs to be included. 

\section{Extended Kalman filter with quaternion states}
To include estimation of an unknown gyroscope bias in \Algorithmref{alg:oriEst-ekfQuat}, the matrices for the time update of the \gls{ekf} need to be adapted to
\begin{subequations}
\begin{align}
F_t &= \begin{pmatrix}
\left( \expq (\tfrac{T}{2} (y_{\omega,t} - \hat{\delta}_{\omega,t \mid t})) \right)^\rightMult & -\tfrac{T}{2} \left( \hat{q}^\text{nb}_{t \mid t} \right)^\leftMult \tfrac{\diff \expq (\delta_{\omega,t})}{\diff \delta_{\omega,t}} \\
0_{3 \times 4} & \mathcal{I}_3
\end{pmatrix}, \\
Q &= \begin{pmatrix} \Sigma_\omega & 0 \\ 0 & \Sigma_{\delta_{\omega,t}} \end{pmatrix}, \qquad G_t = \begin{pmatrix} -\tfrac{T}{2} \left( \hat{q}^\text{nb}_{t \mid t} \right)^\leftMult \tfrac{\diff \expq (e_{\omega,t})}{\diff e_{\omega,t}} & 0 \\ 0 & \mathcal{I}_3 \end{pmatrix}.
\end{align}
\end{subequations}
Note that also for this algorithm a prior on the gyroscope bias needs to be included. Furthermore, the renormalization of the covariance in~\eqref{eq:oriEst-ekfQuatRenorm} needs to be adjusted to the size of the covariance matrix as
\begin{align}
    \label{eq:app-estGyroBias-relinCovFiltOpt}
    J_t = \begin{pmatrix} \tfrac{1}{\| \tilde{q}^\text{nb}_{t \mid t} \|_2^3} \tilde{q}^\text{nb}_{t \mid t} \left( \tilde{q}^\text{nb}_{t \mid t} \right)^\Transp & 0 \\ 0 & \mathcal{I}_3 \end{pmatrix}.
\end{align}

\section{Extended Kalman filter with orientation deviation states}
To include estimation of an unknown gyroscope bias in \Algorithmref{alg:oriEst-ekfOriError}, the update of the linearization point in the time update takes the estimate of the gyroscope bias into account as
\begin{align}
\tilde{q}^\text{nb}_{t+1 \mid t} = \tilde{q}^\text{nb}_{t \mid t} \odot \expq \left( \tfrac{T}{2} ( y_{\omega,t} - \hat{\delta}_{\omega,t \mid t} ) \right).
\end{align}
The matrices $F_t$ and $G_t$ for the time update are adapted to 
\begin{align}
F_t = \begin{pmatrix}
\mathcal{I}_3 & -T \tilde{R}_{t+1 \mid t}^{\text{nb}} \\
0 & \mathcal{I}_3
\end{pmatrix}, \qquad  
G_t = \begin{pmatrix} T \tilde{R}_{t+1 \mid t}^{\text{nb}} & 0 \\ 0 & \mathcal{I}_3 \end{pmatrix}, \end{align}
and $Q$ is given in~\eqref{eq:app-estGyroBias-timeUpdateFiltOpt-GQ}. A prior on the gyroscope bias also needs to be included.

\bibliographystyle{plain} 
\bibliography{references}

\end{document}